\documentclass[10pt,twocolumn,letterpaper]{article}

\usepackage{times}
\usepackage[numbers,sort&compress]{natbib}
\usepackage{amsmath,amssymb,mathtools}
\usepackage{booktabs,multirow,tabularx,array,colortbl}
\usepackage{graphicx}
\usepackage[dvipsnames]{xcolor}
\usepackage{xspace}
\usepackage{microtype}
\usepackage{enumitem}
\usepackage{lineno}
\usepackage[font=footnotesize,skip=5pt,subrefformat=parens]{subcaption}
\DeclareCaptionJustification{halfline}{%
  \leftskip=0pt\rightskip=0pt\parfillskip=0pt plus 0.49\linewidth
  \tolerance=9999\emergencystretch=2em\looseness=-1}
\definecolor{cvprblue}{rgb}{0.21,0.49,0.74}
\definecolor{causaldubrow}{RGB}{235,247,243}
\newcommand{\oursrow}{\rowcolor{causaldubrow}}
\newsavebox{\fittablebox}
\newcommand{\fittable}[3]{%
  \setlength{\tabcolsep}{0pt}\sbox{\fittablebox}{#3}%
  \setlength{\tabcolsep}{\dimexpr(#1-\wd\fittablebox)/(2*#2)\relax}%
  \ifdim\tabcolsep>9pt \setlength{\tabcolsep}{9pt}\fi
  #3}
\usepackage[breaklinks,colorlinks,allcolors=cvprblue]{hyperref}

\newcommand{\method}{VidDub\xspace}
\newcommand{\causalvariant}{Causal two-step VidDub\xspace}
\IfFileExists{tables/model_efficiency.tex}{% Generated model-size interface. Values are backed by model_efficiency_receipt.json.
\newcommand{\ParamsWavTwoLip}{36M}
\newcommand{\ParamsVideoReTalking}{484M}
\newcommand{\ParamsMuseTalk}{971M}
\newcommand{\ParamsDiffTwoLip}{102M}
\newcommand{\ParamsLatentSync}{1.39B}
\newcommand{\ParamsXDub}{14.32B}
\newcommand{\ParamsTeacher}{2.81B}
\newcommand{\ParamsLipForcing}{14.53B}

\newcommand{\ParamsVidDub}{2.81B}
}{}
\IfFileExists{tables/v2_benchmark_stats.tex}{% Generated by analyze_v2_benchmark.py; do not edit manually.
\newcommand{\AIGCPromptCount}{100,000}

\newcommand{\AIGCBenchCount}{256}

\newcommand{\AIGCEmotionCount}{42}
\newcommand{\AIGCSettingCount}{38}
\newcommand{\AIGCGestureCount}{73}
\newcommand{\AIGCAgeCount}{15}

\newcommand{\VTwoDynamicsRows}{%
HDTF & 30 & 1.30 & 1.03 & 6.17 \\
VoxCeleb2 & 30 & 4.83 & 1.82 & 5.86 \\
TalkVid & 500 & 2.14 & 1.45 & 8.94 \\
VidDub-Bench & 256 & 4.57 & 2.29 & 6.31 \\
}
}{}
\IfFileExists{tables/v2_ranking_audit.tex}{% Generated by analyze_v2_rankings.py; do not edit manually.
\newcommand{\VTwoRankingRows}{%
HDTF & 0.92 & -0.10 \\
VoxCeleb2 & 0.98 & -0.30 \\
TalkVid & 0.98 & 0.23 \\
VidDub-Bench & 0.85 & 0.60 \\
}
\newcommand{\RealAIGCRankRho}{0.71}
}{}

\newcommand{\Lfm}{\mathcal{L}_{\mathrm{FM}}}
\newcommand{\Lsync}{\mathcal{L}_{\mathrm{sync}}}
\newcommand{\Lret}{\mathcal{L}_{\mathrm{ret}}}
\setlist[itemize]{leftmargin=*,topsep=3pt,itemsep=2pt}
\newcommand{\arxivtablenobold}{\let\bfseries\mdseries\let\mathbf\mathrm}
\AddToHook{env/tabular/begin}{\arxivtablenobold}
\AddToHook{env/tabular*/begin}{\arxivtablenobold}
\AddToHook{env/tabularx/begin}{\arxivtablenobold}
\AddToHook{env/table/begin}{}
\AddToHook{env/table*/begin}{}
\title{VidDub: Replacement-Aware Supervision for Two-Step Video Dubbing}
\date{}
\author{
\begin{tabular}{c}
Yexin Liu$^{1,2}$, Yunlong Lin$^{2}$, Zehan Wang$^{3}$, Guangxin He$^{2}$, Xinli Xu$^{2}$, Jiajun Zha$^{1}$, Yaofu Liu$^{1}$\\
Yueze Wang$^{4}$, Bowen Ping$^{2}$, Yuanming Li$^{2}$, Amie Wang$^{5}$, Bernard Ghanem$^{6}$, Bing Li$^{6,*}$, Harry Yang$^{1,*}$\\[4pt]
$^{1}$HKUST \quad $^{2}$Tencent \quad $^{3}$ZJU \quad $^{4}$Independent Researcher \quad $^{5}$Tsinghua University \quad $^{6}$KAUST\\[3pt]
GitHub: \url{https://github.com/LAW1223/VidDub}\\[3pt]
$^{*}$Corresponding author
\end{tabular}
}

\begin{document}

\twocolumn[{%
\renewcommand\twocolumn[1][]{#1}%
\maketitle
\begin{center}
  \includegraphics[width=\textwidth]{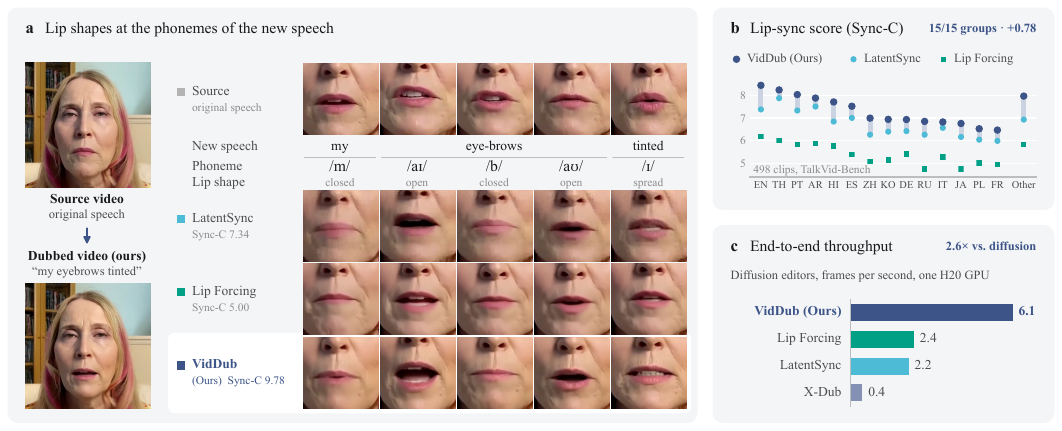}
  \vspace{-12pt}
  \captionof{figure}{\textbf{Resolving source-mouth conflict for multilingual, few-step dubbing.}
  (a) \method replaces the speech of a source video; mouth crops at phonemes of the new speech show closures and openings that follow the expected lip shape. Sync-C is SyncNet confidence (higher is better).
  (b) \method leads all 15 language groups on TalkVid-Bench (+0.78 Sync-C averaged over 498 clips).
  (c) End-to-end throughput of diffusion editors on one H20 GPU, from raw video and
  target speech to the final video file (GAN-based Wav2Lip: 5.2).}
  \vspace{-3pt}
  \label{fig:overview}
\end{center}
}]

\begin{abstract}
Video dubbing aims to synthesize lip movements that match target speech while
preserving the speaker identity, head pose, and background of a source video.
Existing editors are predominantly trained via self-reconstruction under original
paired audio, even though the source mouth becomes a conflicting target when driving
speech is replaced. We introduce \emph{replacement-aware cross-audio supervision}:
paired-audio samples use full-frame reconstruction, whereas unpaired replacement-speech
samples mask out source-mouth reconstruction losses and supervise the mouth solely
through audio--visual synchronization while preserving unmasked facial regions and
background. Because offline dubbing provides the complete source video and target
speech beforehand, we present \method, a full-context diffusion editor that uses
guided trajectory regression to transfer CFG-guided teacher trajectories to a
two-step student and distribution matching to recover visual quality after
compression. We will release the curated training dataset, VidDub-Bench, fixed
paired- and cross-audio evaluation manifests, preprocessing
outputs, and unified evaluation code. Across three training seeds, masking out
conflicting source-mouth reconstruction improves cross-audio Sync-C by
$0.833$. At two NFE, the final student provides the strongest cross-audio
synchronization and visual speech recognition among the evaluated methods,
with competitive visual fidelity and source preservation.
\end{abstract}

\section{Introduction}
\label{sec:intro}

\looseness=-1
Video dubbing synthesizes lip movements that match replacement speech while
preserving the identity, head motion, and scene of a source video. Ideally, a model
would learn from triplets comprising a source video, replacement speech, and a
ground-truth dubbed video whose lip movements match that speech while all other
content remains unchanged. In practice, such ground-truth dubbed videos are
unavailable at scale: collecting them requires the same person to produce different
utterances under nearly identical appearance, pose, motion, and scene conditions.

\looseness=-1
To bypass this missing supervision, existing feed-forward, inpainting, latent-space,
and diffusion editors commonly adopt same-audio self-reconstruction
\cite{wav2lip,videoretalking,dinet,diff2lip,musetalk,latentsync,omnisync}: the model
receives a source video and its original speech and learns to reproduce that video.
This target is valid for paired speech but conflicts with replacement dubbing.
Under replacement speech, the source mouth articulates different content, whereas
identity, pose, background, and non-mouth regions remain valid targets. Applying
reconstruction losses to the whole source therefore penalizes the mouth changes
required by the target speech.

\looseness=-1
This leads to our central question: when ground-truth dubbed videos are unavailable, \textit{how can a model learn from unpaired replacement speech without treating the conflicting source-mouth articulation as ground truth}, while still using valid source supervision to preserve identity, pose, and scene content?

\looseness=-1
We address this question with \emph{replacement-aware cross-audio supervision}.
For paired speech, the complete source video remains a valid reconstruction target.
For unpaired replacement speech, we remove flow-matching, pixel, and perceptual
reconstruction losses inside the mouth, guide lip articulation with audio--visual
synchronization, and retain reconstruction outside the mouth. Replacement speech
therefore controls the region that must change, while the source video continues to
constrain identity, pose, and background.

\looseness=-1
We then distill the replacement-aware full-context teacher into a full-context
two-step student. Matched same- and cross-audio trajectories transfer its CFG-guided
predictions while preserving bidirectional temporal attention, and guided
distribution matching restores visual quality lost during trajectory compression
\cite{dmd2}. These stages have distinct roles: replacement-aware supervision learns
the response to target speech, trajectory regression transfers that response to two
steps, and distribution matching recovers visual quality. Full-context modeling
also matches offline dubbing, where the complete source video and speech are
available before inference. A controlled same-backbone study shows that causal
execution lowers first-output latency for streaming, whereas full-context execution
provides higher offline throughput, lower peak memory, and better visual speech
recognition.

\looseness=-1
We evaluate \method across diverse cross-audio and cross-domain benchmarks.
VidDub-Bench (Sec.~\ref{sec:data-construction}) provides generated videos from 256
disjoint source videos with fixed same- and cross-audio pairings, and we will release the
curated training clips developed in this work. Together with HDTF, VoxCeleb2,
TalkVid-Bench, 20 minute-long pairs (Long-20), independent visual speech recognition (VSR)
\cite{hdtf,voxceleb2,talkvid,vsr_ma}, temporal-shift calibration, and non-mouth
preservation metrics, the evaluation covers speech content, temporal
synchronization, visual fidelity, source preservation, and minute-long stability.
Across three training seeds, masking out conflicting source-mouth reconstruction
yields a $+0.833\pm0.040$ cross-audio Sync-C improvement. Using only two denoising
steps (2 NFE), trajectory regression retains the teacher's synchronization, and
distribution matching recovers most of the FVD increase introduced by compression.
The final model improves cross-audio synchronization and visual speech recognition
over the evaluated external methods while retaining competitive fidelity and source
preservation.

\begin{itemize}[nosep]
  \item \looseness=-1 \textbf{Replacement-aware cross-audio supervision} uses unpaired
  replacement speech without reconstructing the inconsistent source mouth, while
  retaining reconstruction outside the mouth and target-speech synchronization.
  \item \looseness=-1 \textbf{Two-step diffusion distillation}
  transfers matched same- and cross-audio teacher trajectories to a full-context
  two-step student (2 NFE), followed by distribution matching to recover visual
  quality after compression.
  \item \looseness=-1 \textbf{Open training dataset and evaluation benchmark}
  provide curated training clips and VidDub-Bench with fixed same- and cross-audio
  pairings, unified evaluation scripts, and VSR, temporal-alignment, preservation,
  perceptual, and long-video evaluations.
\end{itemize}

\begin{figure*}[!t]
  \centering
  \includegraphics[width=\textwidth]{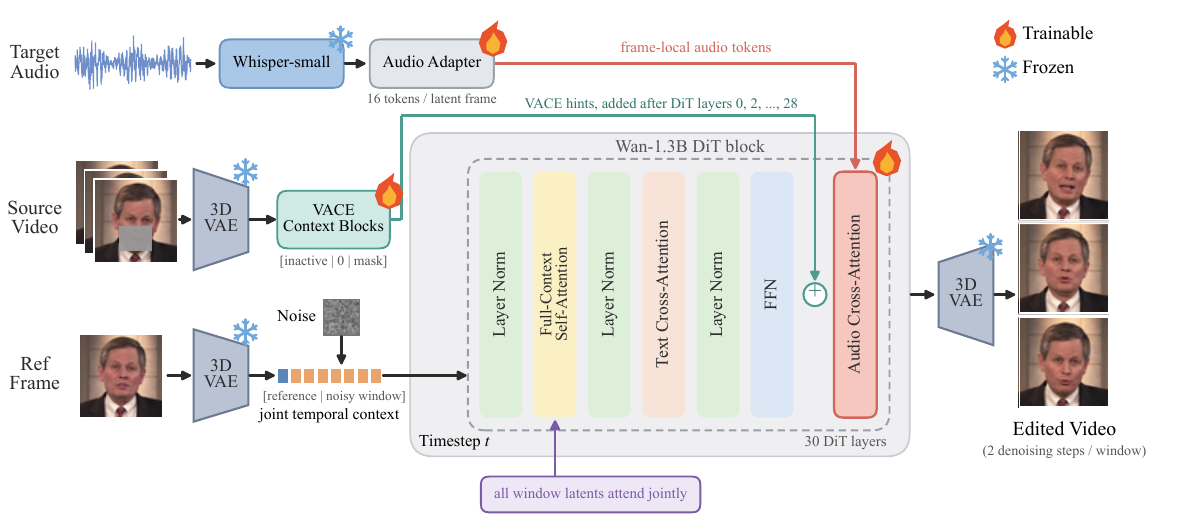}
  \caption{\textbf{Full-context two-step dubbing architecture.} Masked source frames
  and a reference image preserve speaker appearance and pose, target-speech tokens
  drive lip articulation, and bidirectional self-attention jointly denoises the
  complete latent window in two steps.}
  \label{fig:method}
\end{figure*}

\section{Related Work}
\label{sec:related}

\noindent\textbf{Visual speech generation and editing.}
Source-video dubbing methods edit mouth motion in existing footage to match target
speech \cite{wav2lip,videoretalking,dinet,talklip,diff2lip,musetalk,flashlips,latentsync,omnisync,xdub}.
Person-specific neural rendering methods fit a subject from video
\cite{adnerf,geneface,synctalk}. Reference-conditioned portrait animation
synthesizes head motion from an identity image, optionally with an additional pose
source \cite{pcavs,sadtalker,difftalk,dreamtalk,aniportrait,hallo,echomimic,hallo3,vasa1,omniavatar},
and audio-driven generators synthesize talking faces from speech without an identity
reference \cite{diffusedheads}.
Unlike standard self-reconstruction training that penalizes deviations from the
source mouth, our replacement-aware objective masks out source-mouth reconstruction
on unpaired replacement speech during training.
X-Dub \cite{xdub} addresses the lack of paired dubbing data through generative
bootstrapping that constructs pseudo-paired training examples. In contrast, we
train directly with unpaired replacement speech: we retain valid reconstruction
supervision outside the mouth while removing conflicting source-mouth reconstruction
losses, without generating pseudo-paired dubbed videos. We isolate this effect with
matched teacher controls before distilling the resulting full-context editor to two
steps.

\noindent\textbf{Few-step video diffusion.}
Progressive, consistency, adversarial, and distribution-matching distillation methods
accelerate diffusion sampling \cite{progressive_distillation,consistency,lcm,add,dmd,dmd2}.
Causal Forcing \cite{causalforcing,causalforcingpp} and Lip Forcing
\cite{lipforcing} combine autoregressive causal attention with few-step distillation
for online streaming. In contrast, offline video dubbing provides the full source
video and replacement speech prior to generation. We retain full-window bidirectional
attention while distilling the CFG-guided teacher into a two-step student
\cite{cfg,flowmatching}.

\noindent\textbf{Dubbing benchmarks and evaluation metrics.}
HDTF, VoxCeleb2, and TalkVid-Bench provide high-resolution and in-the-wild speech
domains \cite{hdtf,voxceleb2,talkvid}; we add generated footage and analyze its
source-motion statistics. In addition to SyncNet-based synchronization
\cite{syncnet}, we use an independently trained visual speech recognizer
\cite{vsr_ma} to evaluate lip-reading intelligibility alongside pixel
fidelity, identity similarity \cite{arcface}, head pose accuracy \cite{insightface},
FVD \cite{fvd}, and temporal stability over 60-second videos.

\section{Method}
\label{sec:method}

\begin{figure*}[t]
  \centering
  \includegraphics[width=\textwidth]{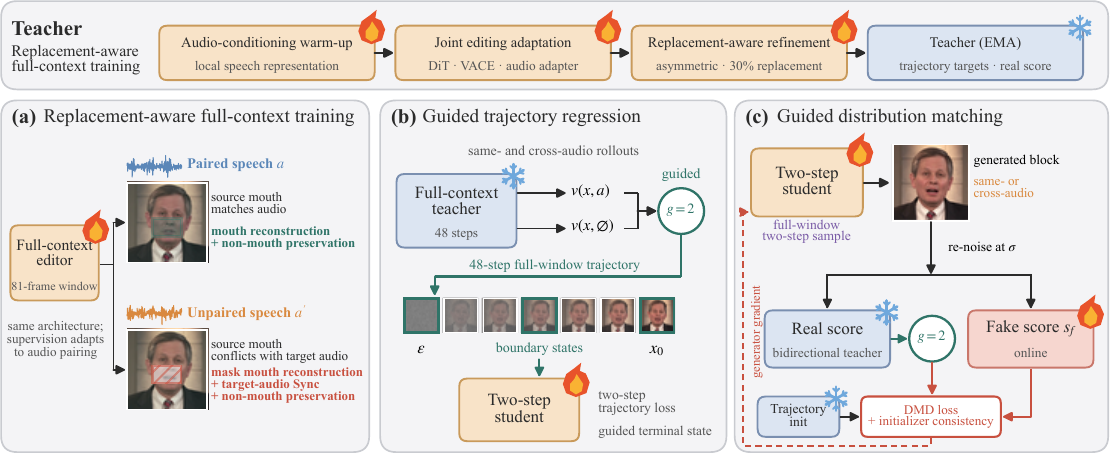}
  \vspace{-9pt}
  \caption{\textbf{Three-stage training and distillation pipeline.}
  Top: full-context teacher training via audio warm-up, joint editing adaptation, and
  replacement-aware refinement. (a) Replacement-aware training masks out source-mouth
  reconstruction under unpaired speech. (b) Guided trajectory regression distills the
  48-step CFG-guided teacher into a two-step student. (c) Guided distribution matching recovers visual realism in the final two-step model.}
  \label{fig:causal-distillation}
\end{figure*}

\noindent\textbf{Overview.}
Given a source video and speech, \method synthesizes synchronized lip
movements while preserving speaker identity, head pose, and background. Our core
insight is that ground-truth reconstruction targets must be decoupled by region when
the driving speech is replaced. Figure~\ref{fig:method} illustrates our
full-context editing architecture, which conditions on masked source frames, a
reference frame, and frame-aligned audio tokens. Figure~\ref{fig:causal-distillation}
presents the three-stage training pipeline: (a) replacement-aware teacher training,
(b) two-step guided trajectory regression, and (c) guided distribution matching. Sections~\ref{sec:task}--\ref{sec:training} detail each component.

\subsection{Replacement-Aware Cross-Audio Supervision}
\label{sec:task}

In source-video dubbing, supervision cannot rely solely on spatial masks; it must
account for whether the speech matches the source video. When original speech
is replaced by unpaired speech, the source mouth motion conflicts with the target
utterance, whereas the unmasked face and background remain valid ground truth.

Given source frames $V^{\rm src}=\{I_t^{\rm src}\}_{t=1}^{T}$ and target waveform
$A^{\rm tgt}$, we aim to synthesize $\hat V$ with target-driven lip articulation
while preserving all non-mouth visual content. The source mouth provides a
consistent reconstruction target only when $A^{\rm tgt}=A^{\rm src}$.

For clean video latent $z$ and Gaussian noise $\epsilon\!\sim\!\mathcal N(0,I)$, flow
matching \cite{flowmatching} interpolates
$x_\sigma=(1-\sigma)z+\sigma\epsilon$ with velocity target
$v^*=\epsilon-z$ and spatially weighted loss
$\Lfm^q=\mathbb E[q\|v_\theta(x_\sigma,\sigma,c)-v^*\|_2^2]/\mathbb E[q]$.
Max-pooling the VACE edit mask yields the latent mouth mask $M$. Paired same-audio
samples use spatial weight $q=w(M)=1+\lambda_M M$ with $\lambda_M=2$ to emphasize mouth reconstruction,
whereas unpaired replacement-speech samples use $q=\bar M=1-M$, masking out
source-mouth velocity regression entirely.
Note that $z$ is encoded directly from the unedited source video $V^{\rm src}$
rather than from a synthetic pseudo-target. Its mouth region therefore reflects the
original speech $A^{\rm src}$; under replacement speech
$A^{\rm tgt}\ne A^{\rm src}$, $\bar M$ keeps the flow-matching loss from pulling
the generated lips back toward it.

This region-decoupled design yields distinct objectives for the two training regimes.
Paired same-audio samples optimize full-frame reconstruction:
\begin{equation}
\begin{aligned}
\mathcal L_{\rm same}={}&\Lfm^{w(M)}
+\lambda_{\rm pix}\mathcal L_{\rm pix}
+\lambda_{\rm perc}\mathcal L_{\rm LPIPS}\\
&+\lambda_{\rm sync}\Lsync
+\lambda_{\rm keep}\mathcal L_{\rm keep}.
\end{aligned}
\label{eq:same-loss}
\end{equation}
For unpaired replacement speech $a'$, source-mouth reconstruction terms are
masked out and $\Lsync$ uses $a'$:
\begin{equation}
\begin{aligned}
\mathcal L_{\rm cross}={}&\Lfm^{\bar M}
+\lambda_{\rm sync}\Lsync
+\lambda_{\rm keep}\mathcal L_{\rm keep}^{\bar M}.
\end{aligned}
\label{eq:cross-loss}
\end{equation}
On a decoded 16-frame window, $\mathcal L_{\rm pix}$ is mouth-region $L_1$ loss,
$\mathcal L_{\rm LPIPS}$ \cite{lpips} is perceptual loss over four sampled mouth
crops, and
$\mathcal L_{\rm keep}^{\bar M}=\|\bar M\odot(D(z)-V^{\rm src})\|_1/\|\bar M\|_1$
measures pixel preservation outside the mouth mask, where $D$ is the VAE decoder. In teacher refinement,
$(\lambda_{\rm pix},\lambda_{\rm perc},\lambda_{\rm sync},\lambda_{\rm keep})
=(0,.10,.02,0)$; pixel and keep terms are used only in student stages. Under replacement
speech, no flow, pixel, or perceptual loss regresses the mouth toward $V^{\rm src}$;
masked source frames provide appearance context while target speech governs lip
synchronization.

Lip synchronization is supervised by a frozen LatentSync-1.6 StableSyncNet
\cite{latentsync}, which extracts embeddings from 16 lower-face frames and the
aligned 16-kHz mel-spectrogram via cosine-log loss
$\Lsync=-\log[\max(\cos(e_v,e_a),10^{-6})]$ without temporal lag search. Gradients
backpropagate through the frozen synchronizer and VAE decoder into the generator.
Test-time evaluation instead uses the standard S3FD \cite{s3fd}/SyncNet-v2
\cite{syncnet,voxceleb2} benchmark pipeline,
complemented by an independent LRS3 VSR reader in Sec.~\ref{sec:sync-validity}.

\subsection{Full-Context Diffusion Architecture}
\label{sec:architecture}

While replacement-aware supervision defines the training objective, our architecture
retains bidirectional self-attention across the entire latent window so that every
generated frame attends to full temporal context. This design matches offline video
dubbing, where the complete source video and target speech are available before
inference.

Given masked source latents $u$, reference latent $r$, target-speech embeddings $a$,
mouth mask $M$, and initial noise $\epsilon$, the two-step generator predicts the
edited latent window in two denoising steps:
\begin{equation}
  \hat z = G_\phi^{(2)}(\epsilon \mid u,r,a,M).
  \label{eq:task}
\end{equation}
We build upon Wan-1.3B with VACE context conditioning \cite{wan,vace} and a 287M
audio adapter over frozen Whisper-small features \cite{whisper}. VACE encodes the
masked source video and reference image into visual context tokens. The audio
adapter maps 16 local Whisper tokens per latent frame into speech conditioning
tokens, and bidirectional DiT self-attention jointly processes all 21 video latents
within each 81-frame window. The two-step student shares the exact network
architecture of the teacher, compressing only the sampling schedule from 48 steps to
2 steps while absorbing classifier-free guidance (CFG) into a single forward pass.

\noindent\textbf{Relation to causal few-step methods.}
Lip Forcing \cite{lipforcing} represents the causal two-step design paradigm, and
Causal Forcing develops causal ODE/DMD objectives
\cite{causalforcing,causalforcingpp}. Our controlled same-backbone comparison
(Supplementary Sec.~B) shows that causal attention reduces first-frame latency but
does not improve offline throughput or output quality. \method therefore combines replacement-aware
cross-audio supervision with direct full-context distillation.

\IfFileExists{tables/lipforcing_rows.tex}{\providecommand{\LipForcingHdtfRow}{Lip Forcing & \ParamsLipForcing & 2 & 8.429 & 6.919 & 0.9517 & 31.37 & 6.884 & 8.284 & 33.80 \\}
\providecommand{\LipForcingVoxRow}{Lip Forcing & \ParamsLipForcing & 2 & 6.408 & 7.831 & 0.9466 & 33.12 & 4.871 & 9.242 & 41.03 \\}
\providecommand{\LipForcingTalkRow}{Lip Forcing & \ParamsLipForcing & 2 & 5.914 & 8.283 & 0.7698 & 24.48 & 5.610 & 8.648 & 12.68 \\}
\providecommand{\LipForcingPublicSummaryRow}{Lip Forcing & \ParamsLipForcing & 2 & 6.884 / 33.80 & 4.871 / 41.03 & 5.610 / 12.68 \\}
}{}

\begin{table*}[t]
\centering
\caption{\textbf{HDTF comparison.} Params counts all neural modules loaded for
inference. HDTF-30 FVD values are small-sample diagnostics.}
\vspace{-3pt}
\label{tab:hdtf}
\fontsize{9pt}{9.5pt}\selectfont
\setlength{\tabcolsep}{2.3pt}
\fittable{\textwidth}{10}{%
\begin{tabular}{l l r rr rr rr r}
\toprule
& & & \multicolumn{4}{c}{Reconstruction} & \multicolumn{3}{c}{Replacement Dubbing} \\
\cmidrule(lr){4-7}\cmidrule(lr){8-10}
Method & Params & NFE & Sync-C$\uparrow$ & Sync-D$\downarrow$ & SSIM$\uparrow$ & PSNR$\uparrow$ & Sync-C$\uparrow$ & Sync-D$\downarrow$ & FVD$\downarrow$ \\
\midrule
Unedited source & -- & 0 & 8.723 & 6.666 & -- & -- & 1.283 & 14.483 & -- \\
\midrule
\multicolumn{10}{l}{\textit{Feed-forward and one-step editors}} \\
\addlinespace[1pt]
Wav2Lip~\cite{wav2lip} & \ParamsWavTwoLip & 1 & 9.177 & 6.555 & 0.9233 & 28.97 & 8.782 & 6.972 & 129.51 \\
VideoReTalking~\cite{videoretalking} & \ParamsVideoReTalking & 1 & 8.801 & 6.593 & 0.8981 & 27.85 & 8.312 & 6.988 & 105.72 \\
MuseTalk 1.5~\cite{musetalk} & \ParamsMuseTalk & 1 & 8.471 & 6.895 & 0.9489 & \textbf{33.74} & 7.116 & 7.880 & 40.10 \\
\midrule
\multicolumn{10}{l}{\textit{Bidirectional diffusion editors (20--50 NFE)}} \\
\addlinespace[1pt]
Diff2Lip~\cite{diff2lip} & \ParamsDiffTwoLip & 25 & 9.670 & \textbf{5.535} & 0.9182 & 30.46 & 8.680 & 6.310 & 99.70 \\
LatentSync 1.6~\cite{latentsync} & \ParamsLatentSync & 20 & 9.363 & 6.029 & \textbf{0.9656} & 32.95 & 8.160 & 7.035 & \textbf{28.05} \\
X-Dub~\cite{xdub} & \ParamsXDub & 50 & 7.792 & 7.810 & 0.8364 & 26.62 & 5.027 & 10.271 & 66.05 \\
Full-context Teacher-48 (ours) & \ParamsTeacher & 48 & \textbf{10.398} & 5.561 & 0.9221 & 30.15 & \textbf{10.162} & \textbf{5.744} & \textbf{26.22} \\
\midrule
\multicolumn{10}{l}{\textit{Two-step diffusion editors}} \\
\addlinespace[1pt]
Lip Forcing~\cite{lipforcing} & \ParamsLipForcing & 2 & 8.429 & 6.919 & 0.9517 & 31.37 & 6.884 & 8.284 & 33.80 \\
\oursrow\method & \ParamsVidDub & 2 & 10.371 & 5.654 & 0.9213 & 30.03 & 10.102 & 5.856 & 27.66 \\
\bottomrule
\end{tabular}}
\vspace{-2pt}
\end{table*}

\enlargethispage{8pt}
\subsection{Trajectory and Distribution Distillation}
\label{sec:training}

Our pipeline proceeds in three stages: learning cross-audio lip
synchronization in the teacher, compressing multi-step sampling into two steps via
trajectory regression, and restoring visual realism via guided distribution matching.

\noindent\textbf{(a) Replacement-aware teacher training.}
Following audio warm-up and joint editing adaptation, teacher refinement mixes 70\%
paired same-audio clips with 30\% unpaired replacement-speech clips, where the
replacement speech is taken from the preceding training sample. All training clips
use 81-frame (3.375\,s) windows aligned from the start of the waveform. Non-mouth
masked flow supervision anchors identity, pose, and background, while $\Lsync$
drives lip movements to match the replacement speech. This 300-update LoRA-64
refinement starts from a mature joint editor, and all reported results use its
EMA-smoothed weights. The stage is short because it changes only the low-rank
residual and audio adapter of an already trained editor; the pretrained visual
representation is retained. We fixed the update budget before final evaluation and
verify the replacement effect with three independent seeds and matched
no-replacement controls. The stage corrects the objective of a trained editor; it
does not show that 300 updates suffice to train a dubbing model from scratch.

\begin{figure*}[t]
  \centering
  \includegraphics[width=0.95\textwidth]{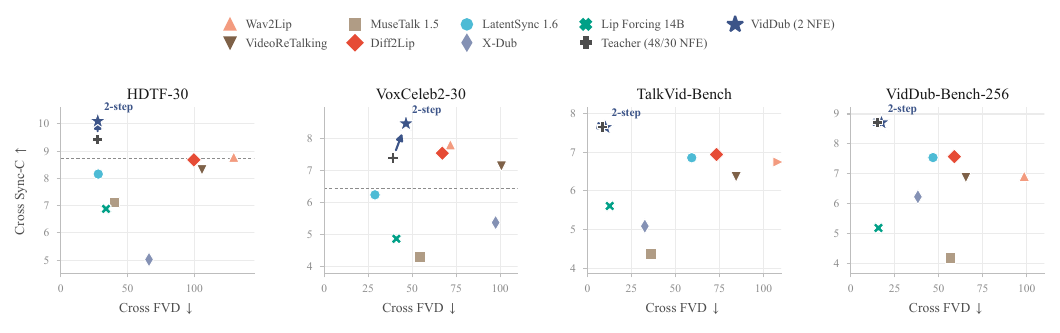}
  \vspace{-0.2em}
  \caption{\textbf{Cross-audio Sync-C versus FVD} on four benchmarks. Dashed lines
  show real-video same-audio Sync-C.}
  \label{fig:external}
  % \vspace{-3pt}
\end{figure*}

\noindent\textbf{(b) Guided trajectory regression.}
Matched full-context trajectories compress the 48-step teacher into a student with
two denoising steps. The teacher uses classifier-free audio guidance
$v^{(g)}=v(x,\varnothing)+g[v(x,a)-v(x,\varnothing)]$. Source, audio, noise,
boundary state, and teacher target are matched within the same full-window trajectory.
For boundary state $x_\tau$, guided-teacher terminal latent $\tilde z_{T,g}$, and
two-step student map $\Psi_\phi^{(2)}$, we optimize
\begin{equation}
\mathcal L_{\rm traj}=
\mathbb E\!\left[
\left\|\Psi_\phi^{(2)}(x_\tau,c)-\operatorname{sg}
(\tilde z_{T,g})\right\|_{w(M)}^2\right]
+\lambda_q^{\rm traj}\mathcal L_q^{s/c},
\label{eq:trajectory}
\end{equation}
where
{\small
\begin{align}
\mathcal L_q^s={}&
\lambda_{\rm pix}^{\rm traj}\mathcal L_{\rm pix}
+\lambda_{\rm perc}^{\rm traj}\mathcal L_{\rm LPIPS}
+\lambda_{\rm sync}^{\rm traj}\Lsync
+\lambda_{\rm keep}^{\rm traj}\mathcal L_{\rm keep}, \\
\mathcal L_q^c={}&
\lambda_{\rm sync}^{\rm traj}\Lsync
+\lambda_{\rm keep}^{\rm traj}\mathcal L_{\rm keep}^{\bar M}.
\end{align}
}
Here $\|\cdot\|_{w(M)}^2$ is the $w(M)$-weighted squared error, and
$\mathcal L_q^{s/c}$ selects $\mathcal L_q^s$ for same audio and
$\mathcal L_q^c$ for replacement speech.
The fixed training set contains 4,096 teacher trajectories, evenly divided between
paired- and cross-audio conditions; replacement utterances share the source domain and language
but come from different source videos. The regression target is the teacher output
conditioned on replacement speech rather than the original source mouth. Matching
all trajectory conditions distills classifier-free guidance into a student that
requires one conditional forward pass per denoising step, while
$\operatorname{sg}$ blocks teacher gradients.
Trajectory regression runs for 400 generator updates on this set and initializes
the two-step student; it is followed by a 200-update fake-score warm-up and 100 DMD
generator updates, each paired with five fake-score updates. The budgets are small
because both stages start from a converged teacher or trajectory student and fit a
fixed endpoint and schedule; they do not relearn appearance or motion. The
supplementary material reports controls at the same 100-update DMD budget.

\begin{table*}[t]
\centering
\caption{\textbf{TalkVid-Bench comparison} on 500 requested clips. ``Scored S/C''
denotes evaluated paired-/cross-audio outputs. Metrics use each method's scored
outputs; text rankings use the shared 308-clip subset reported in the supplement.}
\vspace{-3pt}
\label{tab:talkvid-main}
\fontsize{8.5pt}{9pt}\selectfont
\setlength{\tabcolsep}{2.3pt}
\fittable{\textwidth}{11}{%
\begin{tabular}{l l r c rr rr rr r}
\toprule
& & & & \multicolumn{4}{c}{Reconstruction} & \multicolumn{3}{c}{Replacement Dubbing} \\
\cmidrule(lr){5-8}\cmidrule(lr){9-11}
Method & Params & NFE & Scored S/C & Sync-C$\uparrow$ & Sync-D$\downarrow$ &
SSIM$\uparrow$ & PSNR$\uparrow$ & Sync-C$\uparrow$ & Sync-D$\downarrow$ & FVD$\downarrow$ \\
\midrule
\multicolumn{11}{l}{\textit{Feed-forward and one-step editors}} \\
\addlinespace[1pt]
Wav2Lip~\cite{wav2lip} & \ParamsWavTwoLip & 1 & 499/499 & 7.209 & 7.420 & 0.9218 & 29.83 & 6.750 & 7.932 & 111.19 \\
VideoReTalking~\cite{videoretalking} & \ParamsVideoReTalking & 1 & 402/402 & 6.787 & 7.552 & 0.8948 & 28.00 & 6.365 & 7.868 & 84.51 \\
MuseTalk 1.5~\cite{musetalk} & \ParamsMuseTalk & 1 & 500/499 & 4.968 & 9.106 & 0.9305 & 29.60 & 4.371 & 9.625 & 36.09 \\
\midrule
\multicolumn{11}{l}{\textit{Bidirectional diffusion editors (20--50 NFE)}} \\
\addlinespace[1pt]
Diff2Lip~\cite{diff2lip} & \ParamsDiffTwoLip & 25 & 425/403 & 7.775 & 6.383 & 0.9157 & 30.60 & 6.944 & 7.016 & 73.37 \\
LatentSync 1.6~\cite{latentsync} & \ParamsLatentSync & 20 & 499/499 & 7.516 & 6.965 & 0.8949 & 31.09 & 6.860 & 7.512 & 59.27 \\
X-Dub~\cite{xdub} & \ParamsXDub & 50 & 500/500 & 5.492 & 9.129 & 0.6980 & 21.86 & 5.087 & 9.437 & 32.63 \\
Full-context Teacher-48 (ours) & \ParamsTeacher & 48 & 500/500 &
7.741 & 7.170 & 0.9125 & 28.84 & 7.663 & 7.266 & 8.51 \\
\midrule
\multicolumn{11}{l}{\textit{Two-step diffusion editors}} \\
\addlinespace[1pt]
Lip Forcing~\cite{lipforcing} & \ParamsLipForcing & 2 & 499/499 & 5.914 & 8.283 & 0.7698 & 24.48 & 5.610 & 8.648 & 12.68 \\
\oursrow\method & \ParamsVidDub & 2 & 500/500 &
7.740 & 7.216 & 0.9125 & 28.78 & 7.646 & 7.309 & 10.25 \\
\bottomrule
\end{tabular}}
\vspace{-2pt}
\end{table*}

\begin{table*}[t]
\centering
\caption{\textbf{VidDub-Bench comparison} on each method's successfully evaluated
outputs.}
\vspace{-3pt}
\label{tab:causaldub-bench}
\fontsize{9pt}{9.5pt}\selectfont
\setlength{\tabcolsep}{2.3pt}
\IfFileExists{tables/causaldub256_rows.tex}{\newcommand{\CausalDubBenchRows}{%
\multicolumn{10}{l}{\textit{Feed-forward and one-step editors}} \\
\addlinespace[1pt]
Wav2Lip & \ParamsWavTwoLip & 1 & 8.269 & 7.193 & 0.9063 & 29.44 & 6.922 & 8.536 & 98.62 \\
VideoReTalking & \ParamsVideoReTalking & 1 & 8.097 & 7.088 & 0.8773 & 26.77 & 6.873 & 8.069 & 65.43 \\
MuseTalk 1.5 & \ParamsMuseTalk & 1 & 6.430 & 8.673 & 0.9078 & 28.27 & 4.184 & 10.670 & 56.73 \\
\midrule
\multicolumn{10}{l}{\textit{Bidirectional diffusion editors (20--50 NFE)}} \\
\addlinespace[1pt]
Diff2Lip & \ParamsDiffTwoLip & 25 & \textbf{9.724} & \textbf{5.643} & 0.9125 & 30.36 & 7.575 & 7.309 & 58.87 \\
LatentSync 1.6 & \ParamsLatentSync & 20 & 9.389 & 6.269 & 0.9099 & \textbf{32.05} & 7.544 & 7.730 & 46.77 \\
X-Dub public & \ParamsXDub & 50 & 7.576 & 7.912 & 0.6954 & 22.07 & 6.234 & 9.071 & 38.17 \\
Full-context Teacher-48 (ours) & \ParamsTeacher & 48 & 8.966 & 6.889 & \textbf{0.9141} & 29.22 & \textbf{8.718} & \textbf{7.093} & \textbf{15.23} \\
\midrule
\multicolumn{10}{l}{\textit{Two-step diffusion editors}} \\
\addlinespace[1pt]
Lip Forcing & \ParamsLipForcing & 2 & 6.672 & 8.475 & 0.7820 & 25.36 & 5.192 & 9.971 & 15.78 \\
\oursrow\method & \ParamsVidDub & 2 & 8.993 & 6.905 & 0.9123 & 29.04 & 8.709 & 7.140 & 17.50 \\
}
}{}
\fittable{\textwidth}{10}{%
\begin{tabular}{l l r rr rr rr r}
\toprule
& & & \multicolumn{4}{c}{Reconstruction} & \multicolumn{3}{c}{Replacement Dubbing} \\
\cmidrule(lr){4-7}\cmidrule(lr){8-10}
Method & Params & NFE & Sync-C$\uparrow$ & Sync-D$\downarrow$ & SSIM$\uparrow$ & PSNR$\uparrow$ & Sync-C$\uparrow$ & Sync-D$\downarrow$ & FVD$\downarrow$ \\
\midrule
\CausalDubBenchRows
\bottomrule
\end{tabular}}
\vspace{-4pt}
\end{table*}

\noindent\textbf{(c) Guided distribution matching.}
Trajectory regression matches individual teacher endpoints but does not directly
match the distribution of generated videos. We therefore apply guided distribution
matching (DMD) after trajectory distillation. A student sample $z$ is re-noised to
$x_\sigma$ at noise level $\sigma$, and the frozen guided real score and online fake
score define $\hat z_r=x_\sigma-\sigma v_r$ and
$\hat z_f=x_\sigma-\sigma v_f$. Applied to the complete latent window, DMD2
\cite{dmd2} uses
\begin{equation}
\begin{aligned}
d&=(\hat z_f-\hat z_r)/
\max(\operatorname{mean}|z-\hat z_r|,10^{-6}),\\
\mathcal L_{\rm DMD}&=\tfrac12\|z-\operatorname{sg}(z-d)\|_{w(M)}^2
+\lambda_{\rm ret}\Lret\\
&\quad+\lambda_q^{\rm DMD}\mathcal L_q^{\rm DMD,s/c}.
\end{aligned}
\label{eq:dmd}
\end{equation}
Here $d$ is treated as a constant, and $\Lret$ is a $w(M)$-weighted MSE to the output
of the frozen trajectory student under the same noise and schedule.
The auxiliary DMD objectives are
\begin{align}
\mathcal L_q^{\rm DMD,s}={}&
\lambda_{\rm pix}^{\rm DMD}\mathcal L_{\rm pix}
+\lambda_{\rm perc}^{\rm DMD}\mathcal L_{\rm LPIPS}
+\lambda_{\rm keep}^{\rm DMD}\mathcal L_{\rm keep},\\
\mathcal L_q^{\rm DMD,c}={}&
\lambda_{\rm keep}^{\rm DMD}\mathcal L_{\rm keep}^{\bar M}.
\end{align}
$\mathcal L_q^{\rm DMD,s/c}$ selects the objective for the current audio condition.
The fake score takes five updates per generator update, and half of the DMD
batches use replacement speech.

\subsection{Long-Video Inference}
\label{sec:long-inference}
Long videos are processed using overlapping 81-frame windows conditioned on the
corresponding source frames. Each window uses two denoising steps; consecutive
windows advance by 64 frames and blend the 17-frame overlap. This sliding-window
scheme bounds computation and memory while retaining bidirectional context within
each window.

\subsection{Training Data and Benchmark Construction}
\label{sec:data-construction}

\noindent\textbf{Motivation and structured prompt generation.}
Existing talking-face datasets do not jointly cover the demographic, linguistic,
expressive-motion, and scene diversity required by our dubbing setting
\cite{hdtf,voxceleb2,talkvid,speakervid}. To expand this coverage without
identity-privacy constraints, we construct a taxonomy-driven prompt suite of
100,000 talking-human descriptions using Qwen3.5-397B-A17B \cite{qwen35}
(48,303 Chinese and
51,697 English). Each prompt combines verbatim speech transcripts with stratified
attributes spanning ethnicity and cultural region, age group, gender presentation,
42 emotion dynamics, 73 upper-body gestures, and 38 scene settings, and is rendered
into candidate talking-human videos via MiniMax-H3 inference \cite{minimaxh3}.

\noindent\textbf{Filtering and audio--visual alignment.}
Generated and in-the-wild candidates still exhibit face-tracking drift, extreme
poses, and audio--visual timing offsets. We track the dominant face, extract stable
$512{\times}512$ square crops with union lower-face edit masks, and discard windows
with missing or unstable tracks, invalid crop geometry, extreme head pose, or weak
mouth motion. Following the synchronization-based alignment used by LatentSync
\cite{latentsync}, we apply SyncNet \cite{syncnet} to estimate audio--visual confidence and temporal
offsets, rejecting low-correspondence clips and shifting retained waveforms by the
estimated offset before extracting 81-frame windows at 24\,FPS. This pipeline yields
85,435 model-ready H3 training clips alongside filtered public clips.

\noindent\textbf{VidDub-Bench construction.}
From held-out H3 source videos strictly disjoint from the training and development
splits, we select 256 balanced source videos (128 Chinese, 128 English) to form
VidDub-Bench. It is source-disjoint but shares the training set's MiniMax-H3 domain.
Each
source video is paired with its original audio track for
reconstruction and a fixed replacement waveform drawn from a different source video
for dubbing evaluation, while source frames, face crops, spatial masks, reference
frames, random seeds, and durations remain identical across conditions
(Table~\ref{tab:causaldub-bench}).

\section{Experiments}
\label{sec:experiments}

\IfFileExists{tables/lipforcing_rows.tex}{\providecommand{\LipForcingHdtfRow}{Lip Forcing & \ParamsLipForcing & 2 & 8.429 & 6.919 & 0.9517 & 31.37 & 6.884 & 8.284 & 33.80 \\}
\providecommand{\LipForcingVoxRow}{Lip Forcing & \ParamsLipForcing & 2 & 6.408 & 7.831 & 0.9466 & 33.12 & 4.871 & 9.242 & 41.03 \\}
\providecommand{\LipForcingTalkRow}{Lip Forcing & \ParamsLipForcing & 2 & 5.914 & 8.283 & 0.7698 & 24.48 & 5.610 & 8.648 & 12.68 \\}
\providecommand{\LipForcingPublicSummaryRow}{Lip Forcing & \ParamsLipForcing & 2 & 6.884 / 33.80 & 4.871 / 41.03 & 5.610 / 12.68 \\}
}{}

\subsection{Experimental Setup}

Tables~\ref{tab:teacher-replacement-control}
and~\ref{tab:sft-ablation}, together with the supplementary three-seed study,
evaluate replacement-aware cross-audio supervision and the
teacher$\rightarrow$trajectory$\rightarrow$DMD progression. The remaining tables
test the final model across source domains and video lengths.
Tables~\ref{tab:hdtf}, \ref{tab:talkvid-main}, and
\ref{tab:causaldub-bench} evaluate HDTF-30, TalkVid-Bench-500, and
VidDub-Bench-256 using standardized 81-frame, 24\,FPS, $512{\times}512$ inputs;
full VoxCeleb2-30 results and native source protocols are supplementary.
Table~\ref{tab:long20-external} evaluates 20 fixed 60-second pairs. Model
selection uses a disjoint 256-source development split (dev-256).

Cross-method rankings use the clips successfully evaluated for every method, while
method-specific tables report generation coverage separately. Unless noted,
uncertainty is estimated
with paired source-level bootstrap intervals.
Teacher-48 rows use 48 Euler steps and audio CFG 2, and distillation uses the same
schedule; the teacher-refinement controls in
Tables~\ref{tab:teacher-replacement-control} and~\ref{tab:sft-ablation} use 30 steps.
All external methods use their released checkpoints and recommended inference
settings without target-domain adaptation. Their training corpora, face crops, and
native resolutions therefore differ from ours, and weak reconstruction or VSR
values, particularly for X-Dub and Lip Forcing, reflect the released checkpoints
under our common evaluation protocol.

Same-audio reconstruction cannot directly determine whether generated articulation
follows replacement speech. We therefore report cross-audio synchronization across
four complementary source domains: high-resolution HDTF \cite{hdtf}, in-the-wild
VoxCeleb2 \cite{voxceleb2}, web-video TalkVid-Bench \cite{talkvid}, and the
source-disjoint generated
VidDub-Bench. Ranking correlations and decoded source-motion statistics in
Supplementary Sec.~C further show that reconstruction metrics on a single captured
domain do not substitute for direct cross-audio and cross-source evaluation.

\begin{table}[!t]
\centering
\caption{\textbf{Cross-audio supervision.} Teacher cross-audio Sync-C/FVD with
paired-only versus replacement-aware training.}
% \vspace{-2pt}
\label{tab:teacher-replacement-control}
\fontsize{8.5pt}{9.0pt}\selectfont
\setlength{\tabcolsep}{2.2pt}
\fittable{\columnwidth}{3}{%
\begin{tabular}{lrr}
\toprule
Benchmark & Paired-audio only & Replacement-aware \\
\midrule
HDTF-30 & 7.986 / 26.40 & \textbf{9.422} / 27.49 \\
TalkVid-Bench-50 & 6.460 / 26.14 & \textbf{7.414} / 26.51 \\
H3-Val-50 & 6.952 / 37.68 & \textbf{8.086} / 38.09 \\
VoxCeleb2-30 & 5.729 / 39.95 & \textbf{7.382} / \textbf{39.11} \\
\bottomrule
\end{tabular}}
\vspace{-4pt}
\end{table}

\begin{table}[!t]
\centering
\caption{\textbf{Teacher training and full-context distillation on HDTF-30.}
Compare within blocks; S/C denote same-/cross-audio.}
\vspace{-2pt}
\label{tab:sft-ablation}
\fontsize{8.7pt}{9.2pt}\selectfont
\setlength{\tabcolsep}{2.2pt}
\fittable{\columnwidth}{4}{%
\begin{tabular}{lrrr}
\toprule
Stage & S-SSIM$\uparrow$ & C-Sync-C$\uparrow$ & C-FVD$\downarrow$ \\
\midrule
\multicolumn{4}{l}{\textit{Teacher refinement (30-NFE teachers)}} \\
Joint adaptation & 0.9230 & 5.825 & 33.50 \\
Quality-only refinement & 0.9233 & 6.104 & 29.56 \\
Replacement after quality & 0.9216 & 9.135 & 31.59 \\
\oursrow Replacement-aware teacher & \textbf{0.9279} & \textbf{9.422} & \textbf{27.49} \\
\midrule
\multicolumn{4}{l}{\textit{Distillation (48-NFE teacher / 2-NFE students)}} \\
Replacement-aware teacher & 0.9221 & \textbf{10.162} & \textbf{26.22} \\
Trajectory-only two-step student & \textbf{0.9226} & 10.113 & 31.07 \\
\oursrow\quad + distribution matching & 0.9213 & 10.102 & 27.66 \\
\bottomrule
\end{tabular}}
\vspace{-4pt}
\end{table}

\subsection{Comparison with Prior Methods}

\noindent\textbf{Captured and web sources.}
Figure~\ref{fig:external} and Tables~\ref{tab:hdtf} and~\ref{tab:talkvid-main}
compare the final model on captured and web video.
\method nearly matches Teacher-48 Sync-C on HDTF and TalkVid-Bench
(10.102/10.162 and 7.646/7.663), exceeds all external methods in Sync-C and FVD,
and processes all 500 TalkVid-Bench clips; the teacher has lower FVD.

\IfFileExists{tables/lipforcing_rows.tex}{\providecommand{\LipForcingHdtfRow}{Lip Forcing & \ParamsLipForcing & 2 & 8.429 & 6.919 & 0.9517 & 31.37 & 6.884 & 8.284 & 33.80 \\}
\providecommand{\LipForcingVoxRow}{Lip Forcing & \ParamsLipForcing & 2 & 6.408 & 7.831 & 0.9466 & 33.12 & 4.871 & 9.242 & 41.03 \\}
\providecommand{\LipForcingTalkRow}{Lip Forcing & \ParamsLipForcing & 2 & 5.914 & 8.283 & 0.7698 & 24.48 & 5.610 & 8.648 & 12.68 \\}
\providecommand{\LipForcingPublicSummaryRow}{Lip Forcing & \ParamsLipForcing & 2 & 6.884 / 33.80 & 4.871 / 41.03 & 5.610 / 12.68 \\}
}{}

\noindent\textbf{Generated sources.}
Table~\ref{tab:causaldub-bench} reports each method's valid outputs; the ranking on
the 245 clips valid for every method is supplementary. \method nearly matches Teacher-48 Sync-C and exceeds all
external baselines. Its 17.50 FVD trails Lip Forcing's 15.78 and Teacher-48's 15.23,
with complete generation coverage on all 256 sources.

\subsection{Effect of Replacement-Aware Supervision}

Table~\ref{tab:teacher-replacement-control} first compares teachers trained with and without the
replacement-aware branch under matched initialization and data. All 160 paired
evaluation items across the four benchmarks improve in cross-audio Sync-C, and the
paired 95\% source-bootstrap intervals lie fully above zero on all four sets.
The FVD change is mixed: it rises slightly on HDTF-30, TalkVid-Bench-50, and
H3-Val-50 and falls on VoxCeleb2-30; TalkVid-Bench-50 and H3-Val-50 are held-out
teacher-development subsets.
Table~\ref{tab:sft-ablation} shows both the teacher-development path and direct
full-context distillation. The trajectory-only two-step student preserves
synchronization but increases FVD; distribution matching recovers most of the
increase (31.07 to 27.66) while preserving cross-audio Sync-C.
A three-seed factorial in the supplement separates source-mouth masking, matched
audio dropout, and timestep policy. Masking produces the dominant main effect,
showing that the gain follows from removing conflicting mouth supervision rather
than auxiliary sampling choices. Independent
visual-only recognition on the same outputs yields consistent WER and CER
improvements, linking the synchronization gain to more accurate articulation of the
replacement speech.

\begin{table}[!t]
\centering
\caption{\textbf{Independent visual-only WER} ($\downarrow$) from an LRS3 reader
against shared Whisper references. HDTF/Vox2/VDB (VidDub-Bench) use 30/28/121 English clips;
Teacher-48 uses CFG 2. WER may exceed 1.0 because of insertions.}
\vspace{-1pt}
\label{tab:vsr-wer}
\fontsize{8.7pt}{9.2pt}\selectfont
\setlength{\tabcolsep}{2.1pt}
\fittable{\columnwidth}{4}{%
\begin{tabular}{lrrr}
\toprule
Method & HDTF & Vox2 & VDB \\
\midrule
Real (same audio) & 0.171 & 0.579 & 0.711 \\
\midrule
Wav2Lip~\cite{wav2lip} & 1.093 & 1.009 & 1.020 \\
VideoReTalking~\cite{videoretalking} & 0.915 & 1.056 & 0.975 \\
Diff2Lip~\cite{diff2lip} & 0.957 & 0.963 & 0.990 \\
MuseTalk~\cite{musetalk} & 0.987 & 1.068 & 1.014 \\
LatentSync~\cite{latentsync} & 1.019 & 1.049 & 0.953 \\
X-Dub~\cite{xdub} & 0.952 & 0.949 & 0.949 \\
Lip Forcing~\cite{lipforcing} & 1.093 & 0.977 & 0.977 \\
\midrule
Teacher-48 (ours) & \textbf{0.379} & \textbf{0.724} & \textbf{0.682} \\
\oursrow\method & 0.503 & 0.752 & 0.758 \\
\bottomrule
\end{tabular}}
\vspace{-4pt}
\end{table}

\subsection{Effect of Two-Step Distillation}

\noindent\textbf{Trajectory regression and distribution matching.}
Matched same- and cross-audio trajectories transfer the full-context teacher
predictions without changing its bidirectional temporal attention, after which
distribution matching aligns the student's generated distribution with real videos.
Table~\ref{tab:sft-ablation} reports the
teacher$\rightarrow$trajectory$\rightarrow$DMD path; same-budget final-stage
controls are supplementary. Removing replacement batches from the final stage does
not remove cross-audio supervision altogether, because the frozen trajectory
initializer was trained on cross-audio trajectories.

\subsection{Speech Validation and Source Preservation}

\label{sec:sync-validity}

\noindent\textbf{Temporal-shift calibration.}
We shift the audio of 40 VidDub-Bench outputs per method by known frame offsets.
SyncNet follows each imposed shift exactly, and \method returns zero on all
unshifted clips, verifying temporal sensitivity. Because generated outputs can
exceed real-video Sync-C (10.371 vs.\ 8.723 on HDTF), the metric does not measure
naturalness or absolute speech correctness.

\noindent\textbf{Independent visual speech recognition.}
Table~\ref{tab:vsr-wer} uses an LRS3 visual-only reader \cite{vsr_ma}, independent
of the training synchronizer, against shared Whisper references \cite{whisper}.
\method has lower WER than every external method on all three sources, while the
teacher remains lower on VidDub-Bench (0.682 vs.\ 0.758). Since external outputs
approach WER 1.0 and real-video controls span 0.171--0.711, WER is read
relatively, not as calibrated phoneme accuracy.

\IfFileExists{tables/human_study_compact.tex}{\begin{table}[!t]
\centering
\caption{\textbf{Blinded human evaluation.} 15 bilingual raters,
50 sources; entries are VidDub/Opp./Tie.}
\vspace{-6pt}
\label{tab:human-study}
\fontsize{7.1pt}{7.6pt}\selectfont
\begin{tabular*}{\columnwidth}{@{\extracolsep{\fill}}lccccc@{}}
\toprule
Opponent & Sync & Real. & Identity & Stability & Overall \\
\midrule
Teacher & 28/29/93 & 35/38/77 & 18/18/114 & 29/31/90 & 38/37/75 \\
LatentSync & 123/4/23 & 124/19/7 & 46/100/4 & 47/101/2 & 73/77/0 \\
Lip Forcing & 99/28/23 & 109/32/9 & 56/22/72 & 65/70/15 & 98/47/5 \\
\bottomrule
\end{tabular*}
\vspace{-6pt}
\end{table}
}{}
\noindent\textbf{Blinded human evaluation.}
Table~\ref{tab:human-study} shows that \method matches Teacher-48 overall and is
preferred over Lip Forcing except in stability. Against LatentSync, it wins on
synchronization and realism but trails on identity and stability.

\noindent\textbf{Identity and non-mouth preservation.}\label{sec:preservation}
Table~\ref{tab:preservation} compares corresponding source frames, excluding the
editable lower face for Upper-L1; identity and pose use ArcFace \cite{arcface} and
InsightFace \cite{insightface}. \method combines high cross-audio Sync-C
with the lowest Upper-L1, while LatentSync obtains higher identity and lower pose error;
the result is a trade-off between synchronization and preservation.

\begin{table}[!t]
\centering
\caption{\textbf{Identity/non-mouth preservation on VidDub-Bench.}
Upper-L1 is percent change; $n$ counts valid outputs.}
\vspace{-3pt}
\label{tab:preservation}
\IfFileExists{tables/preservation_rows.tex}{\newcommand{\PreservationRows}{%
Wav2Lip & 6.922 & 1.820 & 0.909 & 0.459 & 256 \\
Diff2Lip & 7.575 & 1.283 & 0.862 & 0.518 & 256 \\
LatentSync & 7.544 & 1.605 & 0.913 & 0.422 & 254 \\
Lip Forcing & 5.192 & 3.761 & 0.900 & 0.928 & 256 \\
Full-context teacher & 8.718 & 1.207 & 0.872 & 0.573 & 256 \\
\oursrow\method & 8.709 & 1.199 & 0.875 & 0.579 & 256 \\
}
}{}
\fontsize{8.6pt}{9.1pt}\selectfont
\setlength{\tabcolsep}{1.9pt}
\fittable{\columnwidth}{6}{%
\begin{tabular}{lrrrrr}
\toprule
Method & Sync-C$\uparrow$ & Upper-L1$\downarrow$ & ID$\uparrow$ & Pose$\downarrow$ & $n$ \\
\midrule
\PreservationRows
\bottomrule
\end{tabular}}
\vspace{-2pt}
\end{table}

\vspace{0pt}
\subsection{Long-Video Evaluation and Efficiency}
\label{sec:efficiency}

\begin{table}[!t]
\centering
\caption{\textbf{Long-20 comparison.} Twenty 60\,s pairs, all valid for every method, use overlapping windows
conditioned on source frames.
Last-C is final-5\,s Sync-C; $\Delta$Keep is the final-minus-initial preservation
change.}
\vspace{-2pt}
\label{tab:long20-external}
\fontsize{7.7pt}{8.3pt}\selectfont
\setlength{\tabcolsep}{1.2pt}
\fittable{\columnwidth}{7}{%
\begin{tabular}{lrrrrrr}
\toprule
Method & S-C$\uparrow$ & S-D$\downarrow$ & Last-C$\uparrow$ &
Keep-L1$\downarrow$ & Chroma$\downarrow$ & $\Delta$Keep \\
\midrule
Wav2Lip~\cite{wav2lip} & 7.861 & 7.484 & 8.032 & 3.045 & 1.045 & +0.009 \\
LatentSync~\cite{latentsync} & 7.771 & 7.083 & 7.844 & 4.018 & 0.397 & +0.685 \\
\oursrow\method & \textbf{9.399} & \textbf{6.162} &
\textbf{9.585} & \textbf{2.493} & \textbf{0.250} & $-0.034$ \\
\bottomrule
\end{tabular}}
\vspace{-4pt}
\end{table}

Table~\ref{tab:long20-external} scores every 5\,s window and the final 5\,s segment.
\method has the strongest mean/final synchronization and lowest measured
non-mouth/color changes across these 20 sequences.
A controlled same-backbone study in the supplementary material shows that
full-context execution has higher throughput, 37\% lower peak memory, and lower VSR
error in all six comparisons than its causal counterpart.
\vspace{-0.2em}
\section{Conclusion}
\vspace{0pt}
\label{sec:conclusion}

\looseness=-1
We presented \method, a full-context two-step dubbing model with replacement-aware
cross-audio supervision. Under unpaired speech, it masks source-mouth reconstruction
while preserving non-mouth regions and target-speech synchronization. Full-context
trajectory regression produces a two-step student, and distribution matching
improves visual quality.
Across four source domains, \method improves replacement-speech synchronization and
visual speech recognition while maintaining competitive fidelity and source
preservation.
Two limitations remain. \method trades some identity and pose preservation for
synchronization relative to LatentSync, and VidDub-Bench shares the generated-video
domain of the training set, where the FVD of \method trails Lip Forcing and the
teacher.

{\small
\bibliographystyle{ieeenat_fullname}
\bibliography{main}
}

\clearpage
\section*{Supplementary Material}

This supplement provides data construction, replacement-pairing rules, training
details, and extended evaluations supporting the main paper. It expands the evidence
for replacement-aware cross-audio supervision, full-context two-step distillation,
and 60-second overlapping-window inference. It also reports a controlled causal variant that isolates
the quality--latency effect of causal conversion. Complete benchmark protocols,
uncertainty, qualitative comparisons, reproducibility details, and ethical safeguards
are included.

\appendix
\setcounter{section}{0}
\renewcommand{\thesection}{\Alph{section}}

\section{Guide to the Supplement}
\label{supp:overview}

This supplement records implementation and evaluation details needed for
reproduction but omitted from the main paper. Intermediate checkpoints are
reported separately from the final replacement-aware teacher, the full-context two-step
student, and the controlled causal variant.
A fixed 16-source subset is used only for diagnosis and a separate 256-source
split is used only for development and model selection. The sole internal benchmark reported in
the paper is the source-disjoint 256-video VidDub-Bench.

The material proceeds from data and teacher construction to few-step distillation,
the controlled causal study, and final evaluation. Controlled ablations measure
target-speech sensitivity and the effects of cross-audio supervision; subsequent
sections document both full-context and causal training paths. Final benchmark
tables cover HDTF, VoxCeleb2, TalkVid-Bench, and VidDub-Bench, followed by complete
same/cross-audio examples and evaluation protocols.

\section{Why the Final Editor Remains Full-Context}
\label{supp:causal-comparison}

The complete source video and replacement speech are available before offline
dubbing, so progressive block emission is not required. We nevertheless trained a
matched \causalvariant using the same backbone family, data, teacher,
trajectory objective, DMD recipe, preprocessing, and evaluation manifests.
Table~\ref{tab:supp-causal-design} isolates the resulting design trade-off.

\begin{table*}[t]
\centering
\caption{\textbf{Matched full-context and causal two-step designs.} Timing is the
median of three warmed runs on one H20 after conditioning is prepared. The 81-frame
time includes denoising and VAE decoding; quality uses identical frozen manifests.}
\label{tab:supp-causal-design}
\small
\setlength{\tabcolsep}{5.0pt}
\begin{tabularx}{\textwidth}{@{}Xrrrrrrr@{}}
\toprule
Design & 81-frame time & Prepared FPS & Peak GiB & VSR wins & Upper-L1 & ID & Pose \\
\midrule
Causal two-step & 8.654\,s & 9.4 & 21.96 & 0/6 & 1.079 & 0.861 & 0.614 \\
\oursrow Full-context two-step (\method) & \textbf{7.165\,s} & \textbf{11.3} &
\textbf{13.77} & \textbf{6/6} & \textbf{1.014} & \textbf{0.885} & \textbf{0.567} \\
\bottomrule
\end{tabularx}
\end{table*}

For bounded offline inference, the full-context model is faster
at 81, 241, 721, and 1,441 frames, uses less peak memory, improves all six visual-only
speech-recognition comparisons, and improves all three preservation measures on
every evaluated benchmark. The two designs are effectively tied on Long-20
synchronization. These results motivate the final full-context architecture while
retaining the \causalvariant as a controlled design study rather than the main method.

\section{Data Construction and Splits}
\label{supp:data}

\begin{figure}[t]
\centering
\includegraphics[width=0.90\columnwidth]{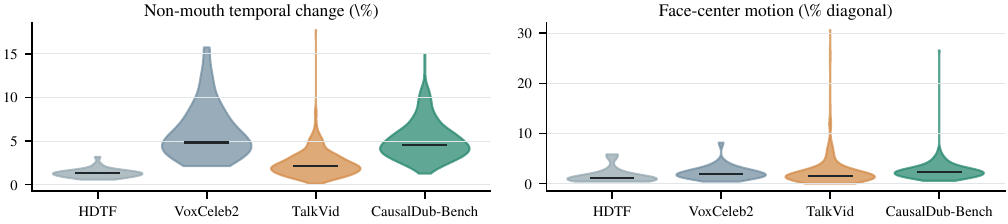}
\caption{\textbf{Source-motion distributions} from 12 sampled frames per video.
Non-mouth change excludes the lower face; center motion is normalized by frame
diagonal.}
\label{fig:supp-source-dynamics}
\end{figure}

\noindent\textbf{Benchmark audit.}
The H3 construction begins with \AIGCPromptCount{} structured talking-human
prompts. The frozen \AIGCBenchCount-source VidDub-Bench contains 81 frames per
source at 24 FPS and fixed replacement-speech partners. Joined prompt metadata contains
\AIGCEmotionCount{} emotion, \AIGCSettingCount{} setting,
\AIGCGestureCount{} gesture, and \AIGCAgeCount{} age labels. These are construction
annotations, so we additionally measure decoded-frame dynamics.
Table~\ref{tab:v2-dynamics} gives the exact medians summarized by
Fig.~\ref{fig:supp-source-dynamics}.

\noindent\textbf{Construction and release unit.}
We generate candidate talking-human clips from the structured prompt set,
then apply the face-track, crop, duration, and audio--visual filters described below.
The benchmark freezes 256 accepted source videos before final model evaluation.
Each source receives one same-audio track and one replacement waveform drawn from a
different source video; source frames, crop, mask, seed, and duration remain fixed
between tracks. Table~\ref{tab:supp-cdb-statistics} summarizes the released unit.

\begin{table}[t]
\centering
\caption{\textbf{VidDub-Bench construction statistics.} Duration and resolution
refer to the model-ready generated source clips; metadata counts denote distinct
labels represented by the frozen benchmark.}
\label{tab:supp-cdb-statistics}
\small
\setlength{\tabcolsep}{3.2pt}
\begin{tabularx}{\columnwidth}{@{}Xr@{}}
\toprule
Statistic & Value \\
\midrule
Structured prompts & \AIGCPromptCount \\
Frozen generated source videos & \AIGCBenchCount \\
Same-audio tracks / replacement pairs & 256 / 256 \\
Frames / duration / frame rate & 81 / 3.375\,s / 24\,FPS \\
Model-ready spatial resolution & $512\times512$ \\
Total unique-source duration & 14.4\,min \\
Emotion / setting labels & \AIGCEmotionCount{} / \AIGCSettingCount \\
Gesture / age-group labels & \AIGCGestureCount{} / \AIGCAgeCount \\
\bottomrule
\end{tabularx}
\end{table}

\noindent\textbf{Release scope.}
We will release the generated benchmark source clips, original and replacement
waveforms, fixed pair manifest, prompt metadata, preprocessing outputs, split and
exclusion manifests, and evaluation scripts. For third-party public benchmarks, the
release contains identifiers and deterministic manifests rather than redistributing
media whose licenses do not permit it.

\begin{table}[t]
\centering
\caption{Source-dynamics audit. Non-mouth $\Delta$ is mean absolute grayscale
change after excluding the detected lower-face region; center motion is normalized
by frame diagonal; scale CV uses detected face-box area.}
\label{tab:v2-dynamics}
\footnotesize
\setlength{\tabcolsep}{2.0pt}
\resizebox{\columnwidth}{!}{%
\begin{tabular}{@{}lrrrr@{}}
\toprule
Source set & $n$ & Non-mouth $\Delta$ & Center motion & Scale CV \\
\midrule
\VTwoDynamicsRows
\bottomrule
\end{tabular}}
\end{table}

\noindent\textbf{Ranking audit.}
For the nine methods present in every benchmark, we compute Spearman correlations
between same-audio Sync-C or SSIM and cross-audio Sync-C. Captured-to-generated
correlation uses the seven external methods and averages each method's cross-audio
Sync-C over HDTF, VoxCeleb2, and TalkVid before comparison with VidDub-Bench.
The reported coefficients summarize these method rankings.

\begin{table}[t]
\centering
\caption{\textbf{Metric- and domain-dependent rankings.} The first two columns compare
same-audio metrics against cross-audio Sync-C within each benchmark.}
\label{tab:supp-ranking-audit}
\small
\begin{tabularx}{\columnwidth}{@{}Xrr@{}}
\toprule
Benchmark & Same Sync-C & Same SSIM \\
\midrule
\VTwoRankingRows
\midrule
Captured$\rightarrow$AIGC cross Sync-C &
\multicolumn{2}{c}{$\rho=\RealAIGCRankRho$} \\
\bottomrule
\end{tabularx}
\end{table}

\noindent\textbf{Source domains.}
The teacher pool combines six domains. H3 \cite{minimaxh3} contains internally
generated talking-human
videos whose records include verbatim speech text, Chinese or English language,
subject attributes, emotion, setting, and the visual prompt. TalkVid
\cite{talkvid} and SpeakerVid-5M \cite{speakervid} provide natural web-video
diversity; HDTF \cite{hdtf} and VoxCeleb2 \cite{voxceleb2} provide standard public
talking-face data; Hallo3 \cite{hallo3} adds reference-portrait and motion diversity.
After filtering,
the final teacher pool contains 322,513 clips: 85,435 H3, 152,713 TalkVid, 57,727
SpeakerVid, 6,527 HDTF, 2,437 VoxCeleb2, and 17,674 Hallo3 clips. Domain-balanced
sampling uses relative weights 30/20/15/12/8/10 in the same order, with per-clip
quality weights inside each domain. The 85,435 retained H3 clips constitute the
released training set.

\noindent\textbf{Face and audio preprocessing.}
For each source video, preprocessing detects landmarks and tracks the dominant face.
Windows with missing or unstable tracks, invalid crop geometry, extreme pose,
insufficient mouth motion, or insufficient audio--visual correspondence are excluded.
SyncNet \cite{syncnet} estimates the temporal offset and confidence; retained audio is shifted by the
estimated offset before window extraction. Each example contains 81 frames at 24 fps,
a stable square face crop resized to $512\times512$, a union lower-face edit mask, one
mechanically selected identity reference, and a 16-kHz mono waveform. Preprocessing
caches the ground-truth crop, masked/inactive source condition, reference latent,
audio features, and empty-text context. For evaluation, frozen cross-audio benchmark
manifests use different source groups while keeping source video, mask, crop, seed,
and frame count fixed.

\noindent\textbf{Teacher and causal manifests.}
Causal adaptation samples 16,384 clips from the 322,513 eligible teacher clips without
replacement under the frozen quality/domain distribution. These clips represent
5,174 H3, 3,450 TalkVid, 2,587 SpeakerVid, 2,070 HDTF, 1,724 Hallo3, and
1,379 VoxCeleb2 clips, which sum to 16,384. The frozen manifest contains
15,346 distinct source IDs and 10,123 leak-free group keys; 1,038 source IDs
contribute a second temporal window. Training manifests, quality weights,
source-video identifiers, group keys, and selected checkpoint hashes are included
in the reproducibility package.

\noindent\textbf{Disjoint development and test sets.}
The development set contains 256 source videos and is used for checkpoint selection.
A fixed 16-source subset is used only for diagnosis. VidDub-Bench contains another
256 source videos, each represented by one 81-frame same-audio example and one fixed
cross-audio pairing, yielding 512 requested outputs per method. Tables report
same- and cross-audio metrics separately and retain track-specific validity; for
example, LatentSync has 254/256 valid cross-audio preservation outputs.
Source-video intersections are zero for train--development,
train--VidDub-Bench, and development--VidDub-Bench. The benchmark is frozen
before final evaluation; no training or checkpoint selection uses its outputs. The
benchmark shares the generated H3 domain with part of training but not source videos.
The 433 group keys derived from the 560 frozen HDTF-30, VoxCeleb2-30, and
TalkVid-Bench clips (26 person, 25 speaker, and 382 YouTube-video groups) match
the exclusion list and
have zero intersection with the eligible pool, causal-training manifest, and
development manifest. Thus all evaluation identities and source videos are held out
at the declared group granularity, while the represented source domains may also
contribute different clips to the training pool.
Every same/cross comparison fixes source video, seed, crop geometry, and frame count.
Because the source mouth is not the target under cross audio, source-mouth
reconstruction metrics are intentionally omitted for that track.

\section{Multilingual Qualitative Results}
\label{supp:diversity}

\begin{figure*}[t]
\centering
\IfFileExists{figures/showcase_diversity.png}{%
  \includegraphics[width=\textwidth]{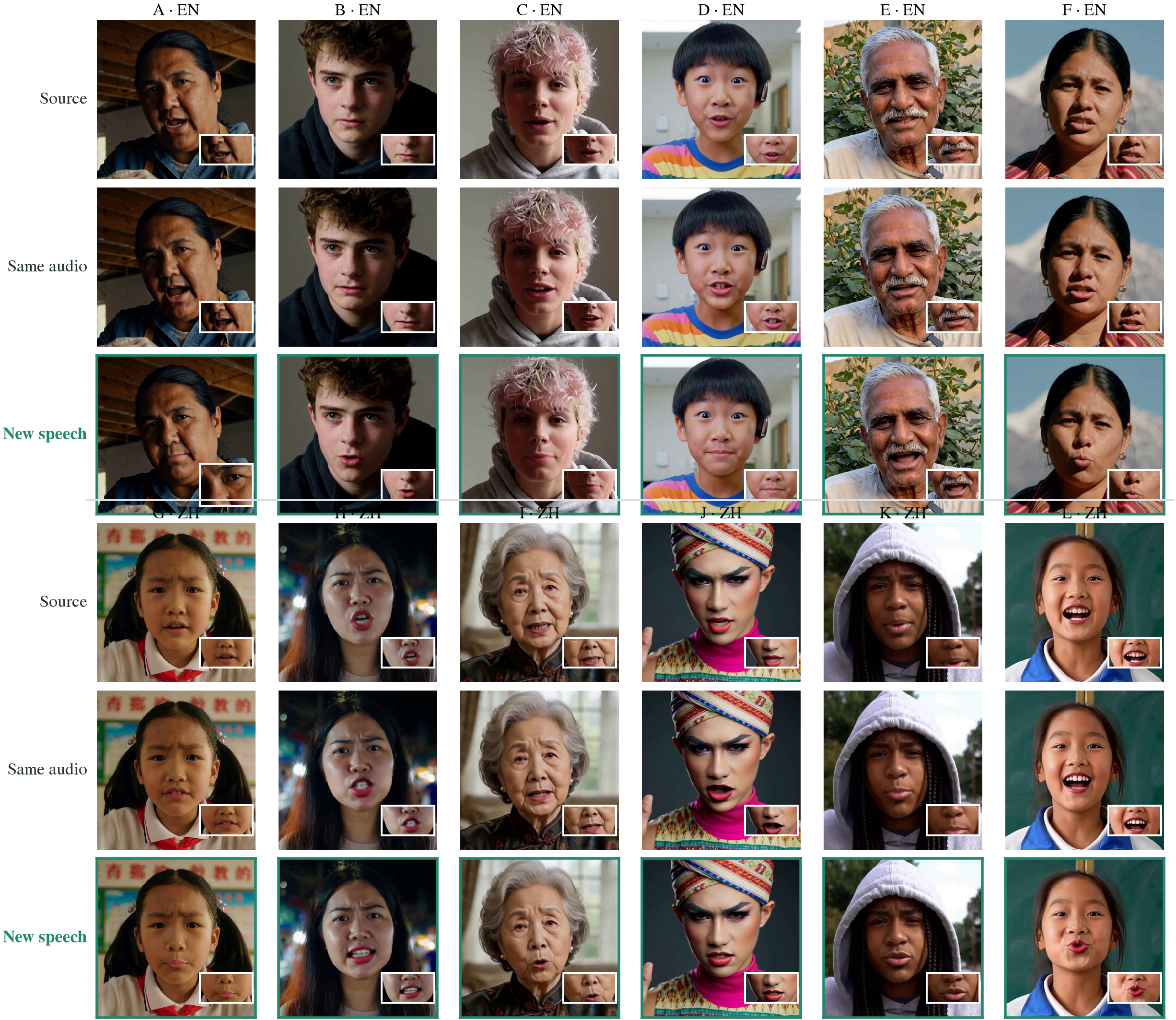}%
}{%
  \fbox{\parbox[c][4.8in][c]{0.96\textwidth}{\centering
  \textbf{Held-out multilingual examples}\par
  The corresponding figure asset is unavailable.}}%
}
\caption{\textbf{Held-out multilingual examples.} Twelve mechanically selected cases
(six English and six Chinese) are shown at matched times. Rows contain the source,
same-audio edit, and replacement-speech edit; insets enlarge the mouth, and green
borders identify replacement-speech outputs.}
\label{fig:supp-diversity}
\end{figure*}

Figure~\ref{fig:supp-diversity} broadens the main-paper teaser with mechanically
selected held-out English and Chinese cases. Each column fixes source and time,
while rows separate the source,
same-audio editing, and replacement-speech editing. Mouth insets expose articulation
that is difficult to inspect at full-frame scale.

\section{Full-Context Teacher Training}
\label{supp:sft}

\subsection{Training chain}

The teacher follows three semantic stages matching the main-paper figure:
\emph{audio-conditioning warm-up} trains only the audio adapter, \emph{joint editing
adaptation} updates DiT, VACE, and the adapter, and \emph{replacement-aware teacher
refinement} adds replacement speech and synchronization supervision behind a DiT
LoRA \cite{lora}. The resulting replacement-aware teacher is used throughout the
paper.
Two independent auxiliary branches study quality-only refinement and replacement
refinement after that quality-only branch; neither is an ancestor of the teacher.
Table~\ref{tab:supp-sft-stages} lists the settings recorded in each run's saved arguments.
``Full'' denotes direct DiT optimization; ``LoRA-64'' denotes a frozen DiT base with
rank-64 adapters merged at export.
The 81-frame training statement in the main method applies to joint adaptation,
replacement-aware refinement, causal adaptation, and distillation; the adapter-only
initialization uses 49-frame clips.

\begin{table*}[t]
\centering
\caption{Full-context teacher training stages. The first three rows form the parent
sequence; the final two rows are independent auxiliary branches. Warm-up and joint
adaptation were run as resumed segments, whose update counts are summed.}
\label{tab:supp-sft-stages}
\small
\resizebox{\textwidth}{!}{%
\begin{tabular}{@{}llcccccccc@{}}
\toprule
Stage & Initialized from & Frames & GPUs & Updates & Adapter LR & DiT & VACE LR & Cross audio & Auxiliary losses \\
\midrule
Audio-conditioning warm-up & scratch adapter & 49 & 48 & 6,000+2,000+9,700 & $10^{-4}\!\rightarrow\!5\!\times\!10^{-5}$ & frozen & frozen & 0 & flow \\
Joint editing adaptation & warm-up & 81 & 48 & 3,000+2,500 & $5\!\times\!10^{-5}$ & full, $10^{-5}$ & $10^{-5}$ & 0 & flow \\
Replacement-aware teacher refinement & joint adaptation & 81 & 16 & 300 & $10^{-5}$ & LoRA-64, $5\!\times\!10^{-5}$ & $2\!\times\!10^{-6}$ & 0.30 & Sync, LPIPS, repr.\ alignment \\
\midrule
\multicolumn{10}{@{}l}{\textit{Auxiliary ablations}} \\
Quality-only refinement & joint adaptation & 81 & 48 & 2,000 & $10^{-5}$ & LoRA-64, $5\!\times\!10^{-5}$ & $2\!\times\!10^{-6}$ & 0 & LPIPS, repr.\ alignment \\
Replacement-after-quality branch & quality-only refinement & 81 & 48 & 300 & $10^{-5}$ & LoRA-64, $5\!\times\!10^{-5}$ & $2\!\times\!10^{-6}$ & 0.30 & Sync, LPIPS, repr.\ alignment \\
\bottomrule
\end{tabular}}
\end{table*}

Teacher refinement and both auxiliary branches use mouth LPIPS weight 0.1 on
same-audio samples and representation alignment at transformer layer 8 with
weight 0.1. The replacement-aware branches add synchronization weight 0.02 for
$\sigma\leq0.8$ and EMA 0.99. The replacement-after-quality branch runs for 300
updates and is evaluated only as an auxiliary ablation; all causal models use the
same replacement-aware teacher. The merged DiT state has the base-model
key layout and requires no LoRA module at inference.

\begin{table*}[t]
\centering
\caption{Empirical teacher-development checkpoints on the same HDTF-30 protocol
(30 same and 30 cross outputs per row). These rows document the effect of the staged
teacher-training recipe and the final teacher choice; because the earlier and final training chains
also differ in accumulated optimization, the table is not presented as a one-factor
causal ablation.}
\label{tab:supp-sft-results}
\small
\IfFileExists{tables/sft_stage_results.tex}{%
  \newcommand{\SftStageRows}{%
Joint editing adaptation & 0.9230 & 7.355 / 8.086 & 5.825 / 9.603 & 33.50 \\
Quality-only refinement, EMA & 0.9233 & 7.561 / 7.845 & 6.104 / 9.256 & 29.56 \\
Replacement-after-quality branch, EMA & 0.9216 & 9.371 / 6.464 & 9.135 / 6.631 & 31.59 \\
Replacement-aware teacher, EMA & 0.9279 & 9.760 / 5.900 & 9.422 / 6.127 & 27.49 \\
}
}{%
  \newcommand{\SftStageRows}{%
  Joint editing adaptation & 0.9230 & 7.355 / 8.086 & 5.825 / 9.603 & 33.50 \\
  Quality-only refinement, EMA & 0.9233 & 7.561 / 7.845 & 6.104 / 9.256 & 29.56 \\
  Replacement-after-quality branch, EMA & 0.9216 & 9.371 / 6.464 & 9.135 / 6.631 & 31.59 \\
  Replacement-aware teacher, EMA & 0.9279 & 9.760 / 5.900 & 9.422 / 6.127 & 27.49 \\
  }%
}
\begin{tabular}{@{}lcccc@{}}
\toprule
Checkpoint & SSIM$\uparrow$ & Same C/D & Cross C/D & Cross FVD$\downarrow$ \\
\midrule
\SftStageRows
\bottomrule
\end{tabular}
\end{table*}

Table~\ref{tab:supp-sft-results} shows why teacher construction matters for the
downstream causal study. Relative to joint editing adaptation, quality-only
refinement lowers cross FVD from 33.50 to 29.56 while slightly increasing cross
Sync-C from 5.825 to 6.104. Applying replacement refinement after that branch raises
cross Sync-C to 9.135 at a small SSIM/FVD cost. Direct replacement-aware teacher
refinement after joint adaptation gives the best joint Sync-C/FVD result
(cross Sync-C 9.422 and FVD 27.49). These staged comparisons have different
accumulated histories and are not single-factor causal effects.

\begin{table*}[t]
\centering
\caption{Matched replacement-sampling control at the shared training endpoint. Both refinements
start from the same joint-adaptation checkpoint and share data, weights, losses,
optimizer, LoRA, EMA, update count, and evaluation pairs. The only configuration
change is replacement probability 0 versus 0.30; when replacement is selected,
the implementation also disables audio dropout. Cells report cross-audio
Sync-C/FVD.}
\label{tab:supp-sft-matched-cross}
\small
\begin{tabular}{@{}lcccc@{}}
\toprule
Teacher refinement & HDTF-30 & TalkVid-Bench-50 & H3-Val-50 & VoxCeleb2-30 \\
\midrule
No replacement branch & 7.986 / 26.40 & 6.460 / 26.14 & 6.952 / 37.68 & 5.729 / 39.95 \\
Replacement-aware (0.30) & 9.422 / 27.49 & 7.414 / 26.51 & 8.086 / 38.09 & 7.382 / 39.11 \\
\bottomrule
\end{tabular}
\end{table*}

Replacement-aware refinement raises cross-audio Sync-C by
1.436/0.954/1.134/1.653 on the four benchmarks. Paired 10,000-resample source
bootstraps give 95\% intervals [1.304, 1.560], [0.863, 1.043],
[0.984, 1.288], and [1.485, 1.823], respectively; all 160 paired items improve.
FVD changes by +1.09/+0.37/+0.40/$-0.84$, so the fidelity direction is mixed.
The matched result identifies a cross-audio synchronization gain with mixed FVD changes.
The intervals are paired source bootstraps over the 160 frozen evaluation items.
TalkVid-Bench-50 is a fixed stratified teacher-development subset of
TalkVid-Bench. H3-Val-50 is a held-out H3 validation subset used during teacher
development. These
teacher-development sets are distinct from both dev-256 and the final
VidDub-Bench-256 evaluation set.

\begin{table}[t]
\centering
\caption{\textbf{Three-seed factorial on dev-256.} Mean effects$\pm$SD separate
source-mouth masking from matched audio dropout and timestep range.}
\label{tab:supp-s3-factorial}
\small
\setlength{\tabcolsep}{1.5pt}
\begin{tabularx}{\columnwidth}{@{}Xrrr@{}}
\toprule
Factor & $\Delta$Sync-C$\uparrow$ & $\Delta$Sync-D$\downarrow$ & $\Delta$FVD$\downarrow$ \\
\midrule
Full objective $-$ paired only & $+1.239_{\pm.012}$ & $-0.951_{\pm.011}$ & $+0.664_{\pm.092}$ \\
\midrule
\oursrow Source-mouth mask (M) & $\mathbf{+0.833_{\pm.040}}$ & $\mathbf{-0.619_{\pm.037}}$ & $-0.230_{\pm.259}$ \\
Audio dropout (D) & $-0.019_{\pm.009}$ & $+0.012_{\pm.008}$ & $-0.044_{\pm.064}$ \\
Timestep range (T) & $-0.206_{\pm.053}$ & $+0.147_{\pm.041}$ & $-0.101_{\pm.056}$ \\
\bottomrule
\end{tabularx}
\end{table}

\subsection{Teacher-refinement ablation protocol}

The teacher study separates historical staged evidence from the matched comparison
in Table~\ref{tab:supp-sft-matched-cross}. Warm-up and joint editing adaptation
measure an association between backbone adaptation and synchronization; the
quality-only branch studies perceptual and representation recovery without a Sync
objective. The matched control then isolates the implemented replacement-sampling
branch by fixing the parent, data, losses, optimizer, and update count. Live versus
EMA export and LoRA versus direct tuning concern optimization stability rather than
objective design.

\begin{table*}[t]
\centering
\caption{Detailed teacher-refinement ablation protocol and interpretation. ``Visual criteria met'' means
SSIM, mouth LPIPS, FVD, and artifact inspection remain within the frozen
tolerance relative to the preceding stage.}
\label{tab:supp-sft-protocol}
\small
\resizebox{\textwidth}{!}{%
\begin{tabular}{@{}lllcl@{}}
\toprule
Comparison & Isolated factor & Primary metrics & Required direction & Interpretation \\
\midrule
Warm-up $\rightarrow$ joint adaptation & backbone/VACE adaptation & same/cross C/D, FVD & cross C up & audio features become usable by the generator \\
Joint adaptation $\rightarrow$ quality-only & LPIPS+repr.\ alignment+LoRA+EMA & SSIM, mouth LPIPS, FVD & visual criteria met & improves visual metrics while reporting synchronization \\
No-replacement $\rightarrow$ replacement-aware refinement & implemented replacement branch & cross C/D, FVD & cross C improves & matched replacement-sampling comparison; dropout semantics disclosed \\
live $\rightarrow$ EMA & parameter averaging & all metrics, failures & lower variance & separates optimization noise from method gain \\
reference/dropout sweep & conditioning strength & cross-audio synchronization, identity & Pareto improvement & tests appearance-prior competition \\
\bottomrule
\end{tabular}}
\end{table*}

\subsection{Audio conditioning and replacement-aware targets}

A frozen Whisper-small encoder \cite{whisper} supplies 13 hidden-state levels. A learned
mixture and projection produce local audio tokens. Each video-latent frame
reads 16 Whisper tokens, approximately 320 ms at 50 Hz, selected by absolute
temporal position. Audio cross-attention follows each of the 30 DiT blocks.
Thus, multi-block causal training gives each video block its corresponding
audio segment rather than a repeated chunk-level embedding.

For ordinary same-audio training, the mouth receives increased flow weight.
For a cross-audio pair $(z,a')$, $a'$ is taken from the preceding sampled clip
on the same training rank; the sampler does not explicitly enforce a different
source video, identity, transcript, temporal non-overlap, or language. All teacher
caches contain 81 frames at 24 fps and audio is read from sample zero through
54,000 at 16 kHz. The swap therefore uses the cached fixed-duration waveform
without a random offset, looping, or runtime resampling. Source-mouth flow
regression is removed:
\begin{equation}
\mathcal L_{\mathrm{flow}}^{\mathrm{cross}}
=\left\|(1-M)\odot\left(v_\theta-v^*\right)\right\|_2^2.
\end{equation}
The non-mouth region remains supervised, audio dropout is disabled, and the
edited mouth is constrained by synchronization. LPIPS and representation-alignment targets tied to
the source mouth are not applied to cross-audio examples. The 0.30 rate
is the setting used for the final teacher.
Teacher refinement draws Gaussian noise and a discrete scheduler timestep uniformly;
replacement examples are restricted to $\sigma\leq0.8$, while same-audio examples
use the full schedule. The scheduler's timestep weight multiplies the normalized
flow error.

\noindent\textbf{Differentiable synchronization objective.}
Training uses the frozen LatentSync-1.6 StableSyncNet checkpoint
\cite{latentsync,latentsync_release}, not the official SyncNet-v2 evaluator
\cite{syncnet,voxceleb2}. A
16-frame, 24-fps decoded window is cropped to the mask-derived
lower face, widened by 10\% horizontally, and resized to $128\times256$. Its
corresponding 16-kHz mel spectrogram has 80 bins at 80 Hz and is resampled to 52
columns. If $e_v,e_a$ are visual and audio embeddings, the loss is
$-\log[\max(\cos(e_v,e_a),10^{-6})]$. Training performs no temporal-lag search.
StableSyncNet parameters remain frozen, but differentiation through the network and
VAE decoder sends the loss to the generator. Final Sync-C/D instead use the official
S3FD \cite{s3fd}/SyncNet-v2 \cite{syncnet,voxceleb2} pipeline described in
Sec.~\ref{supp:evaluation}.

\section{Full-Context Two-Step Distillation}
\label{supp:fullcontext-distillation}

The final student is distilled directly from the replacement-aware
full-context teacher without causal conversion. We cache 4,096 trajectories from
2,048 source clips, with one paired-audio and one fixed same-domain,
same-language, different-source replacement-speech record per source. Teacher
rollouts use 48 Euler steps and audio CFG 2.

Trajectory regression trains the full-context two-step initializer for 400
generator updates. Its pixel/LPIPS/sync/keep coefficients are
$(0.02,0.10,0.02,0.10)$ and its rollout-quality scale is 0.3664846. A 200-update
fake-score warm-up follows with decoded auxiliary losses disabled. Final DMD runs
for 100 generator updates, each accompanied by five fake-score updates; it uses
retention weight 0.10, rollout-quality scale 0.9879127, and
pixel/LPIPS/keep coefficients $(0.02,0.10,0.10)$. The reported checkpoint is
the final DMD student after 100 generator updates; the 50-update checkpoint is
diagnostic only.
These are short adaptation stages rather than training from scratch. Trajectory
regression starts from the mature replacement-aware full-context teacher and uses
frozen matched endpoints; DMD starts from that converged two-step trajectory model
and corrects its generated distribution under a fixed two-step schedule. We freeze
the 400/200/100 update budgets before final evaluation and retain the initializer,
no-guidance, and audio-mixture controls at the same DMD budget. The resulting
evidence supports this bounded distillation recipe, not a general claim that an
arbitrary video diffusion model converges in 100--400 updates.

\begin{figure}[t]
\centering
\includegraphics[width=0.97\columnwidth]{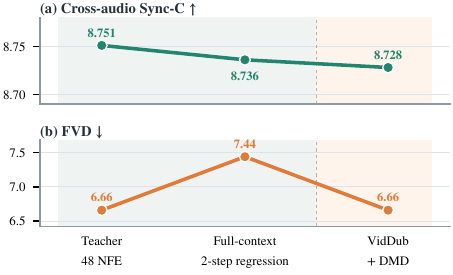}
\caption{\textbf{Full-context distillation on dev-256.} The replacement-aware
teacher, trajectory student, and final DMD student obtain cross-audio
Sync-C/FVD of 8.751/6.66, 8.736/7.44, and 8.728/6.66, respectively.}
\label{fig:supp-transfer-path}
\end{figure}

Figure~\ref{fig:supp-transfer-path} shows that trajectory regression preserves the
teacher's replacement-conditioned response, while DMD recovers the FVD increase
introduced by two-step compression. On the HDTF-30 public benchmark, the
48-step teacher, two-step trajectory student, and final DMD student obtain
cross-audio Sync-C/FVD of 10.162/26.22, 10.113/31.07, and 10.102/27.66,
respectively. This independent benchmark reproduces the preserved synchronization
and partial FVD recovery observed during development. These full-context results
are distinct from the causal diagnostic ablations reported below.

\section{Controlled Causal Variant}
\label{supp:causal-implementation}

This section documents the matched causal control, not the final full-context
VidDub model. Both use Wan-1.3B \cite{wan}, VACE \cite{vace}, and a frozen
Whisper-small encoder \cite{whisper}; the causal control is retained only to isolate
the quality--latency effect of progressive block emission.

\begin{table*}[t]
\centering
\caption{Architecture, sampling, and loss configuration for the three causal stages.
Entries marked ``same'' inherit the causal-adaptation setting. Reported NFE counts
denoising steps only; the \causalvariant additionally performs one clean state-commit
forward per block during inference deployment.}
\label{tab:supp-final-hyperparameters}
\small
\setlength{\tabcolsep}{5pt}
\begin{tabularx}{\textwidth}{@{}>{\raggedright\arraybackslash}p{0.30\textwidth}
  >{\centering\arraybackslash}X>{\centering\arraybackslash}X>{\centering\arraybackslash}X@{}}
\toprule
 & Causal adaptation & Two-step initialization & Guided DMD \\
\midrule
\multicolumn{4}{@{}l}{\textbf{Architecture and sampling}} \\
Backbone & Wan-1.3B + VACE & same & same \\
Audio module & Whisper-small + 287M adapter & same & same \\
Input clip & 81 frames, $512^2$, 24 fps & same & same \\
Latent layout & 1 reference + 21 video latents & same & same \\
Causal blocks & 7 blocks $\times$ 3 latents & same & same \\
Blocks with generator gradient & all 7, sequential backward & 1 sampled trajectory block & 1 sampled rollout block \\
History & clean ground truth & stored causal trajectory & student's causal rollout \\
Precision / checkpointing & bf16 / enabled & bf16 / enabled & bf16 / enabled \\
Reported inference NFE & 48 & 2 & 2 \\
Flow shift & 5 & 5 & 5 \\
Teacher audio guidance & 1 & 2 & 2 \\
Inference audio guidance & 1 & 1 & 1 \\
Replacement-speech sampling & 0.10 online & frozen 1:1 same/cross & 0.50 online \\
Same-audio dropout & 0.10 & 0 & 0 \\
DMD range / fake-score updates & -- & -- & $\sigma\sim\mathcal U[0.02,0.98]$ / 5 \\
\midrule
\multicolumn{4}{@{}l}{\textbf{Losses}} \\
Mouth flow weight & 3 & 3 & 3 \\
Pixel / LPIPS weight & 0.02 / 0.10 & 0.02 / 0.10 & 0.02 / 0.10 \\
Synchronization / non-mouth keep weight & 0.02 / 0.10 & 0.02 / 0.10 & 0 / 0.10 \\
Retention weight & -- & -- & 0.10 \\
Rollout-quality scale & -- & 0.3664846 & 0.9879127 \\
Quality-loss noise range & $\sigma\leq0.5$ & $\sigma\leq1.0$ & $\sigma\leq1.0$ \\
Sync-loss noise range & $\sigma\leq0.8$ & $\sigma\leq0.8$ & -- \\
Quality losses on replacement speech & Sync + non-mouth keep & Sync + non-mouth keep & non-mouth keep only \\
\bottomrule
\end{tabularx}
\end{table*}

\noindent\textbf{Loss coefficients in the main-paper notation.}
We use $\lambda_M=2$, which gives mouth flow weight $1+\lambda_M=3$.
The full-context pixel/LPIPS/sync/keep coefficients are
$(0,0.10,0.02,0)$, and the corresponding causal-adaptation coefficients are
$(0.02,0.10,0.02,0.10)$. For trajectory regression,
$\lambda_q^{\rm traj}=0.3664846$ and
$(\lambda_{\rm pix}^{\rm traj},\lambda_{\rm perc}^{\rm traj},
\lambda_{\rm sync}^{\rm traj},\lambda_{\rm keep}^{\rm traj})
=(0.02,0.10,0.02,0.10)$. For distribution matching,
$\lambda_{\rm ret}=0.10$, $\lambda_q^{\rm DMD}=0.9879127$, and
$(\lambda_{\rm pix}^{\rm DMD},\lambda_{\rm perc}^{\rm DMD},
\lambda_{\rm keep}^{\rm DMD})=(0.02,0.10,0.10)$.
The DMD noise level follows $\sigma\sim\mathcal U[0.02,0.98]$.

% Generated by bench/analyze_talkvid_common_intersection.py.
% Method & success & common Sync-C [95% CI] & common Sync-D [95% CI] & common FVD
\newcommand{\TalkVidCommonRows}{%
Wav2Lip & 499/500 & 6.686 [6.521,6.848] & 7.964 [7.856,8.075] & 112.21 \\
VideoReTalking & 402/500 & 6.361 [6.215,6.509] & 7.859 [7.772,7.953] & 87.80 \\
MuseTalk 1.5 & 498/500 & 4.360 [4.195,4.535] & 9.624 [9.495,9.754] & 36.67 \\
Diff2Lip & 403/500 & 6.904 [6.745,7.067] & 7.021 [6.925,7.116] & 70.21 \\
LatentSync 1.6 & 499/500 & 6.774 [6.611,6.943] & 7.553 [7.450,7.656] & 57.24 \\
X-Dub & 499/500 & 5.095 [4.916,5.275] & 9.391 [9.256,9.523] & 35.21 \\
Lip Forcing & 498/500 & 5.534 [5.352,5.714] & 8.691 [8.545,8.839] & 14.18 \\
Causal two-step VidDub & 500/500 & 7.641 [7.461,7.817] & 7.434 [7.323,7.548] & 14.24 \\
\oursrow\method & 500/500 & 7.571 [7.391,7.753] & 7.340 [7.222,7.459] & 12.06 \\
}

\begin{table*}[t]
\centering
\caption{Optimization and inference configuration. The effective batch equals the
number of workers because each worker processes one clip without gradient
accumulation.}
\label{tab:supp-optimization-settings}
\small
\setlength{\tabcolsep}{5pt}
\begin{tabularx}{\textwidth}{@{}>{\raggedright\arraybackslash}p{0.30\textwidth}
  >{\centering\arraybackslash}X>{\centering\arraybackslash}X>{\centering\arraybackslash}X@{}}
\toprule
 & Causal adaptation & Two-step initialization & Guided DMD \\
\midrule
\multicolumn{4}{@{}l}{\textbf{Optimization}} \\
Optimizer / schedule & AdamW / cosine & AdamW / cosine & AdamW / cosine \\
Adapter or student learning rate & $5\times10^{-6}$ & $2\times10^{-6}$ & $5\times10^{-7}$ \\
Wan/VACE learning rate & $1\times10^{-6}$ & $5\times10^{-7}$ & $2\times10^{-7}$ \\
Fake-score learning rate (adapter / base) & -- & -- & $1\times10^{-6}$ / $5\times10^{-7}$ \\
$(\beta_1,\beta_2)$ & $(0.9,0.999)$ & $(0.9,0.999)$ & $(0.9,0.999)$ \\
Weight decay / grad.\ clip & 0.01 / 1.0 & 0.01 / 1.0 & 0.01 / 1.0 \\
LR schedule / minimum ratio & cosine / 0.1 & cosine / 0.1 & cosine / 0.1 \\
Warmup updates & 50 & 50 & 20 \\
Declared / selected updates & 800 / 400 & 400 / 400 & 100 / 100 \\
GPUs / effective batch & 32 / 32 & 16 / 16 & 16 / 16 \\
Microbatch / grad.\ accumulation & 1 / 1 & 1 / 1 & 1 / 1 \\
Data-loader workers / GPU & 2 & 2 & 2 \\
Random seed & 20261008 & 20261008 & 20261008 \\
\midrule
\multicolumn{4}{@{}l}{\textbf{Inference}} \\
Inference branches & 1 & 1 & 1 \\
Inference NFE & -- & 2 & 2 \\
Persistent state & clean block KV cache & clean block KV cache & bounded clean block KV cache \\
\bottomrule
\end{tabularx}
\end{table*}

The \causalvariant partitions 21 video latents into seven blocks of three after
a separate reference latent. It uses absolute temporal positions,
block-causal self-attention, and frame-local audio cross-attention. Every
target block receives independent noise and timestep samples. Clean history is
detached for standard teacher-forced AR adaptation. To fit all-block training
in memory, blocks are forwarded and backpropagated sequentially, with each loss
divided by the number of selected blocks before one optimizer update.

\noindent\textbf{Flow and pairing details.}
For $u\sim\mathcal U(0,1)$, causal adaptation uses
$\sigma=5u/(1+4u)$, $x_\sigma=(1-\sigma)z+\sigma\epsilon$, and velocity target
$v^*=\epsilon-z$. The VACE mask channels are reduced by a channel maximum and
clamped to obtain latent support $M$. The pointwise squared velocity error is
multiplied by $1+2M$ for same audio and $1-M$ for replacement speech, then divided
by the mean weight. With probability 0.10, the online sampler draws a donor
uniformly from the 16,384 training clips; it replaces an exact clip-ID collision
with the following manifest item but imposes no group, identity, transcript,
language, or silence filter. Under this manifest, the induced same-group and
same-language probabilities are 0.0294\% and 50.40\%, respectively. DMD uses the
same online rule at probability 0.50.

Trajectory generation instead freezes 2,048 source rows and emits both same- and
cross-audio records, yielding 4,096 records in a 1:1 ratio. Each cross donor is
deterministically selected from the same domain and language but a different source
group. Thus the trajectory-regression configuration's online replacement probability zero does not
mean that trajectory regression lacks replacement examples.

\noindent\textbf{Decoded quality terms.}
The complete available latent prefix is decoded and the trailing 16 frames are
scored. Pixel loss is L1 inside the editable support; VGG-LPIPS \cite{lpips} compares the
mask bounding box on four evenly spaced frames; keep loss is L1 outside the support.
Pixel and LPIPS are enabled only for same audio, whereas keep applies to both
branches. Sync applies when audio is present and $\sigma\leq0.8$. Consequently,
causal and trajectory quality are
\begin{align}
\mathcal L_q^s&=.02\mathcal L_{\rm pix}+.10\mathcal L_{\rm LPIPS}
 +.02\mathcal L_{\rm sync}+.10\mathcal L_{\rm keep},\\
\mathcal L_q^c&=.02\mathcal L_{\rm sync}+.10\mathcal L_{\rm keep}^{\bar M}.
\end{align}
Trajectory regression scales either branch by 0.3664846. Its $1+2M$ regression
target is the matched teacher trajectory's terminal latent, so the cross-audio
mouth target is a replacement-conditioned pseudo-target rather than source video.

At inference, conditional and null-audio KV caches are independent when CFG is
requested. A clean $t=0$ pass commits a completed block; no noisy intermediate
is written into persistent history. For long streams, the cache retains the separate
reference and a fixed number of most-recent video latent frames; cached keys preserve
their absolute RoPE positions \cite{rope}, while the absolute generation endpoint continues to
increase. The final distilled model uses only the conditional branch with audio CFG
equal to one. Regression tests verify that the unbounded default is bitwise unchanged,
bounded and unbounded short rollouts agree when the cap covers the sequence, and the
retained KV length stays constant over many causal blocks.

\begin{table}[t]
\centering
\caption{Prepared 81-frame inference efficiency. Medians are computed over ten
timed runs after two warm-up runs. The scope includes condition preparation, causal
denoising, and VAE decoding, but excludes face detection, paste-back, and source
encoding.}
\label{tab:supp-efficiency-breakdown}
\small
\setlength{\tabcolsep}{7pt}
\begin{tabular}{@{}lr@{}}
\toprule
Quantity & Median \\
\midrule
Condition preparation & 0.014 s \\
Time to first block & 0.678 s \\
Denoising & 5.632 s \\
VAE decode & 2.960 s \\
Prepared-clip total & 8.606 s \\
Throughput & 9.412 fps \\
Real-time factor & 2.550 \\
Peak allocated memory & 21.94 GiB \\
Peak reserved memory & 22.53 GiB \\
\bottomrule
\end{tabular}
\end{table}

Long audio uses overlapping Whisper windows blended on one global 50-Hz timeline,
then projects audio tokens once for all absolute latent positions. The Wan VAE exposes
causal streaming encode/decode using the same internal convolution feature state as
full-clip inference. On a real 81-frame clip, streaming and full encode/decode are
bitwise identical (maximum latent and decoded error 0). The frozen long-video set
contains 20 strict contiguous HDTF source/target chains supporting genuine 10, 30,
and 60-second prefixes; video is never looped or concatenated across identities.

\begin{figure*}[t]
\centering
\IfFileExists{figures/streaming_smoke_strip.pdf}{%
  \includegraphics[width=0.96\textwidth]{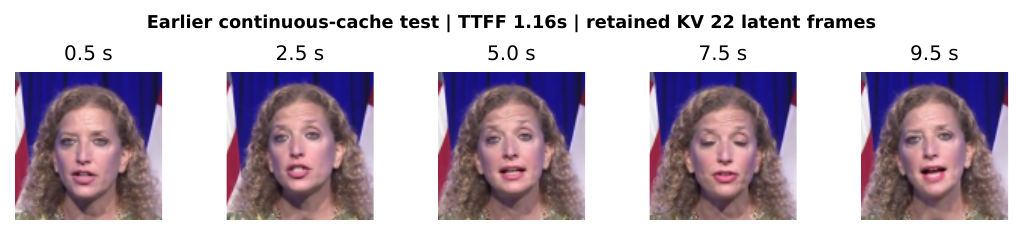}%
}{%
  \fbox{\parbox[c][1.6in][c]{0.96\textwidth}{\centering
    \textbf{Streaming implementation validation}\par
    The corresponding figure asset is unavailable.}}%
}
\caption{\textbf{Streaming implementation validation.} A four-step
checkpoint generates a real continuous 10-second HDTF chain with one persistent
bounded cache, overlapping long-audio Whisper features, and streaming VAE state.
The absolute cache endpoint reaches 65 latent frames while retained KV saturates at
22 (one reference plus 21 video frames). TTFF is 1.16s and peak allocated memory is
23.04GiB. Official SyncNet evaluation in two independent five-second windows gives
Sync-C/D $10.086/6.407$ and $9.719/6.725$, with zero offset in both windows; thus the
persistent cache has zero measured \emph{offset} drift in this interface test.
The later 60-second audit exposes lower-face collapse despite stable synchronization.
The measured end-to-end system RTF is 5.97.}
\label{fig:supp-streaming-smoke}
\end{figure*}

\IfFileExists{tables/final_long_rows.tex}{%
\begin{figure*}[t]
\centering
\includegraphics[width=0.96\textwidth]{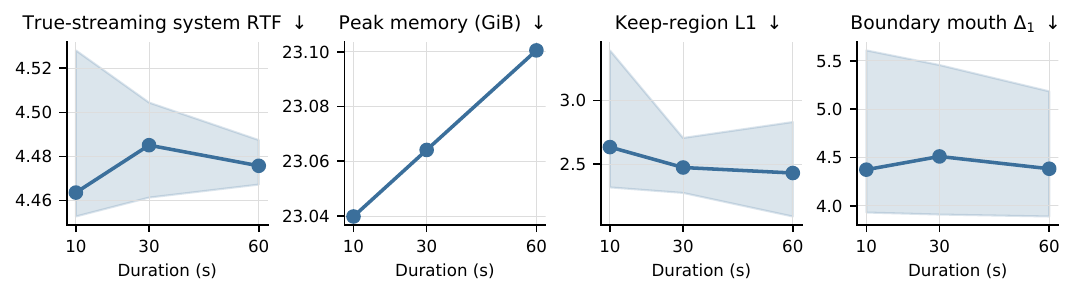}
\input{tables/final_long_caption.tex}
\caption{\FinalLongCaption}
\label{fig:supp-final-long-streaming}
\end{figure*}

\begin{table*}[t]
\centering
\caption{\textbf{Persistent-cache architecture selection.} Values are medians
over 20 contiguous pairs; the combined synchronization, preservation, and visual
measurements motivate the overlapping-window inference used by the final model for stable long-video generation.}
\label{tab:supp-final-long}
\small
\newcommand{\FinalLongRows}{%
10 & 20 & 4.46 & 23.04 & 9.577 / 6.604 & 2.634 & 4.373 & 0 \\
30 & 20 & 4.49 & 23.06 & 9.539 / 6.760 & 2.474 & 4.510 & 0 \\
60 & 20 & 4.48 & 23.10 & 9.577 / 6.814 & 2.430 & 4.383 & 0 \\
}

\resizebox{\textwidth}{!}{%
\begin{tabular}{@{}r r r r c r r r@{}}
\toprule
Duration (s) & Sources & RTF $\downarrow$ & Peak GiB $\downarrow$ & Sync-C / D &
Keep L1 $\downarrow$ & Boundary mouth $\Delta_1\downarrow$ & Sync failures \\
\midrule
\FinalLongRows
\bottomrule
\end{tabular}}
\end{table*}
}{}

\IfFileExists{tables/long_comparison_60s_rows.tex}{%
% Generated by paper/figures/analyze_long_comparison.py.
% Method & NFE & peak GiB [CI] & Sync-C [CI] & Sync-D [CI] & keep L1 [CI] & color [CI] & failures
\newcommand{\LongComparisonRows}{%
Causal AR-48 (CFG 2) & 48 & 29.09 [29.09,29.09] & 9.153 [8.774,9.505] & 6.995 [6.832,7.172] & 2.701 [2.232,3.264] & 0.689 [0.627,0.748] & 0 \\
Two-step trajectory & 2 & 23.27 [23.27,23.27] & 9.019 [8.638,9.376] & 7.130 [6.961,7.313] & 2.930 [2.474,3.478] & 0.867 [0.803,0.931] & 0 \\
\oursrow Guided DMD-2 & 2 & 23.10 [23.10,23.10] & 9.163 [8.773,9.523] & 7.003 [6.831,7.193] & 2.821 [2.358,3.376] & 0.713 [0.659,0.770] & 0 \\
}

\begin{table*}[t]
\centering
\caption{\textbf{Persistent-cache comparison at 60 seconds.}
Brackets are paired 95\% source-bootstrap intervals; synchronization and preserved
region metrics are reported under the same protocol.}
\label{tab:supp-long-comparison}
\small
\resizebox{\textwidth}{!}{%
\begin{tabular}{@{}lrrrrrrr@{}}
\toprule
Method & NFE & Peak GiB & Sync-C$\uparrow$ & Sync-D$\downarrow$ &
Keep L1$\downarrow$ & Color$\downarrow$ & Sync failures \\
\midrule
\LongComparisonRows
\bottomrule
\end{tabular}}
\end{table*}
}{}

\subsection{Extended causal-adaptation ablations}
\label{supp:ar-diagnostics}

The controlled 160-update experiments start from the same teacher checkpoint, use
effective batch 16, and retain the first 160 updates of the same 800-update
cosine schedule. Table~\ref{tab:supp-ar-variants} records each controlled change.

\begin{table*}[t]
\centering
\caption{Controlled causal-adaptation variants. Only the listed factor changes relative to
B3 unless stated otherwise.}
\label{tab:supp-ar-variants}
\small
\resizebox{\textwidth}{!}{%
\begin{tabular}{@{}lllc@{}}
\toprule
ID & Intervention & Diagnostic purpose & Blocks per sample \\
\midrule
A3 & one uniformly sampled block & sparse-coverage reference & 1/7 \\
B3 & sample without replacement & block-coverage sensitivity & 3/7 \\
B7 & supervise every block & complete block coverage & 7/7 \\
S08 & $\sigma_{\rm sync}:0.5\rightarrow0.8$ & noise-range sensitivity & 3/7 \\
SW06 & $\lambda_{\rm sync}:0.02\rightarrow0.06$ & loss-weight sensitivity & 3/7 \\
X30 & S08 plus 30\% counterfactual audio & replacement-speech supervision & 3/7 \\
HM50 & suppress historical mouth with probability 0.5 & history-conditioning sensitivity & 3/7 \\
C2 & two-latent chunks, five sampled blocks & chunk-length sensitivity & 5/10 \\
\bottomrule
\end{tabular}}
\end{table*}

\begin{table*}[t]
\centering
\caption{Complete 160-update controlled results for the causal-adaptation
variants. ``Three-block reference'' is B3; all
rows use the same 16-source protocol. FVD is reported as same/cross.}
\label{tab:supp-ar-diagnostic-results}
\small
\resizebox{\textwidth}{!}{%
\begin{tabular}{@{}lccccccc@{}}
\toprule
Intervention & SSIM$\uparrow$ & Mouth LPIPS$\downarrow$ & Same C/D & Cross C/D & FVD S/C$\downarrow$ & Primary change \\
\midrule
Three-block reference & 0.9310 & 0.1441 & 7.610 / 7.866 & 6.181 / 8.758 & 41.35 / 57.15 & 3/7 blocks \\
Broader Sync exposure & 0.9311 & 0.1439 & 7.981 / 7.582 & 6.807 / 8.232 & 41.36 / 56.20 & $\sigma_{\rm sync}:0.5\!\rightarrow\!0.8$ \\
Larger Sync weight & 0.9310 & 0.1435 & 7.827 / 7.714 & 6.507 / 8.532 & 40.93 / 56.38 & $\lambda_{\rm sync}:0.02\!\rightarrow\!0.06$ \\
Counterfactual speech & 0.9217 & 0.1707 & 8.619 / 7.251 & 7.921 / 7.492 & 54.41 / 63.82 & 30\% cross audio \\
Weaker mouth history & 0.9323 & 0.1382 & 7.539 / 7.917 & 6.227 / 8.712 & 39.79 / 56.00 & 50\% suppression \\
Shorter chunks & 0.9259 & 0.1551 & 7.563 / 7.843 & 6.337 / 8.615 & 45.46 / 57.05 & two latent frames \\
All-block coverage & 0.9325 & 0.1404 & 7.664 / 7.819 & 6.328 / 8.600 & 40.55 / 55.55 & all seven blocks \\
\bottomrule
\end{tabular}}
\end{table*}

With the shifted flow schedule, $\sigma\leq0.5$ occurs much less frequently
than $\sigma\leq0.8$. After excluding an undersized first block, expected AR
synchronization activation is approximately 12.9\%, compared with roughly
41.3\% effective exposure in the replacement-after-quality branch after accounting for cross-audio and audio
dropout. Logged activation rates of 0.10--0.14 agree with this diagnosis.

\subsubsection{Guidance-scale ablation}

For a fixed three-block diagnostic checkpoint at update 160, cross-audio CFG is
varied without retraining. The
monotonic change in Table~\ref{tab:supp-cfg} shows that the output remains
sensitive to the audio-conditioning scale after causal adaptation.

\begin{table}[t]
\centering
\caption{Cross-audio guidance scan on the pilot subset. Mouth $\Delta_1$ is measured
against CFG 1 and is not a reconstruction-error metric.}
\label{tab:supp-cfg}
\small
\begin{tabular}{@{}rcccc@{}}
\toprule
CFG & Sync-C$\uparrow$ & Sync-D$\downarrow$ & $|$offset$|$ & Mouth $\Delta_1$ \\
\midrule
0.0 & 1.878 & 13.260 & 10.19 & 9.04 \\
1.0 & 6.181 & 8.758 & 0.00 & 0.00 \\
1.5 & 7.245 & 7.915 & 0.06 & 3.82 \\
2.0 & 7.607 & 7.638 & 0.07 & 6.03 \\
3.0 & 7.889 & 7.536 & 0.00 & 9.69 \\
\bottomrule
\end{tabular}
\end{table}

\subsubsection{Longer AR and history-policy results}

A0--A4 train for 800 updates. R0--R3 start from the selected
Flow+quality+Sync checkpoint at update 400 and continue for 400
updates. R3 uses a four-step rollout only to form training history; every row
in Table~\ref{tab:supp-ar-full} is evaluated with 50 denoising steps.

\begin{table*}[t]
\centering
\caption{Completed development-set causal-adaptation results. FVD uses the same
I3D16 protocol for every row.}
\label{tab:supp-ar-full}
\small
\resizebox{\textwidth}{!}{%
\begin{tabular}{@{}lcccccc@{}}
\toprule
Method & SSIM$\uparrow$ & Mouth LPIPS$\downarrow$ & Same C$\uparrow$ & Cross C$\uparrow$ & FVD same$\downarrow$ & FVD cross$\downarrow$ \\
\midrule
Full-context teacher & 0.951154 & 0.096476 & 8.505424 & 8.050768 & 6.676098 & 11.701082 \\
A0 Flow & 0.939031 & 0.126039 & 7.628839 & 6.194756 & 8.257593 & 12.564611 \\
A1 Flow+Q & 0.938839 & 0.125288 & 7.631446 & 6.158610 & 8.078579 & 12.433051 \\
A2 Flow+Sync & 0.939122 & 0.125797 & 7.796448 & 6.468818 & 7.990434 & 12.506445 \\
A3 Flow+Q+Sync & 0.938953 & 0.124887 & 7.776736 & 6.450379 & 7.767604 & 12.377647 \\
A4 + representation alignment & 0.938046 & 0.126844 & 7.829486 & 6.560639 & 7.800708 & 12.940533 \\
R0 GT history & 0.939363 & 0.123910 & 7.755741 & 6.390372 & 7.706146 & 12.012290 \\
R1 resample 25\% & 0.939077 & 0.124809 & 7.757215 & 6.386939 & 7.676899 & 12.010987 \\
R2 resample 50\% & 0.938943 & 0.125203 & 7.766012 & 6.404347 & 7.666487 & 11.945611 \\
R3 rollout 25\% & 0.938248 & 0.126147 & 7.716166 & 6.388346 & 7.314490 & 11.618155 \\
\bottomrule
\end{tabular}}
\end{table*}

The full-context teacher gives the best joint result. Among
history-policy continuations, R0 has the strongest reconstruction, A4 has the
highest cross Sync-C, and R3 has the lowest FVD.

\subsection{Controlled causal initialization and DMD ablations}
\label{supp:fewstep}

\subsubsection{Causal consistency initialization}

Our Causal-Forcing++ adaptation \cite{causalforcingpp} is pinned to upstream commit
\texttt{da3ddf1a590f30c9edc568e11c2f7717\allowbreak 0995c31c}. Online causal consistency
distillation uses an adjacent frozen-teacher Euler step followed by an EMA
student prediction on a shifted 48-point grid. All blocks share a sampled exit
step and inference uses $x_0$ re-noising. The optimizer follows the upstream
configuration: AdamW \cite{adamw} with learning rate $2\times10^{-6}$,
$\beta_1=0$, $\beta_2=0.999$, weight decay 0.01, gradient clipping 10, and EMA
0.99. The dubbing-specific auxiliary losses are calibrated relative to the CD gradient and
reported separately from pure CD.

\subsubsection{Causal-consistency initializer selection}

Table~\ref{tab:supp-cd-results} records the 200-update initializer comparison.
The re-noise initializer provides the strongest joint visual and cross-audio operating
point and is therefore selected for the subsequent guidance-transfer stage.

\begin{table*}[t]
\centering
\caption{Pilot-subset four-step consistency results. Pure CD uses only causal
consistency; the auxiliary variant adds calibrated valid GT/Sync losses. The rows compare execution and
quality at the calibrated pilot settings.}
\label{tab:supp-cd-results}
\small
\resizebox{\textwidth}{!}{%
\begin{tabular}{@{}lccccccc@{}}
\toprule
Model & Updates & SSIM$\uparrow$ & Mouth LPIPS$\downarrow$ & Same C$\uparrow$ & Cross C$\uparrow$ & Cross D$\downarrow$ & FVD S/C$\downarrow$ \\
\midrule
CF re-noise initializer & 0 & \textbf{0.94580} & \textbf{0.11099} & 7.923 & \textbf{6.418} & \textbf{8.622} & \textbf{30.82} / \textbf{48.62} \\
Pure CD (online) & 200 & 0.94409 & 0.11232 & 7.930 & 6.395 & 8.612 & 31.68 / 49.41 \\
Pure CD (EMA) & 200 & 0.94496 & 0.11201 & 7.939 & 6.333 & 8.663 & 31.97 / 49.10 \\
CD + auxiliaries (online) & 200 & 0.94379 & 0.11197 & \textbf{7.965} & 6.325 & 8.670 & 31.40 / 49.35 \\
CD + auxiliaries (EMA) & 200 & 0.94538 & 0.11176 & 7.915 & 6.346 & 8.658 & 32.16 / 50.31 \\
\bottomrule
\end{tabular}}
\end{table*}

\subsubsection{Asymmetric DMD implementation}

Our implementation follows DMD2 and Causal Forcing
\cite{dmd2,causalforcing}. It contains a frozen bidirectional
real score, an online bidirectional fake score, five fake-score updates per
generator update by default, and one sampled active block retaining generator
gradients. The objective is evaluated on the selected branch's complete final
rollout (two steps for the final model; four only for the diagnostic branch).
A frozen initializer with paired noise provides an anchor consistency target.

For each fake-score update, $\sigma\sim\mathcal U[0.02,0.98]$ and the online
velocity model regresses $\epsilon-z_{\rm student}$ on a noised, detached student
rollout. The real score is the frozen bidirectional teacher. At the generator
update, both scores parameterize clean estimates
$\hat z_r=x_\sigma-\sigma v_r$ and
$\hat z_f=x_\sigma-\sigma v_f$. If $P_b$ is one on one uniformly sampled latent
block and zero elsewhere, the stop-gradient direction is
\begin{equation}
d_b=\frac{P_b\odot(\hat z_f-\hat z_r)}
{\max(\operatorname{mean}_{P_b}|z-\hat z_r|,10^{-6})}.
\end{equation}
$P_b$ has one channel and broadcasts over the latent channels. The denominator
and score direction are computed without gradients. The generator surrogate is
$\frac12\|z-\operatorname{sg}(z-d_b)\|_{1+2M}^2$.

The initializer-consistency loss compares the selected student block to the frozen two-step initializer's
rollout under identical input noise and schedule:
\begin{equation}
\mathcal L_{\rm ret}=
\|P_b\odot(z_{\rm student}-z_{\rm anchor})\|_{1+2M}^2 .
\end{equation}
Its coefficient is 0.10. DMD quality has
$\mathcal L_{q,\rm DMD}^s=.02\mathcal L_{\rm pix}
+.10\mathcal L_{\rm LPIPS}+.10\mathcal L_{\rm keep}$ and
$\mathcal L_{q,\rm DMD}^c=.10\mathcal L_{\rm keep}^{\bar M}$, both scaled by
0.9879127. Thus replacement DMD disables source-mouth pixel/LPIPS and uses no
training Sync loss; it retains only valid non-mouth reconstruction.

During training, only the real score uses conditional and null-audio forwards:
\begin{equation}
s_r^{(g)}=s_r(x_t,\varnothing)+g
\left[s_r(x_t,a)-s_r(x_t,\varnothing)\right].
\end{equation}
The student rollout uses one conditional forward per denoising step. The fake score learns
the student distribution rather than guided teacher outputs. Every reported
DMD model is evaluated with inference audio CFG equal to one.

\begin{table*}[t]
\centering
\caption{Four-step guidance absorption and the unified two/four-step study.
All DMD models use real-teacher guidance only during training and one inference
branch. Candidates use the same stage-specific parents and shared fake-score
initialization within each step-count branch for controlled comparison.}
\label{tab:supp-dmd-sweep}
\small
\begin{tabular}{@{}lccccccc@{}}
\toprule
Study & NFE & Real $g$ & Branches & SSIM & Mouth LPIPS & Cross C/D & FVD S/C \\
\midrule
AR parent & 50 & -- & 1 & 0.9388 & 0.1261 & 7.951 / 7.603 & 9.19 / 13.99 \\
Four-step trajectory & 4 & -- & 1 & 0.9326 & 0.1403 & 8.033 / 7.562 & 10.38 / 14.45 \\
Four-step DMD & 4 & 2.0 & 1 & 0.9377 & 0.1288 & 8.547 / 7.264 & 9.13 / 14.30 \\
Unified trajectory/DMD & 2, 4 & 2.0 & 1 & \multicolumn{4}{c}{selected under common quality, sync, and failure criteria} \\
\bottomrule
\end{tabular}
\end{table*}

\IfFileExists{tables/trajectory_guidance_control_rows.tex}{%
% Generated by paper/analyze_trajectory_guidance_control.py.
\newcommand{\TrajectoryGuidanceControlRows}{%
Unguided trajectories & 1 & 8.302 [8.159,8.449] & 7.135 [7.035,7.236] & 1.996 [1.952,2.042] & 10.52 \\
Guided trajectories & 2 & 8.641 [8.499,8.788] & 7.069 [6.971,7.168] & 1.974 [1.931,2.020] & 11.48 \\
}

\begin{table*}[t]
\centering
\caption{Matched trajectory-guidance control on the 256-source development
manifest. Only teacher trajectory CFG changes; both two-step trajectory students use one
conditional inference branch with CFG 1. Brackets are 95\% paired
source-bootstrap intervals; Sync uses the 255-source common valid set. FVD is an
aggregate point estimate.}
\label{tab:supp-trajectory-guidance-control}
\small
\resizebox{\textwidth}{!}{%
\begin{tabular}{@{}lrrrrr@{}}
\toprule
Trajectory teacher & Teacher CFG & Sync-C$\uparrow$ & Sync-D$\downarrow$ &
Keep L1$\downarrow$ & FVD$\downarrow$ \\
\midrule
\TrajectoryGuidanceControlRows
\bottomrule
\end{tabular}}
\end{table*}
}{}

\begin{table}[t]
\centering
\caption{Exact values for the sequential causal path. Replacement-speech adaptation
precedes two-branch guidance. This causal autoregressive protocol is distinct from
the full-context two-step trajectory initializer in the main-paper ablation.}
\label{tab:supp-sequential-causal}
\small
\setlength{\tabcolsep}{2.2pt}
\begin{tabularx}{\columnwidth}{@{}Xrrrr@{}}
\toprule
Stage & Br. & SSIM$\uparrow$ & Sync-C$\uparrow$ & FVD$\downarrow$ \\
\midrule
Teacher & 2 & 0.9476 & 8.751 & 6.66 \\
Direct causalization & 1 & 0.9387 & 7.993 & 8.21 \\
\quad + replacement speech & 1 & 0.9390 & 8.121 & 8.33 \\
\quad + guidance & 2 & 0.9369 & 8.726 & 8.91 \\
\bottomrule
\end{tabularx}
\end{table}

\IfFileExists{tables/paper_dmd_ablation_rows.tex}{%
\newcommand{\PaperDmdAblationNarrative}{Relative to an unguided real score ($g=1$), the complete $g=2$ model raises cross Sync-C by 0.3503 (95\% CI [0.3217,0.3785]), while changing same-audio SSIM by -0.0014 ([-0.0016,-0.0013]) and mouth LPIPS by +0.0029 ([0.0026,0.0033]). Thus real-score guidance strengthens cross-audio control at a small same-audio visual cost. Disabling initializer retention changes cross Sync-C by +0.0053 ([-0.0078,0.0182]), which is unresolved; its same-audio SSIM change is only -0.00002 ([-0.00004,-0.00001]). Removing replacement batches during DMD only leaves cross Sync-C unresolved (+0.0052, [-0.0089,0.0198]) and improves cross Sync-D by 0.0158; it improves same-audio SSIM by 0.0011 ([0.0010,0.0012]) and mouth LPIPS by 0.0030, with a keep-L1 cost of 0.0051. The frozen ODE initializer still uses paired and cross-audio trajectories. FVD values remain aggregate point estimates and are not assigned per-video paired confidence intervals.}

\begin{table*}[t]
\centering
\caption{Ablation of the final full-context two-step DMD model under the 256-source
development protocol. Guidance, initializer retention, and the DMD batch mixture
each change one field from the final full-context DMD student. The frozen trajectory
initializer was trained with paired and cross-audio trajectories. SSIM and mouth LPIPS are
same-audio metrics; cross-audio reporting uses synchronization and FVD. Inference
uses one conditional forward per denoising step with audio CFG 1.}
\label{tab:supp-dmd-attribution}
\small
\newcommand{\PaperDmdAblationRows}{%
Complete full-context DMD & 2.0 & 0.1 & 0.9467 & 0.1159 & 8.728 / 6.910 & 4.96 / 6.66 \\
Unguided real score ($g=1$) & 1.0 & 0.1 & 0.9481 & 0.1130 & 8.378 / 6.944 & 4.41 / 6.75 \\
No initializer retention & 2.0 & 0 & 0.9467 & 0.1159 & 8.733 / 6.907 & 4.99 / 6.65 \\
No replacement batches during DMD & 2.0 & 0.1 & 0.9478 & 0.1129 & 8.733 / 6.894 & 4.87 / 6.56 \\
}

\resizebox{\textwidth}{!}{%
\begin{tabular}{@{}lcccccc@{}}
\toprule
Variant & Real $g$ & Init.\ retention & Same SSIM$\uparrow$ & Same mouth LPIPS$\downarrow$ & Cross C/D & FVD S/C$\downarrow$ \\
\midrule
\PaperDmdAblationRows
\bottomrule
\end{tabular}
}
\end{table*}
\PaperDmdAblationNarrative
}{}

\subsection{Matched guidance factorial and target-audio response}
\label{supp:decisive-review}

Table~\ref{tab:supp-guidance-factorial} completes the predeclared
trajectory-guidance by DMD-guidance factorial. All four cells use the same
teacher, causal parent architecture, training rows, update schedule, two-step
inference, evaluator, and retention weight; the CFG-1 trajectory branch uses its
own trajectory-regression parent and parent-matched fake score. Both guidance locations improve
Sync-C. DMD guidance trades this gain for higher Sync-D and FVD, whereas
trajectory guidance also improves Sync-D. The small Sync-C interaction
($0.014$) indicates approximately additive effects under this training configuration.

\begin{table*}[t]
\centering
\caption{Complete trajectory-guidance $\times$ DMD-guidance factorial at generator
update 50 on the frozen 256-source development set. Each cell scores 512/512
same/cross outputs; inference always uses one audio branch with CFG 1.}
\label{tab:supp-guidance-factorial}
\small
\begin{tabular}{@{}lrrrrrr@{}}
\toprule
Cell & $g_{\rm traj}$ & $g_{\rm DMD}$ & Retention &
Cross Sync-C$\uparrow$ & Cross Sync-D$\downarrow$ & Cross FVD$\downarrow$ \\
\midrule
F11 & 1 & 1 & 0.1 & 8.371 & 7.070 & \textbf{7.21} \\
F12 & 1 & 2 & 0.1 & 8.578 & 7.120 & 7.66 \\
F21 & 2 & 1 & 0.1 & 8.512 & \textbf{6.954} & 7.26 \\
F22 (complete) & 2 & 2 & 0.1 & \textbf{8.732} & 7.023 & 7.74 \\
\bottomrule
\end{tabular}
\end{table*}

For optimization variability, Table~\ref{tab:supp-seeds} compares the complete
recipe with a matched closest-recipe control trained in our codebase: F11 with retention
disabled. This control is not claimed to reproduce the public Lip Forcing
implementation. Each seed is evaluated on the same source-disjoint 64-source
subset. Source-bootstrap uncertainty and training-seed variation are distinct;
the table reports the latter directly.

\begin{table}[t]
\centering
\caption{Three-seed endpoint variability at update 50 (mean$\pm$sample SD);
128/128 same/cross outputs score for every seed and condition.}
\label{tab:supp-seeds}
\scriptsize
\setlength{\tabcolsep}{3pt}
\begin{tabular}{@{}lrrr@{}}
\toprule
Endpoint & Cross Sync-C$\uparrow$ & Cross Sync-D$\downarrow$ & Cross FVD$\downarrow$ \\
\midrule
Matched control P1 & $8.254\pm0.012$ & $7.176\pm0.008$ & $\mathbf{20.79\pm0.05}$ \\
Complete P4 & $\mathbf{8.593\pm0.018}$ & $\mathbf{7.145\pm0.017}$ & $22.66\pm0.20$ \\
\bottomrule
\end{tabular}
\end{table}

Table~\ref{tab:supp-audio-intervention} fixes source video, editable mask,
reference, and noise while changing only target audio. ``Unrelated'' is a
deterministic derangement of the replacement utterances; ``null T/S'' remuxes
the learned-null output with the designated target/source waveform for
evaluation. Correct and unrelated speech retain high synchronization with the
waveform that actually drove generation. Learned-null outputs instead align
more with source than target, rejecting a generic mouth-motion explanation.
Keep-region SSIM remains nearly unchanged, while the complete model's FVD cost
relative to the matched control remains explicit.

\begin{table*}[t]
\centering
\caption{Fixed-noise target-audio intervention on 32 development sources.
T/S denote target/source-audio remuxes of identical learned-null pixels under each controlled intervention.}
\label{tab:supp-audio-intervention}
\small
\begin{tabular}{@{}lrrrrrrr@{}}
\toprule
Model & Correct C$\uparrow$ & Shifted C$\uparrow$ & Unrelated C$\uparrow$ &
Null T C$\uparrow$ & Null S C$\uparrow$ & Keep SSIM$\uparrow$ & Correct FVD$\downarrow$ \\
\midrule
48-step teacher & 8.112 & 7.584 & 8.032 & 1.813 & 3.499 & \textbf{0.97994} & \textbf{28.96} \\
Matched control P1 & 8.262 & 7.706 & 8.236 & 1.606 & 3.168 & 0.97837 & 35.41 \\
F21 $(2,1)$ & 8.431 & 7.916 & 8.370 & \textbf{1.583} & 3.140 & 0.97848 & 35.38 \\
F22 $(2,2)$ & \textbf{8.625} & \textbf{8.183} & \textbf{8.592} & 1.639 & 3.281 & 0.97845 & 39.83 \\
\bottomrule
\end{tabular}
\end{table*}

\subsubsection{Controlled AR--trajectory--DMD evaluation protocol}

The causal parent is frozen after development-set selection. All two- and four-step
trajectory or consistency candidates share this checkpoint, and each distribution-
matching branch starts from one selected initializer with a branch-specific shared
fake score. Table
\ref{tab:supp-stage-gates} records the ablations and stopping rules fixed before final
evaluation.

\begin{table*}[t]
\centering
\caption{Stage-gated causal conversion experiments. Each row changes one
factor relative to the selected parent for that stage.}
\label{tab:supp-stage-gates}
\small
\resizebox{\textwidth}{!}{%
\begin{tabular}{@{}llll@{}}
\toprule
Stage & Variants & Question & Selection criterion \\
\midrule
AR coverage & A3, B3, B7 & Does supervision cover temporal/audio positions? & better mouth LPIPS and FVD without cross-sync loss \\
AR audio & B7, +S08, +cross 5/10\%, warmup & Can audio evidence increase without X30 artifacts? & cross C/D improve; visual tolerance passes \\
AR history & GT, suppression, resample, audio-consistent rollout & How does history construction affect causal generation? & long-horizon and cross metrics improve jointly \\
Initializer & Euler4, re-noise4, trajectory, CD0, CD1 & Which four-step objective retains the AR parent? & joint development quality/sync Pareto frontier \\
DMD guidance & real $g=1,1.5,2$ & Can teacher CFG be absorbed into one branch? & cross C/D improve at inference CFG 1 \\
DMD initializer consistency & none, trajectory anchor, +quality, +cross keep/Sync & What prevents color and mouth artifacts? & FVD/LPIPS/color improve without synchronization loss \\
\bottomrule
\end{tabular}}
\end{table*}

\begin{figure*}[t]
\centering
\IfFileExists{figures/benchmark_pareto.pdf}{%
  \includegraphics[width=\textwidth]{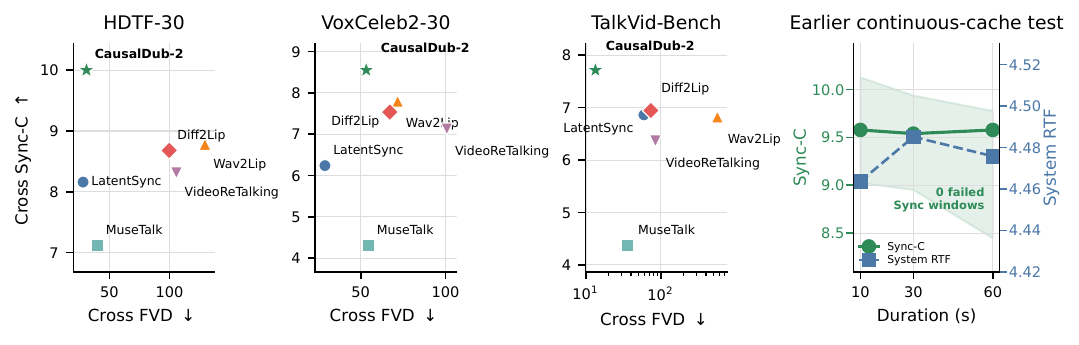}%
}{%
  \fbox{\parbox[c][1.9in][c]{0.96\textwidth}{\centering
    \textbf{Synchronization--quality Pareto figure}\par
    The corresponding figure asset is unavailable.}}%
}
\caption{The final two-step model improves audio--visual synchronization, while the
bidirectional models define the reported FVD frontier. The 10/30/60-second
persistent-cache comparison motivates the overlapping-window inference
used by the final model for stable long-video generation.}
\label{fig:supp-fewstep-pareto}
\end{figure*}

\section{Evaluation Protocol}
\label{supp:evaluation}

\noindent\textbf{Synchronization.} We use the official S3FD \cite{s3fd}/SyncNet-v2
\cite{syncnet,voxceleb2} path and
report Sync-C, Sync-D, estimated offset, tracking failures, and the number of outputs
used by each mean. Same-audio and cross-audio scores are never pooled.
Generated videos can obtain higher Sync-C than real same-audio video because the
metric rewards evaluator-aligned, sometimes exaggerated mouth motion. Training and
evaluation also use different frozen synchronizers. We therefore interpret Sync-C
as a relative synchronization diagnostic, not a calibrated measure of naturalness
or phoneme correctness.

\noindent\textbf{Temporal-shift calibration.}
For 40 VidDub-Bench cross-audio outputs per method, we shift the waveform by
$k\in\{0,\pm1,\pm2,\pm3,\pm5\}$ video frames and rerun the same evaluator. The
estimated offset changes by exactly $-k$ throughout $|k|\leq3$ for every method.
At zero shift, the \causalvariant, its teacher, and LatentSync return offset zero on all
40 clips. This is the temporal-shift test summarized in the main paper.

\noindent\textbf{Independent visual speech recognition.}
Driving utterances are transcribed with Whisper-large-v3 \cite{whisper}, while generated video is
read by the visual-only LRS3 model of Ma et al.~\cite{vsr_ma}; this reader shares no
weights or objective with either the frozen training synchronizer or SyncNet-v2.
The final benchmark uses the 30/28/121 English HDTF/VoxCeleb2/VidDub-Bench clips
reported in the main paper. The three-seed factorial uses the 212 clips valid across
all 27 trained runs/checkpoints (24 factorial runs plus three paired-only anchors):
masking source-mouth reconstruction under unpaired replacement speech lowers
WER by $0.144\pm0.009$ and CER by $0.116\pm0.007$, with the favorable direction
preserved for every seed and all 12 controlled cell comparisons.
External cross-audio WERs near 1.0 and the 0.711 VidDub-Bench real-video control
limit absolute interpretation; these results support relative content attribution
only.

\begin{table*}[t]
\centering
\caption{\textbf{Same-audio sanity check for the visual-only reader.}
Each entry reports same/cross-audio WER on HDTF, VoxCeleb2, and VidDub-Bench,
respectively. All displayed cohorts have zero reader failures. Lower WER for the
same-audio reconstruction confirms that the near-one cross-audio values do not arise
from a universally failed crop or reader invocation; the remaining high same-audio
values caution against treating this reader as calibrated phoneme accuracy.}
\label{tab:supp-vsr-same-sanity}
\small
\setlength{\tabcolsep}{4.0pt}
\begin{tabular}{@{}lccc@{}}
\toprule
Method & HDTF S/C & VoxCeleb2 S/C & VidDub-Bench S/C \\
\midrule
Wav2Lip & 0.879 / 1.093 & 0.893 / 1.009 & 0.919 / 1.020 \\
VideoReTalking & 0.755 / 0.915 & 0.998 / 1.056 & 0.846 / 0.975 \\
Diff2Lip & 0.731 / 0.957 & 0.850 / 0.963 & 0.746 / 0.990 \\
MuseTalk & 0.504 / 0.987 & 0.970 / 1.068 & 0.890 / 1.014 \\
LatentSync & 0.470 / 1.019 & 0.881 / 1.049 & 0.712 / 0.953 \\
X-Dub & 0.839 / 0.952 & 0.932 / 0.949 & 0.890 / 0.949 \\
Lip Forcing & 0.816 / 1.093 & 0.937 / 0.977 & 0.890 / 0.977 \\
\bottomrule
\end{tabular}
\end{table*}

\noindent\textbf{Reconstruction and preservation.} SSIM \cite{ssim}, PSNR, mouth
LPIPS \cite{lpips}, mouth
L1, and outside-mask preservation are valid for same-audio reconstruction. For
cross audio we do not compare the generated mouth with the source mouth. FVD
\cite{fvd} with I3D features \cite{i3d} and FID \cite{fid} are reported for both
conditions using identical feature extraction
and window grouping within each table.
HDTF and VoxCeleb2 FVD use only 30 sources and are treated as small-sample
diagnostics; TalkVid-Bench and VidDub-Bench provide the larger-set distribution
comparisons.

For cross-audio source preservation, corresponding source frames remain valid ground
truth outside the editable lower face. We derive one stable source face crop per frame
and use MediaPipe Face Mesh \cite{mediapipeface} source landmarks to set a
horizontal boundary eight pixels above the
topmost lip landmark. Upper-L1 is the mean RGB error above this boundary, thereby
excluding the full lower face and paste-back blend region. ArcFace cosine
\cite{arcface} and InsightFace 3D-landmark pitch/yaw/roll error \cite{insightface}
are evaluated every fourth frame using the
same source-derived crops. All methods share frame indices, crops, and source targets;
confidence intervals bootstrap source identities.

\begin{table*}[t]
\centering
\caption{Cross-audio source preservation on the three frozen evaluation sets.
Upper-L1 is reported in percentage points; lower Upper-L1 and pose error and higher
identity cosine are better. LatentSync has 254/256 VidDub-Bench outputs; all other
requested outputs are present for evaluation.}
\label{tab:supp-preservation}
\IfFileExists{tables/preservation_benchmark_rows.tex}{\newcommand{\PreservationBenchmarkRows}{%
HDTF / Wav2Lip & 2.580 & 0.928 & 0.432 & 30 \\
HDTF / Diff2Lip & 1.753 & 0.888 & 0.464 & 30 \\
HDTF / LatentSync & 1.646 & 0.970 & 0.228 & 30 \\
HDTF / Lip Forcing & 1.669 & 0.934 & 0.340 & 30 \\
HDTF / Teacher & 0.873 & 0.911 & 0.415 & 30 \\
HDTF / Full-context 2-step & 0.914 & 0.902 & 0.449 & 30 \\
VOX2 / Wav2Lip & 1.697 & 0.913 & 0.573 & 30 \\
VOX2 / Diff2Lip & 1.065 & 0.893 & 0.619 & 30 \\
VOX2 / LatentSync & 1.009 & 0.951 & 0.560 & 30 \\
VOX2 / Lip Forcing & 0.995 & 0.905 & 0.553 & 30 \\
VOX2 / Teacher & 0.912 & 0.877 & 0.629 & 30 \\
VOX2 / Full-context 2-step & 0.931 & 0.877 & 0.671 & 30 \\
VidDub-Bench / Wav2Lip & 1.820 & 0.909 & 0.459 & 256 \\
VidDub-Bench / Diff2Lip & 1.283 & 0.862 & 0.518 & 256 \\
VidDub-Bench / LatentSync & 1.605 & 0.913 & 0.422 & 254 \\
VidDub-Bench / Lip Forcing & 3.761 & 0.900 & 0.928 & 256 \\
VidDub-Bench / Teacher & 1.207 & 0.872 & 0.573 & 256 \\
VidDub-Bench / Full-context 2-step & 1.199 & 0.875 & 0.579 & 256 \\
}
}{}
\small
\setlength{\tabcolsep}{5pt}
\begin{tabular}{@{}lrrrr@{}}
\toprule
Benchmark / method & Upper-L1$\downarrow$ & ID cosine$\uparrow$ & Pose error ($^\circ$)$\downarrow$ & $n$ \\
\midrule
\PreservationBenchmarkRows
\bottomrule
\end{tabular}
\end{table*}

\noindent\textbf{Temporal behavior and efficiency.} We measure mouth temporal
variation, block-boundary variation, stable-face color drift, denoising and
end-to-end latency, time to first block, peak memory, throughput, and real-time
factor. Long-video tests use 10, 30, and 60-second inputs and report cache
growth, synchronization offset, and identity drift over time.

\noindent\textbf{Compute accounting.}
As in the main paper, NFE counts denoising steps only. The \causalvariant additionally
performs one clean state-commit forward per block; timing tables include that forward
unless a prepared-clip scope is explicitly stated.

\begin{figure}[t]
\centering
\includegraphics[width=0.92\columnwidth]{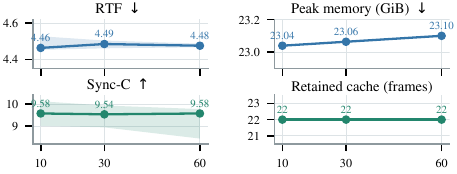}
\caption{\textbf{Long-video execution design.} The persistent-cache run motivates
overlapping windows conditioned on source frames by jointly comparing runtime, memory, synchronization,
and lower-face stability over 10--60 seconds.}
\label{fig:supp-long-horizon}
\end{figure}

\noindent\textbf{Matched architecture timing.}
Table~\ref{tab:supp-causal-design} reports the controlled full-context/causal
comparison, timed after conditioning is prepared. Figure~1(c) of the main paper
separately compares public pipelines with one protocol: each pipeline is timed from
raw video and audio to the final video file on the same five HDTF sources at 4\,s
and 8\,s, and throughput is 25 divided by the fitted seconds per video second.

\noindent\textbf{Statistics.} Per-video comparisons are paired by source using
10,000 bootstrap resamples. Confidence intervals bootstrap source identities. For FVD, all temporal
feature windows from one source are resampled as a group. The pilot subset is diagnostic only; VidDub-Bench is evaluated once after
the selection recipe is frozen.

\section{Long-20 External Comparison}
\label{supp:long20-external}

All methods receive the same 20 frozen 60-second CFR source videos and replacement
waveforms. Wav2Lip uses the official GAN checkpoint; LatentSync 1.6 uses its official
20-step, guidance-1.5 inference. \method uses 81-frame overlapping windows conditioned on source frames,
a 64-frame stride, and 17-frame overlap rather than one persistent rollout. We require
full video/audio decode, score twelve
non-overlapping 5-second SyncNet windows, and count failures rather than dropping
them. For source preservation, full-frame outputs are transformed by the exact
source-derived face crop before corresponding-frame comparison. Keep-L1 excludes the
editable lower face; stable-face chroma is CIELAB $ab$ displacement
\cite{cielab}. Horizon changes
use complete 5-second bins, excluding codec-induced one-frame terminal bins.

\begin{table*}[t]
\centering
\caption{Long-20 results (means over 20 paired sequences). First/last are
5-second Sync-C windows; slope is Sync-C change per minute. $\Delta$ metrics are
last-minus-first complete-bin changes. Runtime is omitted because public pipeline
timing scopes are not uniform across methods in this comparison.}
\label{tab:supp-long20-external}
\small
\setlength{\tabcolsep}{3.0pt}
\resizebox{\textwidth}{!}{%
\begin{tabular}{@{}lrrrrrrrrrr@{}}
\toprule
Method & Valid & Sync-C$\uparrow$ & Sync-D$\downarrow$ & First C & Last C &
Slope/min & Keep-L1$\downarrow$ & Chroma$\downarrow$ &
$\Delta$Keep & $\Delta$Chroma \\
\midrule
Wav2Lip~\cite{wav2lip} & 20/20 & 7.861 & 7.484 & 8.147 & 8.032 & $-0.146$ &
3.045 & 1.045 & +0.009 & +0.003 \\
LatentSync 1.6~\cite{latentsync} & 20/20 & 7.771 & 7.083 & 8.029 & 7.844 & $-0.112$ &
4.018 & 0.397 & +0.685 & +0.006 \\
\oursrow\method & 20/20 & \textbf{9.399} & \textbf{6.162} & \textbf{9.755} &
\textbf{9.585} & $-0.046$ & \textbf{2.493} & \textbf{0.250} &
$-0.034$ & +0.006 \\
\bottomrule
\end{tabular}}
\end{table*}

\begin{table}[t]
\centering
\caption{\textbf{Source re-anchoring.} Pair-14 compares persistent execution with
81-frame source re-anchoring and 17-frame overlap.}
\label{tab:supp-long-reanchor}
\small
\setlength{\tabcolsep}{3.0pt}
\begin{tabularx}{\columnwidth}{@{}Xrr@{}}
\toprule
Metric & Persistent & Re-anchored \\
\midrule
Sync-C $\uparrow$ & 9.578 & \textbf{9.747} \\
Sync-D $\downarrow$ & 6.772 & \textbf{6.106} \\
Peak memory (GiB) $\downarrow$ & 23.10 & \textbf{22.38} \\
Skin chroma MAE $\downarrow$ & 7.48 & \textbf{2.43} \\
Lower-face stability & no & yes \\
\bottomrule
\end{tabularx}
\end{table}

\method completes all 20 sequences with the highest mean and final synchronization.
Its near-zero slope, preserved-region change, and stable-face chroma show that the
overlapping full-context windows maintain both speech synchronization and surrounding
visual evidence throughout the minute.

\section{Benchmark Suite and Baseline Protocol}
\label{supp:benchmarks}

\begin{table*}[t]
\centering
\caption{Benchmark construction and \method execution. Duration/FPS describes the
native source clip scored by each benchmark, not the model-call tensor. Every
\method call uses an 81-frame, 24-FPS, $512{\times}512$ window; longer source
clips use shifted-tail windows, and Long-20 uses 17-frame overlap.}
\label{tab:supp-benchmarks}
\small
\resizebox{\textwidth}{!}{%
\begin{tabular}{@{}lcccll@{}}
\toprule
Benchmark & Sources & Source duration / native FPS & Tracks & Cross pairing & \method execution \\
\midrule
HDTF-30 & 30 & 8.0s / 25 & same + cross & fixed cyclic, different speaker & 81-frame windows, shifted tail \\
VoxCeleb2-30 & 30 & 4.0s / 25 & same + cross & fixed cyclic, different speaker & 81-frame windows, shifted tail \\
TalkVid-Bench & 500 & 5.042s / 24 & same + cross & deterministic within stratum & 81-frame windows, shifted tail \\
VidDub-Bench & 256 & 3.375s / 24 & same + cross & fixed multilingual pairs & one 81-frame full-context window \\
Frozen-50 & 50 & 3.375s / 24 & cross & identical clips for compared methods & matched blinded comparison \\
Long-20 & 20 & 60s / 25 & cross primary & frozen contiguous pairs & 81-frame windows, 17-frame overlap \\
\bottomrule
\end{tabular}}
\end{table*}

\IfFileExists{tables/lipforcing_rows.tex}{\providecommand{\LipForcingHdtfRow}{Lip Forcing & \ParamsLipForcing & 2 & 8.429 & 6.919 & 0.9517 & 31.37 & 6.884 & 8.284 & 33.80 \\}
\providecommand{\LipForcingVoxRow}{Lip Forcing & \ParamsLipForcing & 2 & 6.408 & 7.831 & 0.9466 & 33.12 & 4.871 & 9.242 & 41.03 \\}
\providecommand{\LipForcingTalkRow}{Lip Forcing & \ParamsLipForcing & 2 & 5.914 & 8.283 & 0.7698 & 24.48 & 5.610 & 8.648 & 12.68 \\}
\providecommand{\LipForcingPublicSummaryRow}{Lip Forcing & \ParamsLipForcing & 2 & 6.884 / 33.80 & 4.871 / 41.03 & 5.610 / 12.68 \\}
}{}

Table~\ref{tab:vox2} gives the complete VoxCeleb2-30 comparison that is not
tabulated in the main paper. Table~\ref{tab:talkvid} reproduces the complete
main-paper TalkVid-Bench comparison for the supplementary record. Params include every neural module loaded and
used by the generation path (generative backbone, conditioning modules, audio encoder,
and decoder/restorer), but exclude generic face detectors and geometric post-processing.
\method uses a 1.3B Wan backbone; 2.441B denotes the adapted generator components,
whereas 2.81B is the complete inference stack reported in benchmark tables. It uses a
cached empty-text embedding, so the inactive T5 encoder is not counted.

\IfFileExists{tables/causaldub_common_intersection_rows.tex}{%
% Generated by bench/analyze_causaldub_common_intersection.py.
\newcommand{\VidDubBenchCommonRows}{%
Wav2Lip & 255/256 & 6.930 & 8.514 & 99.05 \\
VideoReTalking & 252/256 & 6.883 & 8.052 & 66.18 \\
MuseTalk 1.5 & 253/256 & 4.193 & 10.650 & 57.24 \\
Diff2Lip & 255/256 & 7.588 & 7.286 & 59.17 \\
LatentSync 1.6 & 250/256 & 7.541 & 7.732 & 46.40 \\
X-Dub & 251/256 & 6.254 & 9.050 & 39.09 \\
Full-context Teacher-48 (ours) & 253/256 & 8.707 & 7.100 & 15.62 \\
Lip Forcing & 253/256 & 5.197 & 9.967 & 16.14 \\
Causal two-step VidDub & 253/256 & 8.669 & 7.240 & 17.92 \\
\oursrow\method & 253/256 & 8.702 & 7.141 & 17.96 \\
}
}{}
\begin{table*}[t]
\centering
\caption{VidDub-Bench cross-audio failure and common-intersection audit. Every
method is requested to generate 256 clips. ``Scoreable'' requires an output, a
successful SyncNet record, and matching tables/GT feature files; all metric
columns use the identical 245-item intersection throughout.}
\label{tab:causaldub-common}
\small
\setlength{\tabcolsep}{5pt}
\begin{tabularx}{\textwidth}{@{}Xcrrr@{}}
\toprule
Method & Scoreable & Sync-C $\uparrow$ & Sync-D $\downarrow$ & FVD $\downarrow$ \\
\midrule
\VidDubBenchCommonRows
\bottomrule
\end{tabularx}
\end{table*}

The main table reports means over each method's valid outputs and generation coverage,
whereas cross-method ranking uses the same 245 clips for every method. On this shared
subset, \method obtains 8.702 Sync-C versus 7.588 for the strongest external method
and 8.707 for its full-context teacher, preserving 99.9\% of teacher Sync-C; the
matched \causalvariant obtains 8.669.

\begin{table*}[t]
\centering
\caption{VoxCeleb2-30 results under in-the-wild pose and compression.}
\label{tab:vox2}
\small
\setlength{\tabcolsep}{2.6pt}
\begin{tabularx}{\textwidth}{@{}X l r rr rr rr r@{}}
\toprule
& & & \multicolumn{4}{c}{Same Audio} & \multicolumn{3}{c}{Cross Audio} \\
\cmidrule(lr){4-7}\cmidrule(lr){8-10}
Method & Params & NFE & Sync-C$\uparrow$ & Sync-D$\downarrow$ & SSIM$\uparrow$ & PSNR$\uparrow$ & Sync-C$\uparrow$ & Sync-D$\downarrow$ & FVD$\downarrow$ \\
\midrule
\multicolumn{10}{@{}l}{\textit{Feed-forward and one-step editors}} \\
\addlinespace[1pt]
Wav2Lip~\cite{wav2lip} & \ParamsWavTwoLip & 1 & 8.362 & 6.779 & 0.9297 & 31.34 & 7.800 & 7.173 & 71.73 \\
VideoReTalking~\cite{videoretalking} & \ParamsVideoReTalking & 1 & 7.697 & 6.971 & 0.9087 & 28.95 & 7.124 & 7.394 & 100.75 \\
MuseTalk 1.5~\cite{musetalk} & \ParamsMuseTalk & 1 & 5.698 & 8.464 & 0.9317 & 32.19 & 4.305 & 9.673 & 54.59 \\
\midrule
\multicolumn{10}{@{}l}{\textit{Bidirectional diffusion editors (20--50 NFE)}} \\
\addlinespace[1pt]
Diff2Lip~\cite{diff2lip} & \ParamsDiffTwoLip & 25 & 8.075 & \textbf{6.234} & 0.9341 & 33.36 & 7.533 & 6.662 & 67.15 \\
LatentSync 1.6~\cite{latentsync} & \ParamsLatentSync & 20 & 7.315 & 7.064 & \textbf{0.9533} & \textbf{34.59} & 6.239 & 8.020 & \textbf{28.89} \\
X-Dub~\cite{xdub} & \ParamsXDub & 50 & 6.298 & 8.598 & 0.8302 & 25.91 & 5.375 & 9.254 & 97.45 \\
Full-context Teacher-30 (CFG 1) & \ParamsTeacher & 30 & 7.641 & 6.971 & 0.9302 & 31.88 & 7.382 & 7.119 & 39.11 \\
\midrule
\multicolumn{10}{@{}l}{\textit{Two-step diffusion editors}} \\
\addlinespace[1pt]
\LipForcingVoxRow
\oursrow\method & \ParamsVidDub & 2 & 8.609 & 6.470 & \textbf{0.9262} & \textbf{31.20} & \textbf{8.454} & \textbf{6.579} & \textbf{46.52} \\
\bottomrule
\end{tabularx}
\end{table*}

\begin{table*}[t]
\centering
\caption{TalkVid-Bench results. $^\dagger$Wav2Lip runs at resize factor 1; an earlier run used
the adapter default of 2 and wrote half-resolution video, which corrupted visual
metrics (Sync-C/D nearly unchanged). For 13 clips at 1440p/4K, face detection exceeds
GPU memory at factor 1; these use factor 2 and are upscaled. The primary face
detector initially rejected three clear frontal-face clips at its default threshold.
A lower-threshold fallback, invoked only after a complete primary-detector miss,
recovers these clips; \method, the \causalvariant, and Teacher-48 consequently generate
and score all 500 requested clips in both tracks. Other methods use their method-specific
successful subsets; the main-paper table reports every scored count.}
\label{tab:talkvid}
\small
\setlength{\tabcolsep}{2.6pt}
\begin{tabularx}{\textwidth}{@{}X l r rr rr rr r@{}}
\toprule
& & & \multicolumn{4}{c}{Same Audio} & \multicolumn{3}{c}{Cross Audio} \\
\cmidrule(lr){4-7}\cmidrule(lr){8-10}
Method & Params & NFE & Sync-C$\uparrow$ & Sync-D$\downarrow$ & SSIM$\uparrow$ & PSNR$\uparrow$ & Sync-C$\uparrow$ & Sync-D$\downarrow$ & FVD$\downarrow$ \\
\midrule
\multicolumn{10}{@{}l}{\textit{Feed-forward and one-step editors}} \\
\addlinespace[1pt]
Wav2Lip$^\dagger$~\cite{wav2lip} & \ParamsWavTwoLip & 1 & 7.209 & 7.420 & 0.9218 & 29.83 & 6.750 & 7.932 & 111.19 \\
VideoReTalking~\cite{videoretalking} & \ParamsVideoReTalking & 1 & 6.787 & 7.552 & 0.8948 & 28.00 & 6.365 & 7.868 & 84.51 \\
MuseTalk 1.5~\cite{musetalk} & \ParamsMuseTalk & 1 & 4.968 & 9.106 & \textbf{0.9305} & 29.60 & 4.371 & 9.625 & 36.09 \\
\midrule
\multicolumn{10}{@{}l}{\textit{Bidirectional diffusion editors (20--50 NFE)}} \\
\addlinespace[1pt]
Diff2Lip~\cite{diff2lip} & \ParamsDiffTwoLip & 25 & 7.775 & \textbf{6.383} & 0.9157 & 30.60 & 6.944 & \textbf{7.016} & 73.37 \\
LatentSync 1.6~\cite{latentsync} & \ParamsLatentSync & 20 & 7.516 & 6.965 & 0.8949 & \textbf{31.09} & 6.860 & 7.512 & 59.27 \\
X-Dub~\cite{xdub} & \ParamsXDub & 50 & 5.492 & 9.129 & 0.6980 & 21.86 & 5.087 & 9.437 & 32.63 \\
Full-context Teacher-48 (ours) & \ParamsTeacher & 48 &
7.741 & 7.170 & 0.9125 & 28.84 & 7.663 & 7.266 & \textbf{8.51} \\
\midrule
\multicolumn{10}{@{}l}{\textit{Two-step diffusion editors}} \\
\addlinespace[1pt]
\LipForcingTalkRow
\oursrow\method & \ParamsVidDub & 2 & \textbf{7.740} & 7.216 &
\textbf{0.9125} & \textbf{28.78} & \textbf{7.646} & 7.309 & \textbf{10.25} \\
\bottomrule
\end{tabularx}
\end{table*}

\begin{table*}[t]
\centering
\caption{TalkVid cross-audio failure and common-intersection audit. ``Success'' requires
an output, successful SyncNet record, and matching tables/GT feature files.
All remaining metrics use the identical 308-item intersection (258 source videos);
brackets are 95\% source-cluster bootstrap intervals with 10,000 resamples. FVD is
recomputed on that intersection and kept as a point estimate because it is a global
feature-distribution functional.}
\label{tab:talkvid-common}
\small
\setlength{\tabcolsep}{4.0pt}
\begin{tabularx}{\textwidth}{@{}Xcrrr@{}}
\toprule
Method & Success & Sync-C $\uparrow$ [95\% CI] & Sync-D $\downarrow$ [95\% CI] & FVD $\downarrow$ \\
\midrule
\TalkVidCommonRows
\bottomrule
\end{tabularx}
\end{table*}

The common-intersection conclusion agrees with the method-specific audit. \method
reduces FVD from 14.24 to 12.06 relative to the matched causal variant, while the
causal variant has a 0.070 Sync-C advantage on this subset. Across methods, moving
from each native successful subset to the strict common set changes mean Sync-C by
at most 0.086; failed outputs remain explicit and are never recast as scored examples.

All methods use the same decoding, face-crop, synchronization, and
feature-extraction protocol when released checkpoints permit. We report
Wav2Lip, VideoReTalking, Diff2Lip, MuseTalk 1.5, LatentSync 1.6, public X-Dub,
Lip Forcing 14B and its OmniAvatar-LS 1.3B/14B variants \cite{lipforcing}, our
full-context teacher, the causal AR model,
selected two/four-step initializers, and final \method. If a public
method cannot support cross audio, the evaluated long-sequence protocol, or the same resolution, the
missing capability is marked instead of silently changing the protocol.
OmniAvatar-LS receives a fixed reference frame extracted from each source and
is evaluated as an avatar-generation baseline. Because it is not conditioned
on the complete source video, its SSIM, mouth LPIPS, and source-motion retention
are marked not applicable rather than compared with editing methods.

\begin{table*}[t]
\centering
\caption{Complete cross-audio benchmark summary used by the main-paper comparison.
Each cell reports Sync-C / FVD; higher Sync-C and lower FVD are better. These displayed
means use each method's valid outputs and are accompanied by generation coverage
in the detailed tables. Cross-method ranking is verified separately on the shared
TalkVid-308 and VidDub-Bench-245 valid subsets in
Tables~\ref{tab:talkvid-common} and~\ref{tab:causaldub-common}.}
\label{tab:supp-public-results}
\small
\begin{tabular}{@{}lcccc@{}}
\toprule
Method & NFE & HDTF-30 & VoxCeleb2-30 & TalkVid-Bench \\
\midrule
Wav2Lip~\cite{wav2lip} & 1 & 8.782 / 129.51 & 7.800 / 71.73 & 6.750 / 111.19 \\
VideoReTalking~\cite{videoretalking} & 1 & 8.312 / 105.72 & 7.124 / 100.75 & 6.365 / 84.51 \\
Diff2Lip~\cite{diff2lip} & 25 & 8.680 / 99.70 & 7.533 / 67.15 & 6.944 / 73.37 \\
MuseTalk 1.5~\cite{musetalk} & 1 & 7.116 / 40.10 & 4.305 / 54.59 & 4.371 / 36.09 \\
X-Dub~\cite{xdub} & 50 & 5.027 / 66.05 & 5.375 / 97.45 & 5.087 / 32.63 \\
LatentSync 1.6~\cite{latentsync} & 20 & 8.160 / 28.05 & 6.239 / 28.89 & 6.860 / 59.27 \\
Full-context teacher (30/30/48 NFE) & 30/30/48 & 9.422 / 27.49 & 7.382 / 39.11 & 7.663 / 8.51 \\
Causal diagnostic (4 NFE) & 4 & 9.537 / 38.43 & 7.789 / 70.05 & 7.456 / 15.16 \\
Final \method & 2 & 10.102 / 27.66 & 8.454 / 46.52 & 7.646 / 10.25 \\
\bottomrule
\end{tabular}
\end{table*}

Editing baselines receive the complete source video and target waveform under their
released preprocessing. The full-context teacher uses 30 or 48 denoising steps,
whereas \method uses two denoising steps and one conditional forward at each step.
The unusually low TalkVid SSIM of public X-Dub (0.6980) and Lip Forcing (0.7698)
is reported as observed under their released input contracts; we retain the outputs
for qualitative inspection and do not use those values alone to claim perceptual
superiority.
The four-NFE causal diagnostic in Table~\ref{tab:supp-public-results} is not the
matched \causalvariant evaluated at two NFE in
Tables~\ref{tab:supp-causal-design} and~\ref{tab:supp-frozen50}.
OmniAvatar-LS receives only one reference image and is therefore excluded from
source-motion preservation comparisons.

On this frozen 50-pair set, the \causalvariant and the two-branch full-context
teacher evaluated at 48 NFE obtain 8.772 and 8.816 Sync-C, respectively, with their
corresponding FVD reported in the same table.
Against the closest public causal few-step baseline, Lip Forcing 14B, the
\causalvariant uses
the same reported NFE and one conditional branch while improving Sync-C
from 3.973 to 8.772; it additionally executes one clean state-commit forward
per block, while Lip Forcing obtains the lower FVD. The \causalvariant also improves both Sync-C and FVD over
LatentSync on this subset; the result establishes the reported synchronization
ordering under the evaluated inference protocols.

Figure~\ref{fig:supp-fewstep-pareto} summarizes cross Sync-C/FVD on three public
benchmarks. Its 10/30/60-second curves document the persistent-cache architecture comparison;
the final Long-20 evaluation uses the overlapping-window inference reported in
Table~\ref{tab:supp-long20-external}.

\IfFileExists{tables/frozen50_results.tex}{%
\begin{table*}[t]
\centering
\caption{Frozen 50-pair cross-audio comparison. All methods use the same
clips and evaluation protocol; these results are separate from the 256-source
benchmark protocol.}
\label{tab:supp-frozen50}
\small
\newcommand{\FrozenFiftyRows}{%
Full-context teacher & 1.3B & 48 & 2 & 8.816 & 6.950 & 43.86 \\
LatentSync 1.6 & -- & 20 & 1 & 7.629 & 7.589 & 69.93 \\
Lip Forcing & 14B & 2 & 1 & 3.973 & 11.164 & 34.70 \\
Causal two-step VidDub & 1.3B & 2 & 1 & 8.772 & 7.087 & 49.04 \\
}

\resizebox{\textwidth}{!}{%
\begin{tabular}{@{}lcrrrrr@{}}
\toprule
Method & Backbone & NFE & Br. & Sync-C$\uparrow$ & Sync-D$\downarrow$ & FVD$\downarrow$ \\
\midrule
\FrozenFiftyRows
\bottomrule
\end{tabular}
}
\end{table*}
}{}

\section{Qualitative Evaluation and Artifact Taxonomy}
\label{supp:qualitative}
The following twenty figures cover ten videos: four uniformly spaced complete cases
from HDTF-30 and VoxCeleb2-30, and two complete VidDub-Bench cases. Every case has
separate same- and cross-audio tracks and seven synchronized temporal samples per
method. Public-benchmark figures compare the source, six external editors, the
full-context teacher, and \method; VidDub-Bench figures retain the source,
LatentSync, teacher, and \method for readability. Cross-audio panels additionally
show the target waveform and fixed acoustic time markers. Selection manifests and
seeds accompany all figures to prevent frame- or case-level cherry-picking.

The failure taxonomy includes weak articulation, exaggerated mouth motion, teeth or
tongue flicker, boundary discontinuity, color drift, identity drift, and accumulated
audio offset. Audible videos, tracking failures, and cases where metric ordering
disagrees with visual judgment are included in the supplementary-video package.

\IfFileExists{figures/qualitative_appendix_figures.tex}{%
  \begin{figure*}[p]
\centering
\includegraphics[width=.985\textwidth]{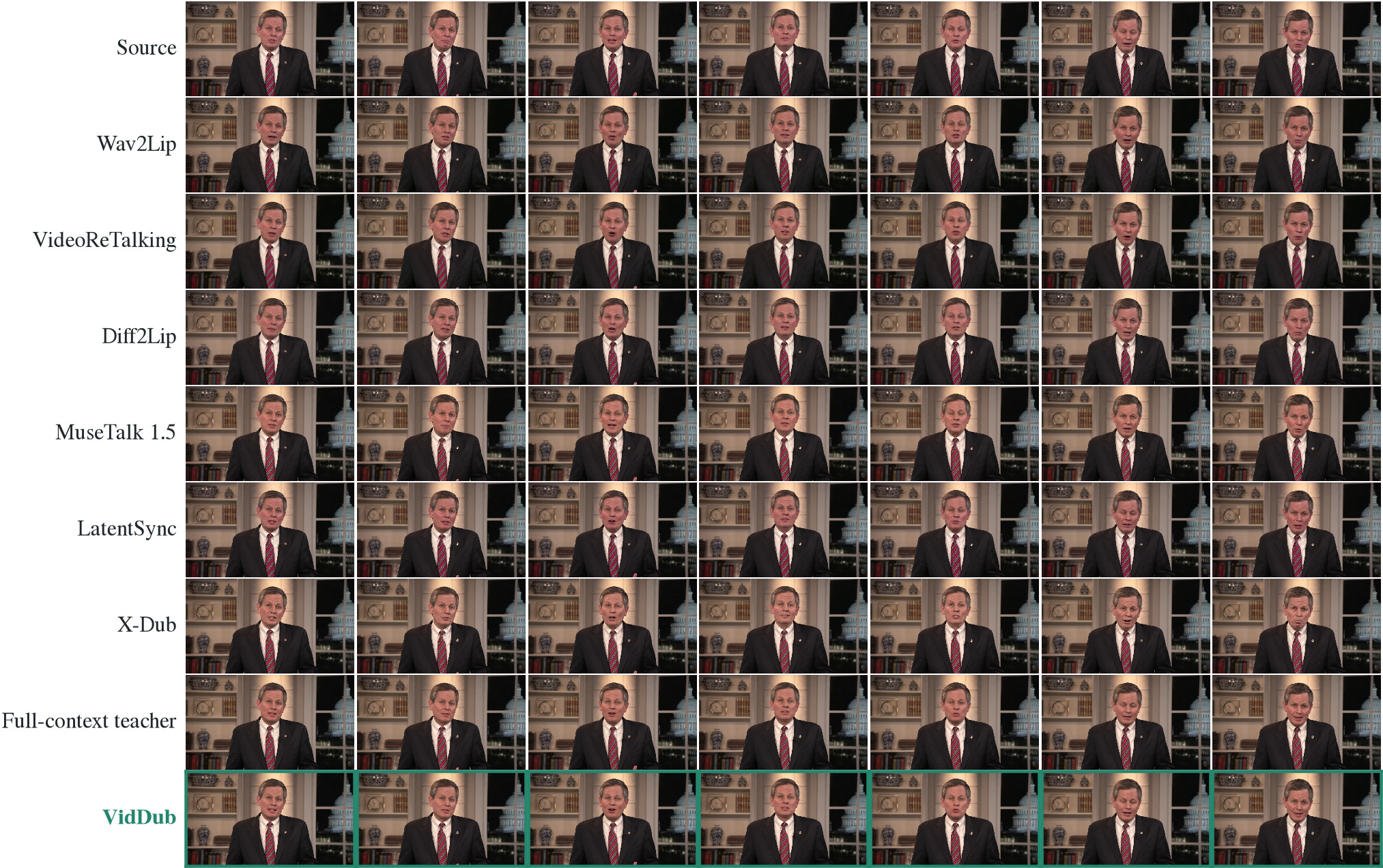}
\caption{HDTF-30 case 1, same-audio track. Rows compare the source and editing methods at matched temporal positions. S1--S7 mark the same seven temporal positions across methods.}
\label{fig:qual-hdtf-1-same}
\end{figure*}

\begin{figure*}[p]
\centering
\includegraphics[width=.985\textwidth]{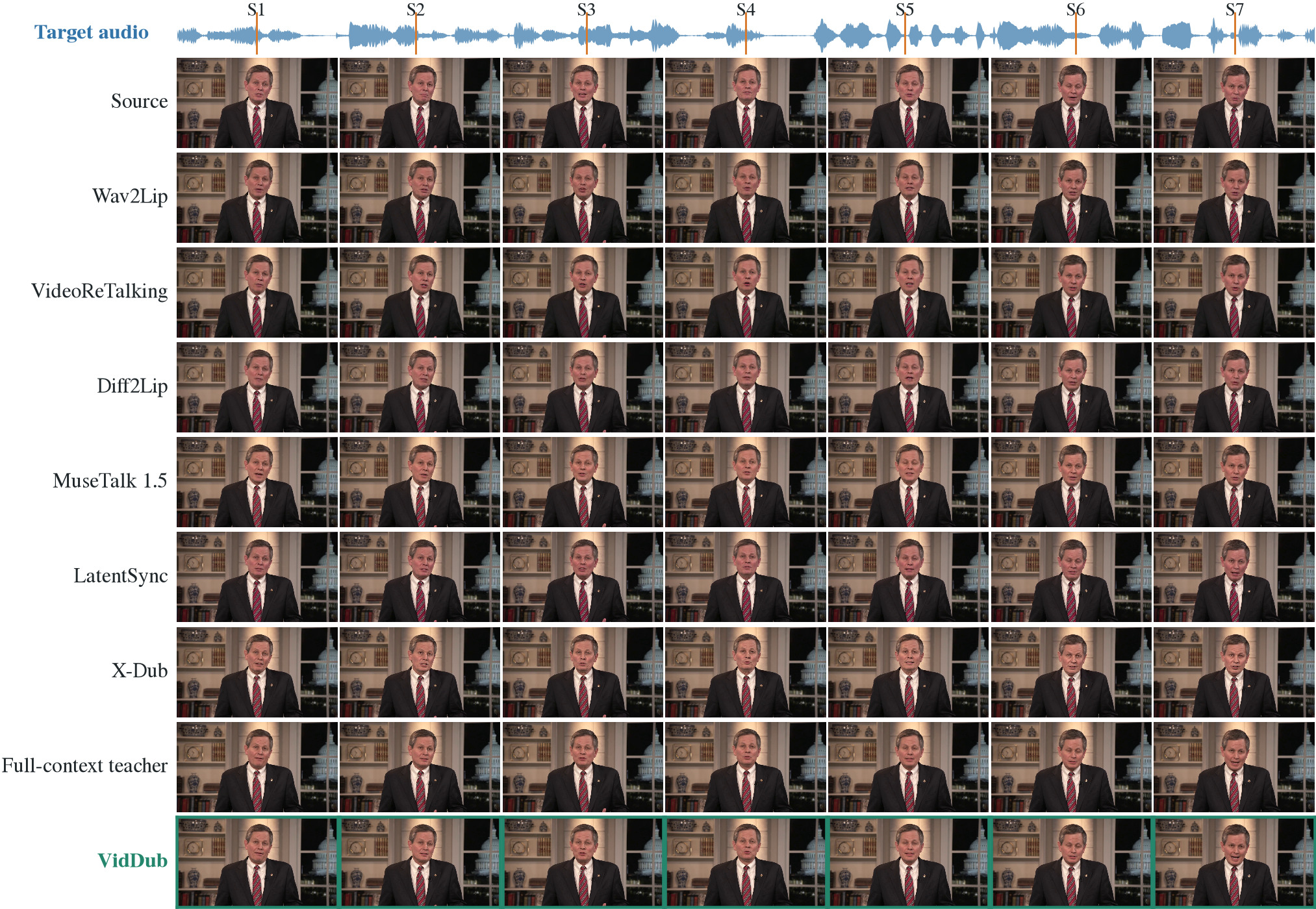}
\caption{HDTF-30 case 1, cross-audio track. Rows compare the source and editing methods at matched temporal positions. The panel includes the actual target waveform; S1--S7 mark aligned acoustic segments rather than phoneme ground truth.}
\label{fig:qual-hdtf-1-cross}
\end{figure*}

\begin{figure*}[p]
\centering
\includegraphics[width=.985\textwidth]{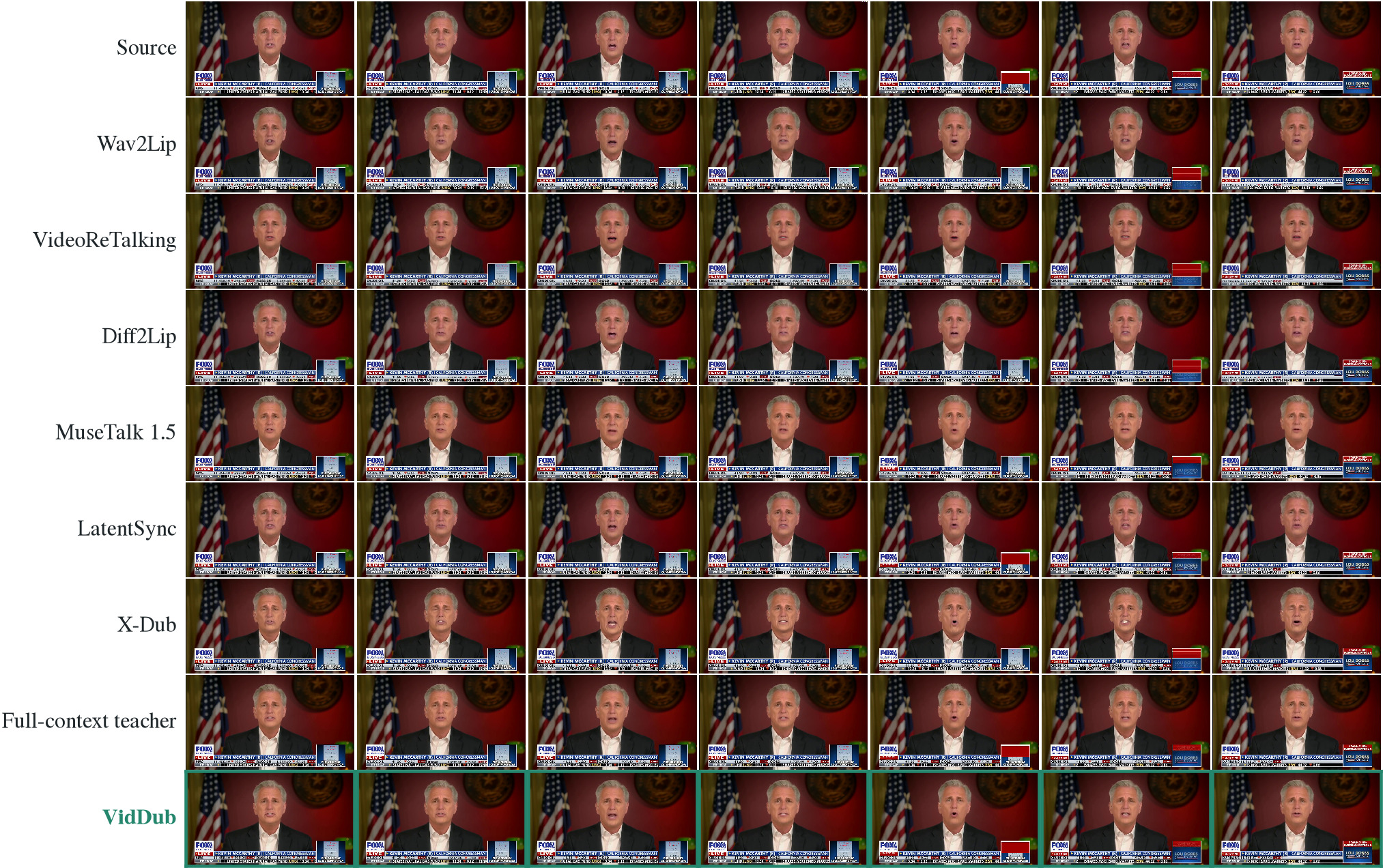}
\caption{HDTF-30 case 2, same-audio track. Rows compare the source and editing methods at matched temporal positions. S1--S7 mark the same seven temporal positions across methods.}
\label{fig:qual-hdtf-2-same}
\end{figure*}

\begin{figure*}[p]
\centering
\includegraphics[width=.985\textwidth]{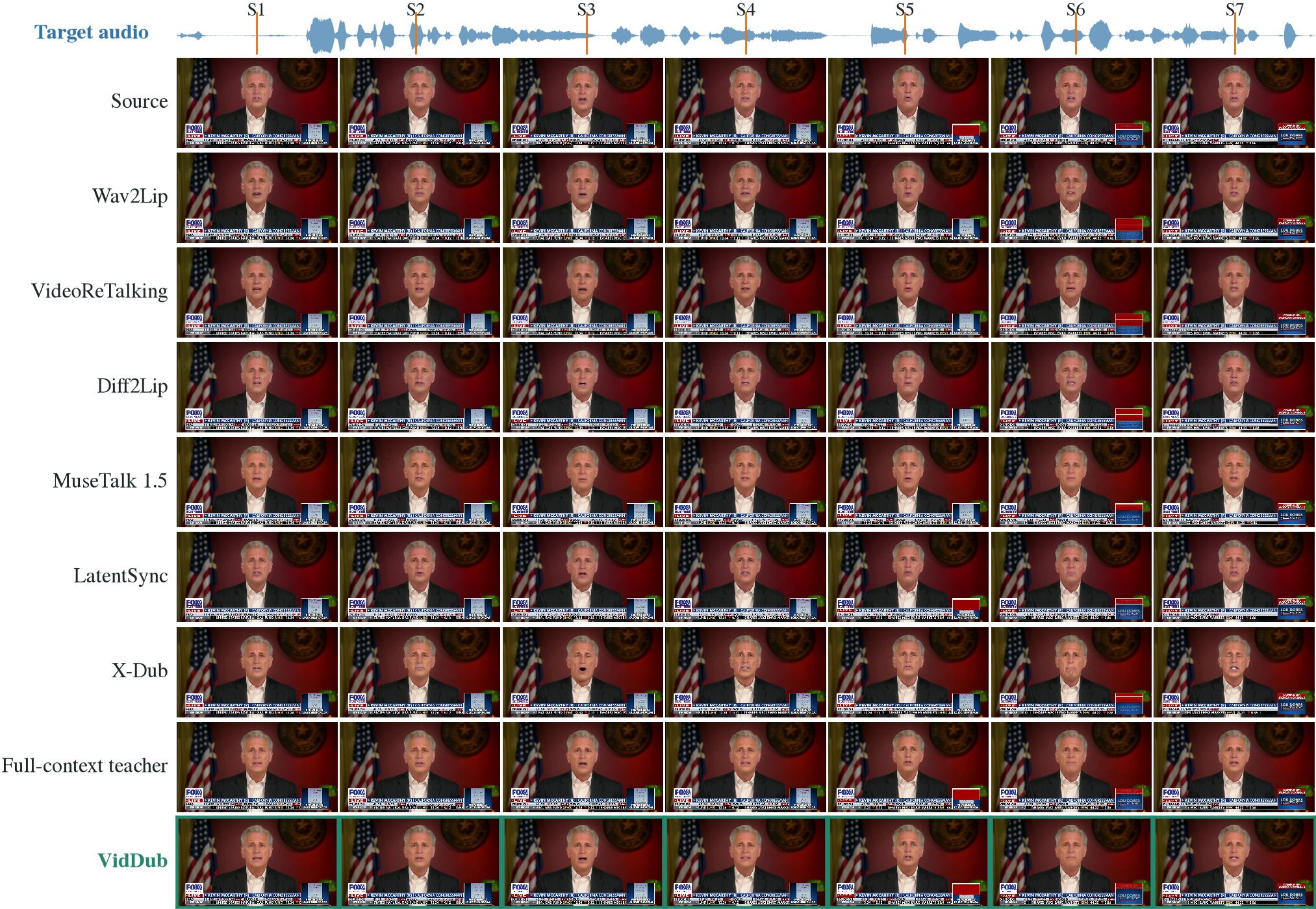}
\caption{HDTF-30 case 2, cross-audio track. Rows compare the source and editing methods at matched temporal positions. The panel includes the actual target waveform; S1--S7 mark aligned acoustic segments rather than phoneme ground truth.}
\label{fig:qual-hdtf-2-cross}
\end{figure*}

\begin{figure*}[p]
\centering
\includegraphics[width=.985\textwidth]{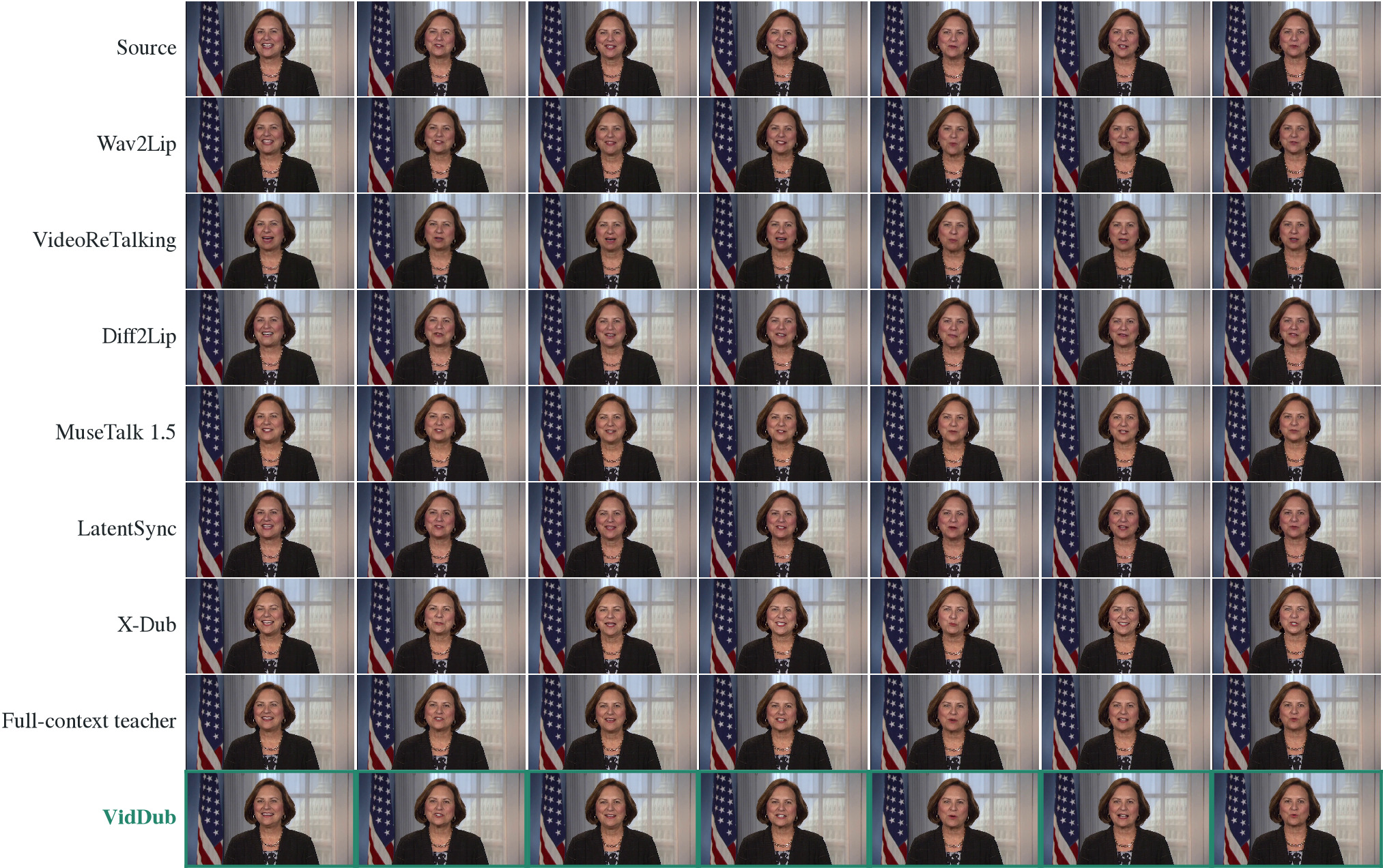}
\caption{HDTF-30 case 3, same-audio track. Rows compare the source and editing methods at matched temporal positions. S1--S7 mark the same seven temporal positions across methods.}
\label{fig:qual-hdtf-3-same}
\end{figure*}

\begin{figure*}[p]
\centering
\includegraphics[width=.985\textwidth]{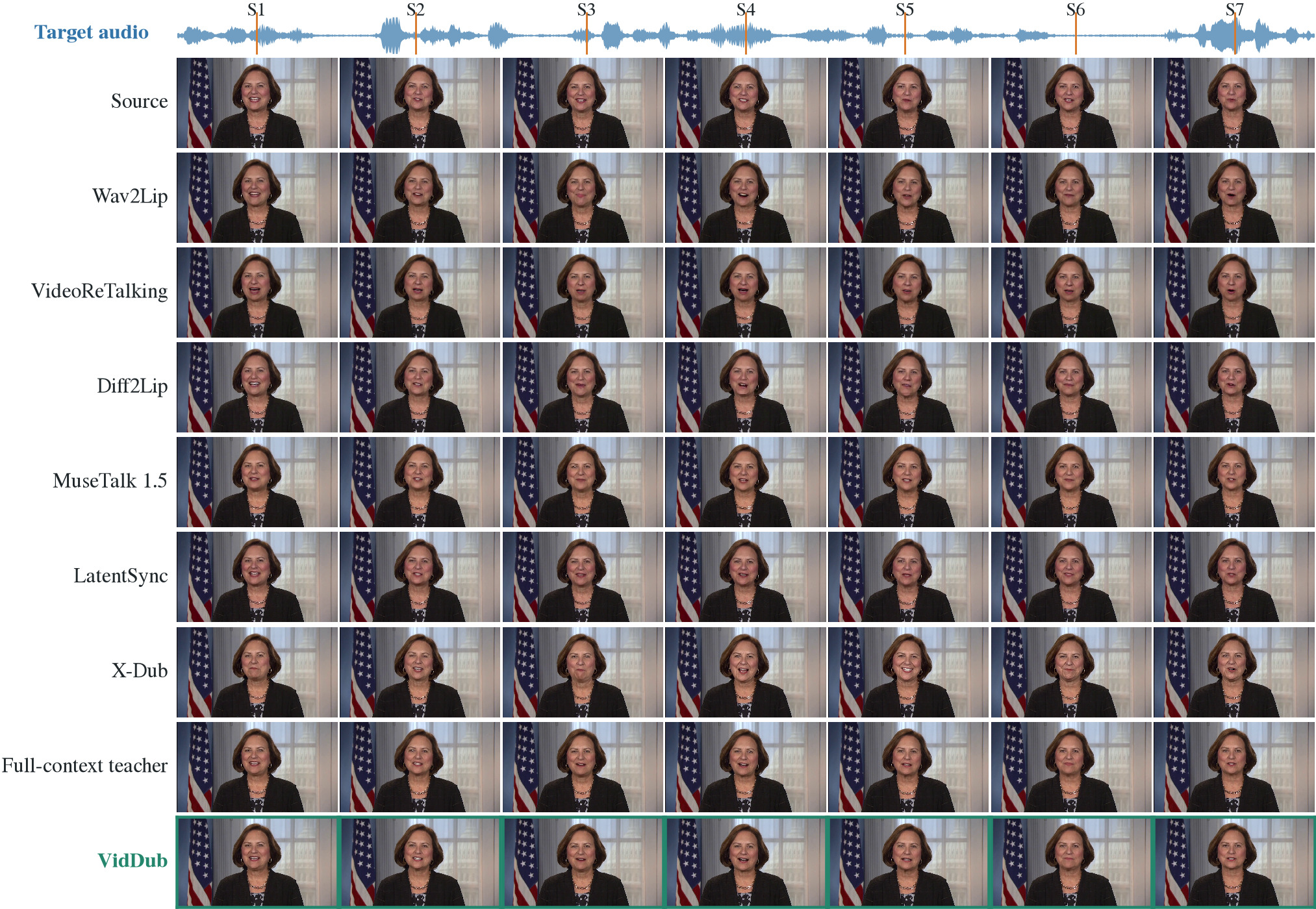}
\caption{HDTF-30 case 3, cross-audio track. Rows compare the source and editing methods at matched temporal positions. The panel includes the actual target waveform; S1--S7 mark aligned acoustic segments rather than phoneme ground truth.}
\label{fig:qual-hdtf-3-cross}
\end{figure*}

\begin{figure*}[p]
\centering
\includegraphics[width=.985\textwidth]{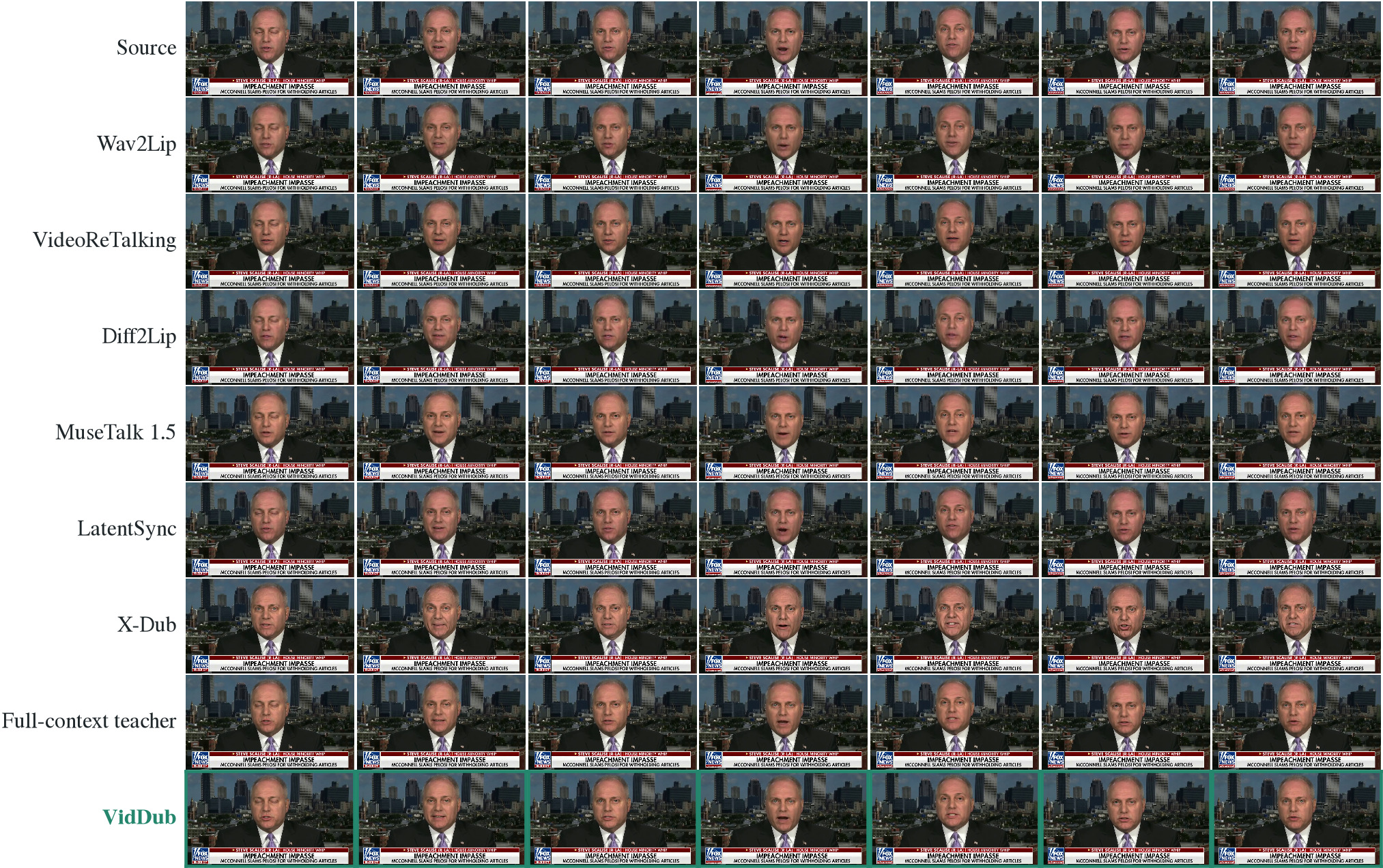}
\caption{HDTF-30 case 4, same-audio track. Rows compare the source and editing methods at matched temporal positions. S1--S7 mark the same seven temporal positions across methods.}
\label{fig:qual-hdtf-4-same}
\end{figure*}

\begin{figure*}[p]
\centering
\includegraphics[width=.985\textwidth]{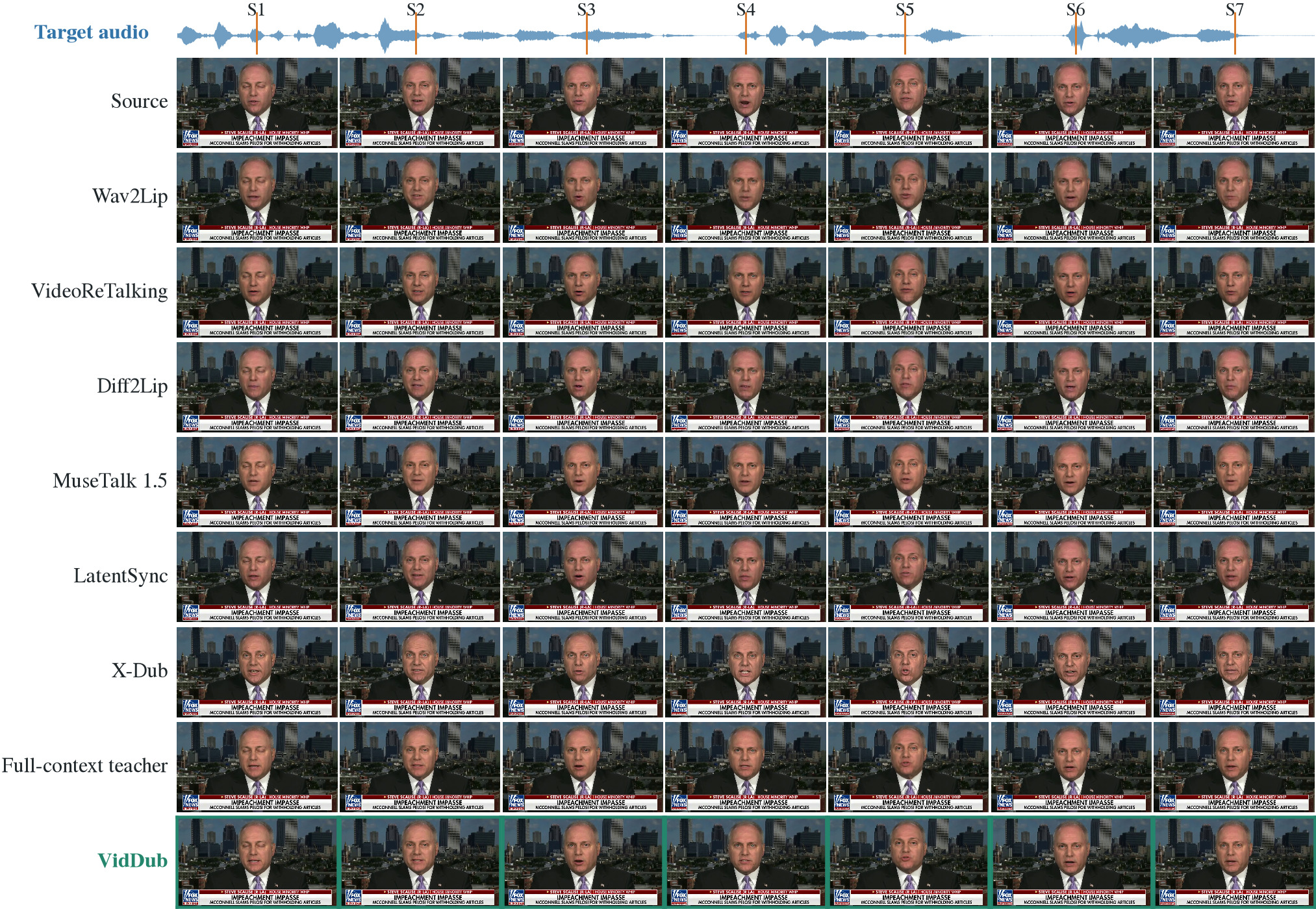}
\caption{HDTF-30 case 4, cross-audio track. Rows compare the source and editing methods at matched temporal positions. The panel includes the actual target waveform; S1--S7 mark aligned acoustic segments rather than phoneme ground truth.}
\label{fig:qual-hdtf-4-cross}
\end{figure*}

\begin{figure*}[p]
\centering
\includegraphics[width=.985\textwidth]{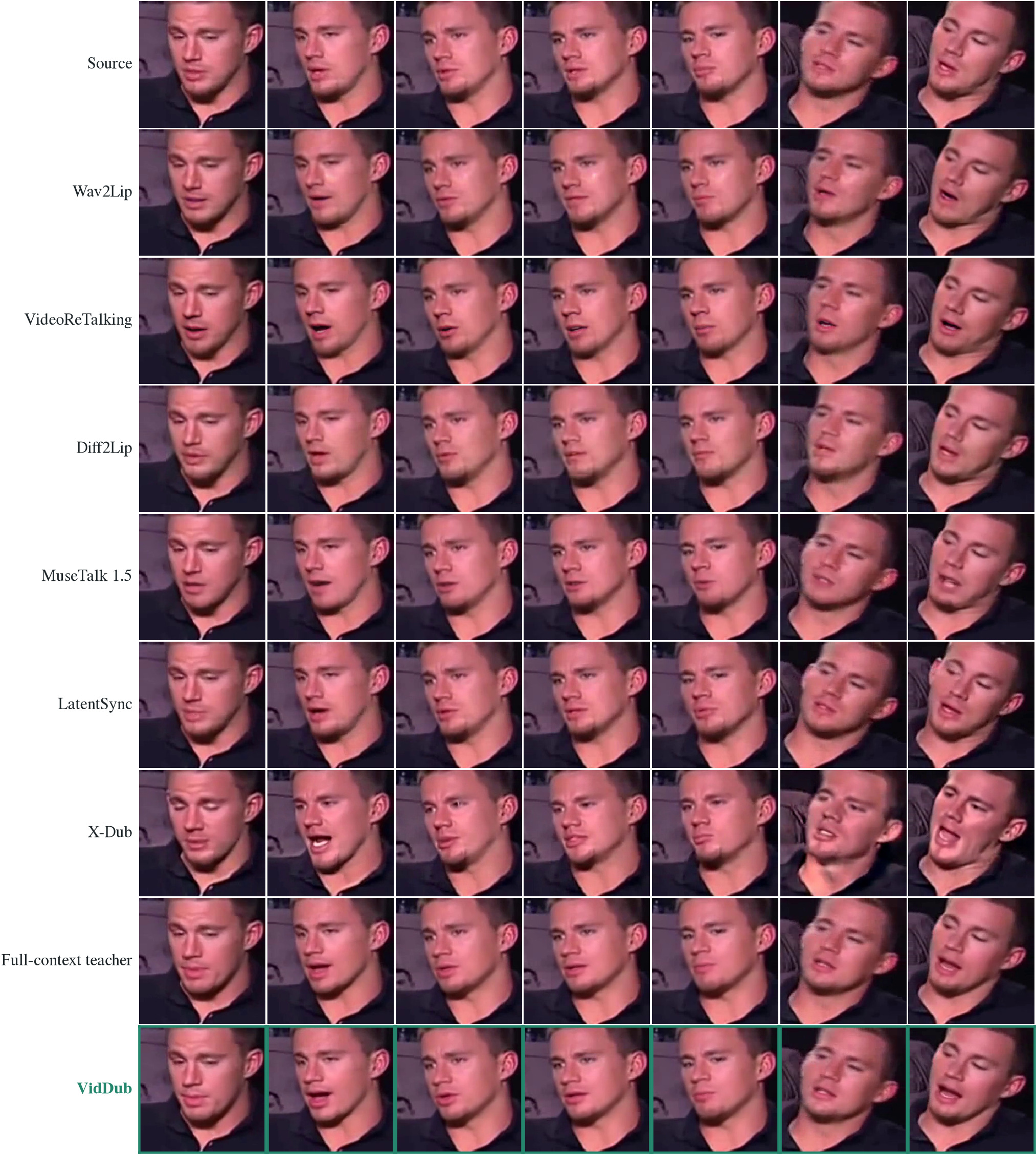}
\caption{VoxCeleb2-30 case 1, same-audio track. Rows compare the source and editing methods at matched temporal positions. S1--S7 mark the same seven temporal positions across methods.}
\label{fig:qual-vox2-1-same}
\end{figure*}

\begin{figure*}[p]
\centering
\includegraphics[width=.985\textwidth]{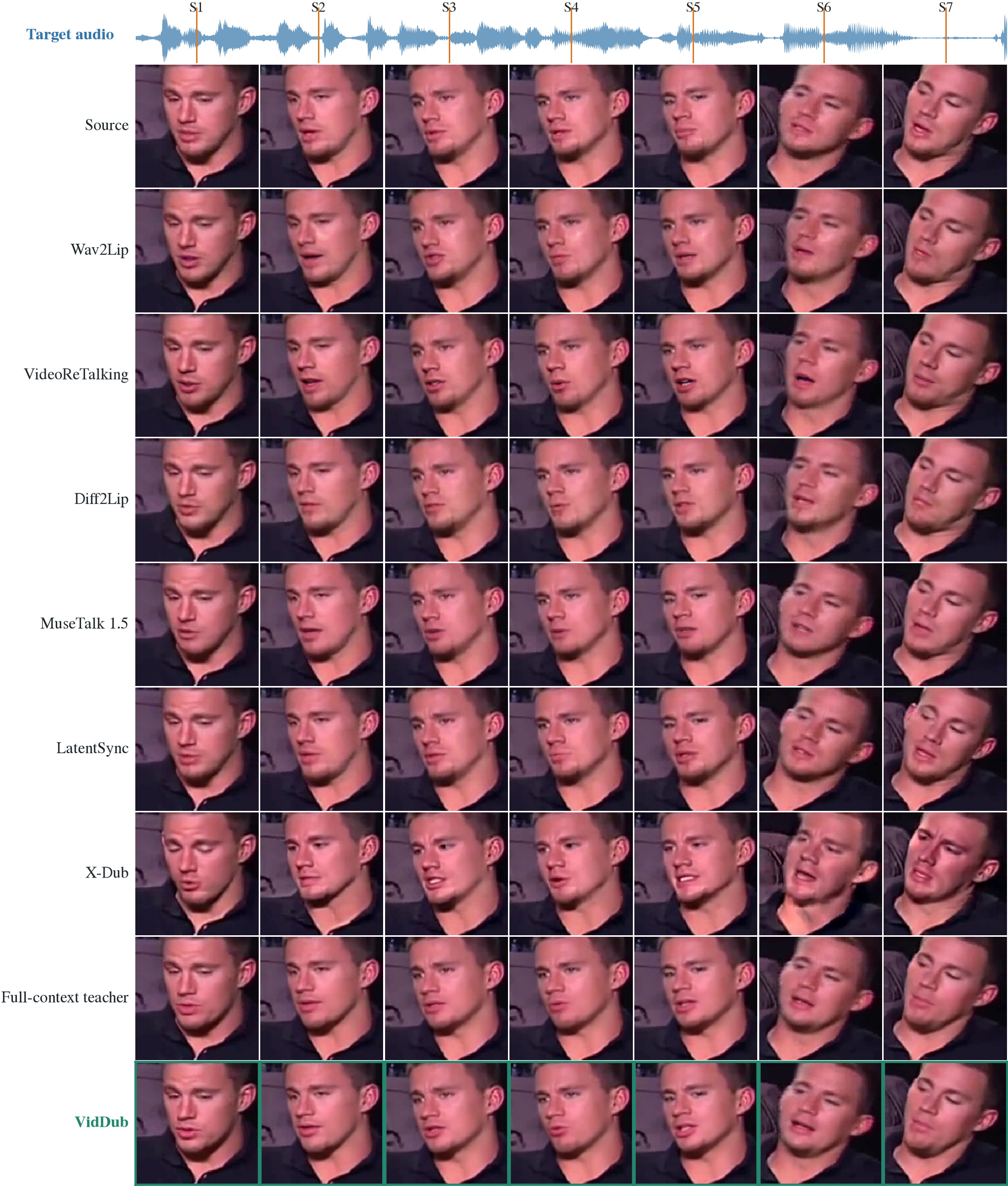}
\caption{VoxCeleb2-30 case 1, cross-audio track. Rows compare the source and editing methods at matched temporal positions. The panel includes the actual target waveform; S1--S7 mark aligned acoustic segments rather than phoneme ground truth.}
\label{fig:qual-vox2-1-cross}
\end{figure*}

\begin{figure*}[p]
\centering
\includegraphics[width=.985\textwidth]{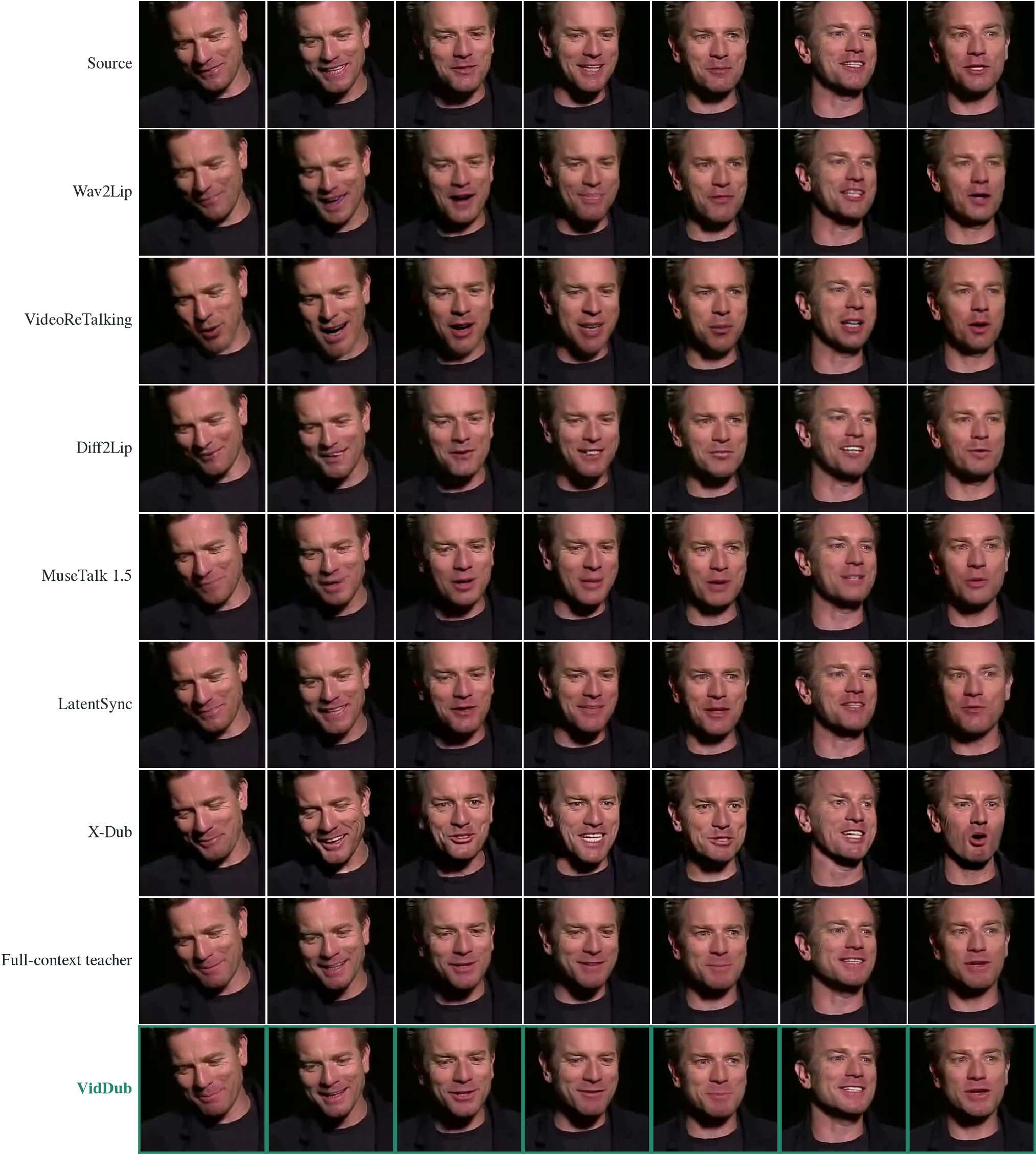}
\caption{VoxCeleb2-30 case 2, same-audio track. Rows compare the source and editing methods at matched temporal positions. S1--S7 mark the same seven temporal positions across methods.}
\label{fig:qual-vox2-2-same}
\end{figure*}

\begin{figure*}[p]
\centering
\includegraphics[width=.985\textwidth]{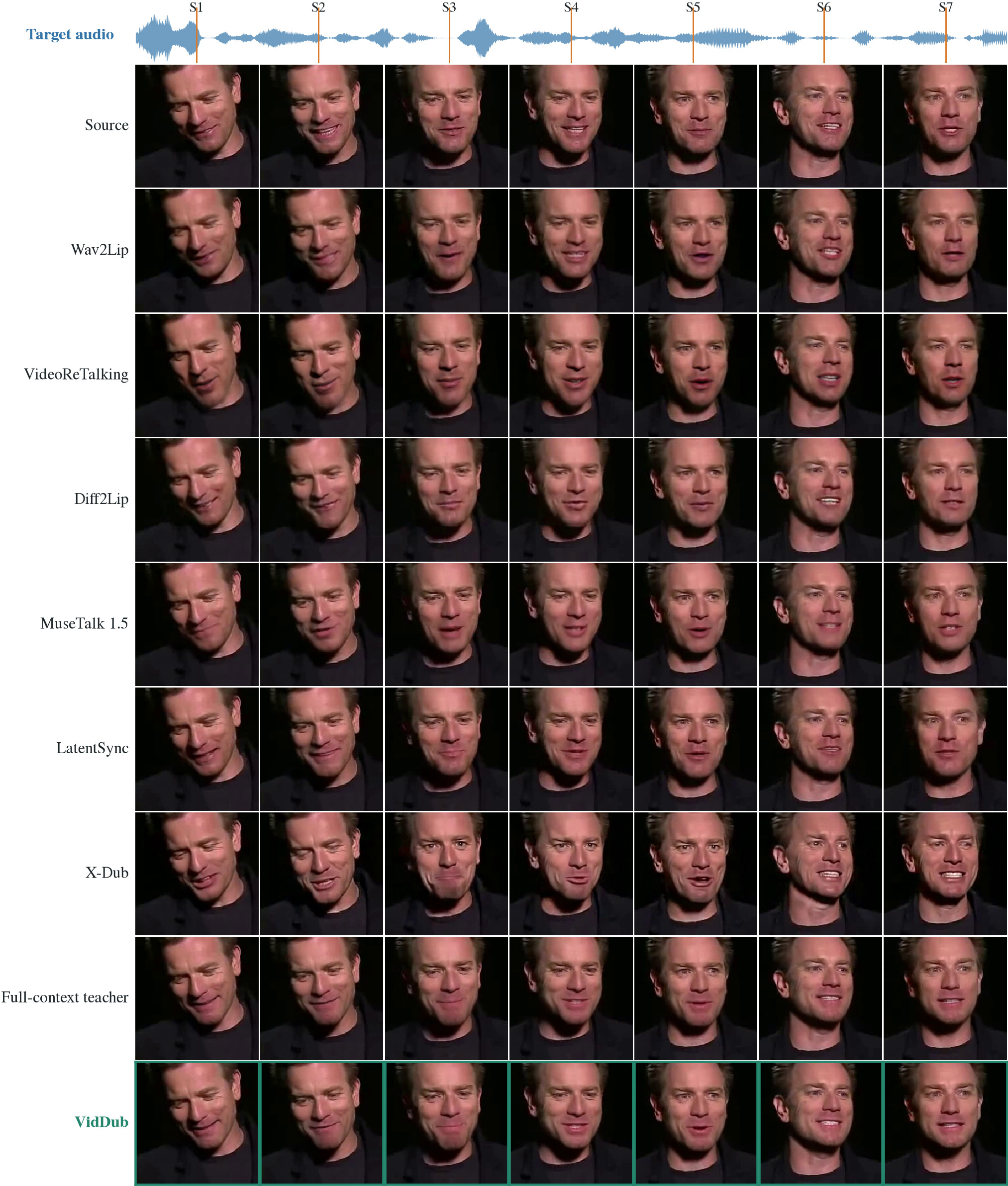}
\caption{VoxCeleb2-30 case 2, cross-audio track. Rows compare the source and editing methods at matched temporal positions. The panel includes the actual target waveform; S1--S7 mark aligned acoustic segments rather than phoneme ground truth.}
\label{fig:qual-vox2-2-cross}
\end{figure*}

\begin{figure*}[p]
\centering
\includegraphics[width=.985\textwidth]{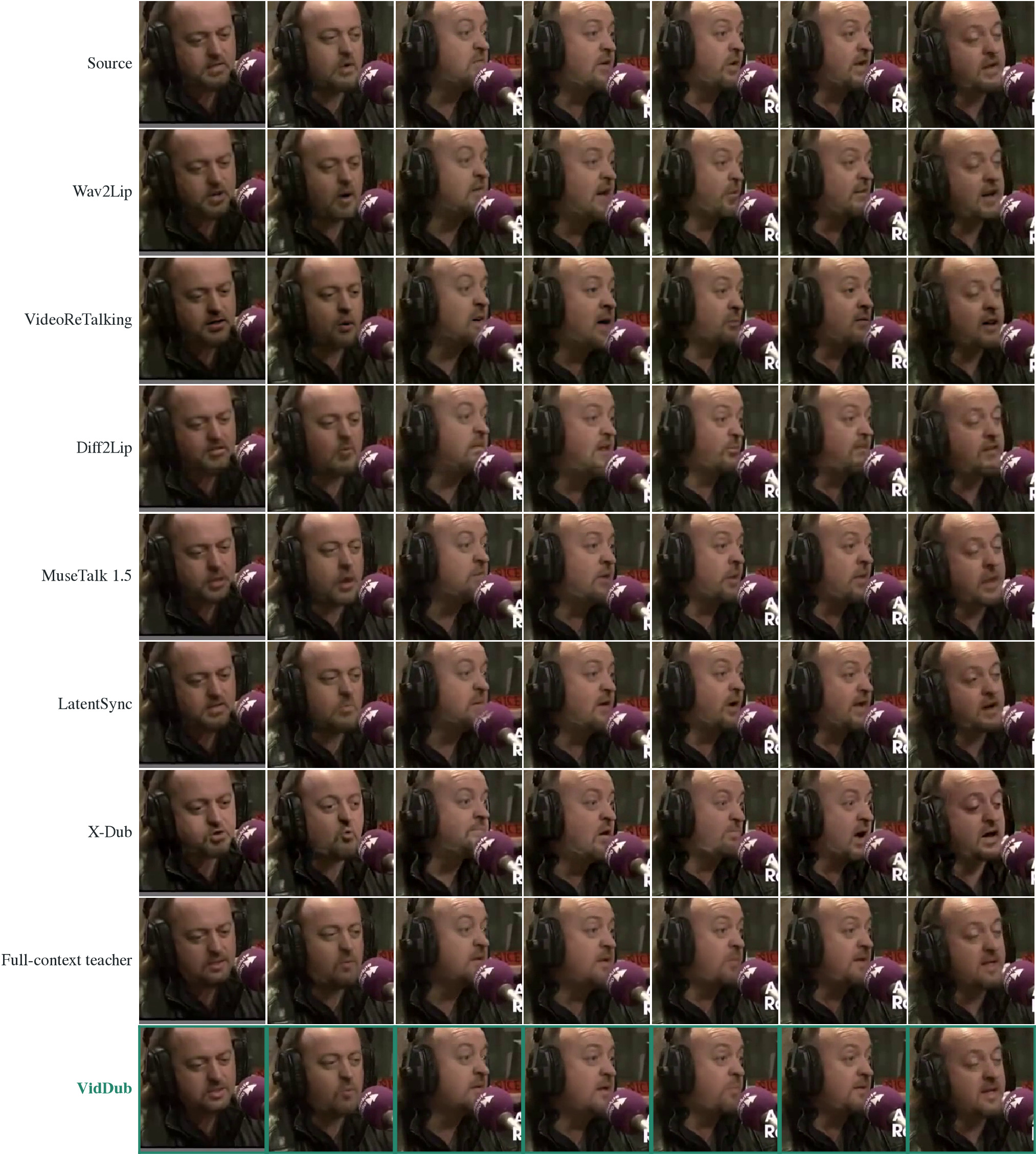}
\caption{VoxCeleb2-30 case 3, same-audio track. Rows compare the source and editing methods at matched temporal positions. S1--S7 mark the same seven temporal positions across methods.}
\label{fig:qual-vox2-3-same}
\end{figure*}

\begin{figure*}[p]
\centering
\includegraphics[width=.985\textwidth]{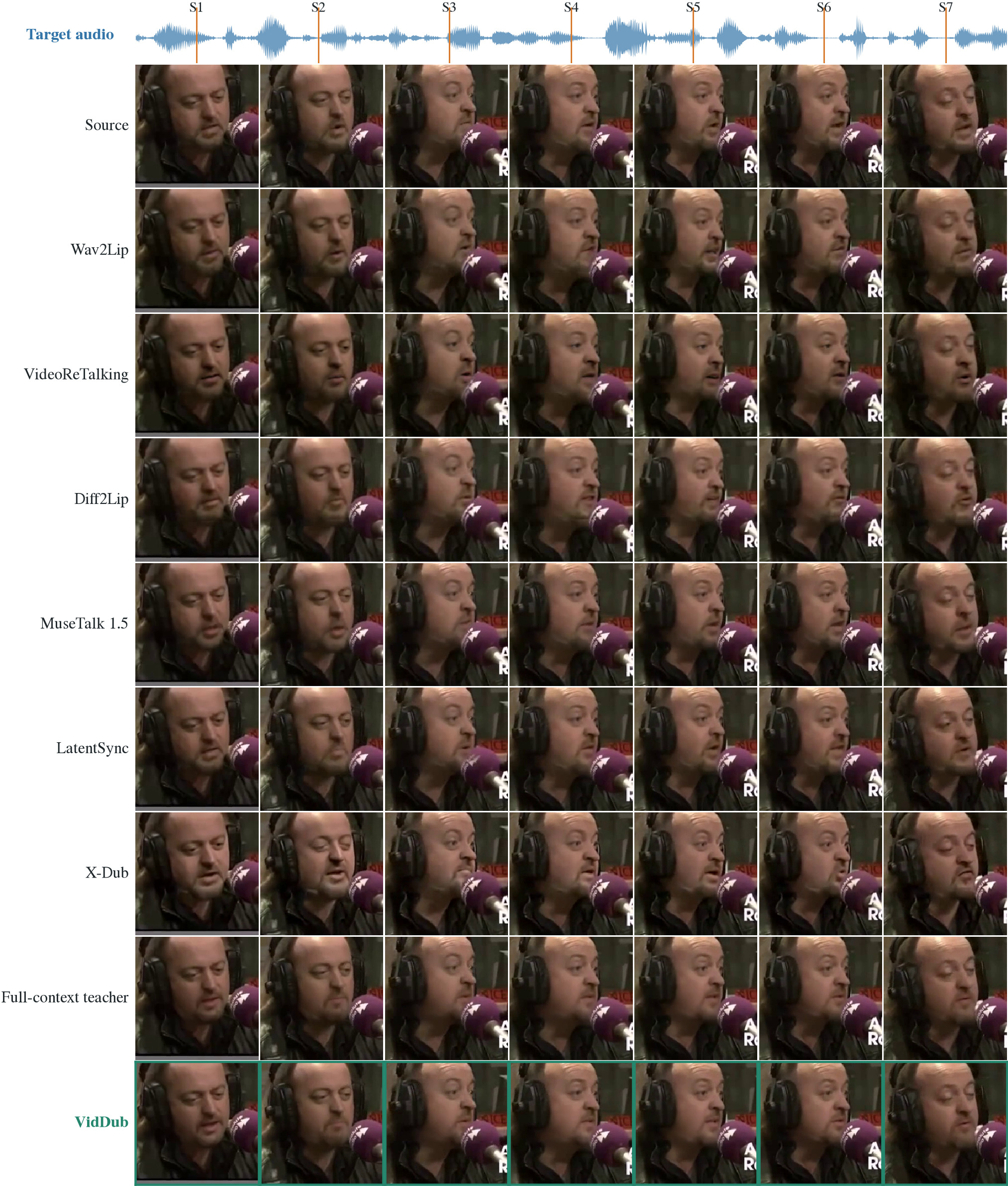}
\caption{VoxCeleb2-30 case 3, cross-audio track. Rows compare the source and editing methods at matched temporal positions. The panel includes the actual target waveform; S1--S7 mark aligned acoustic segments rather than phoneme ground truth.}
\label{fig:qual-vox2-3-cross}
\end{figure*}

\begin{figure*}[p]
\centering
\includegraphics[width=.985\textwidth]{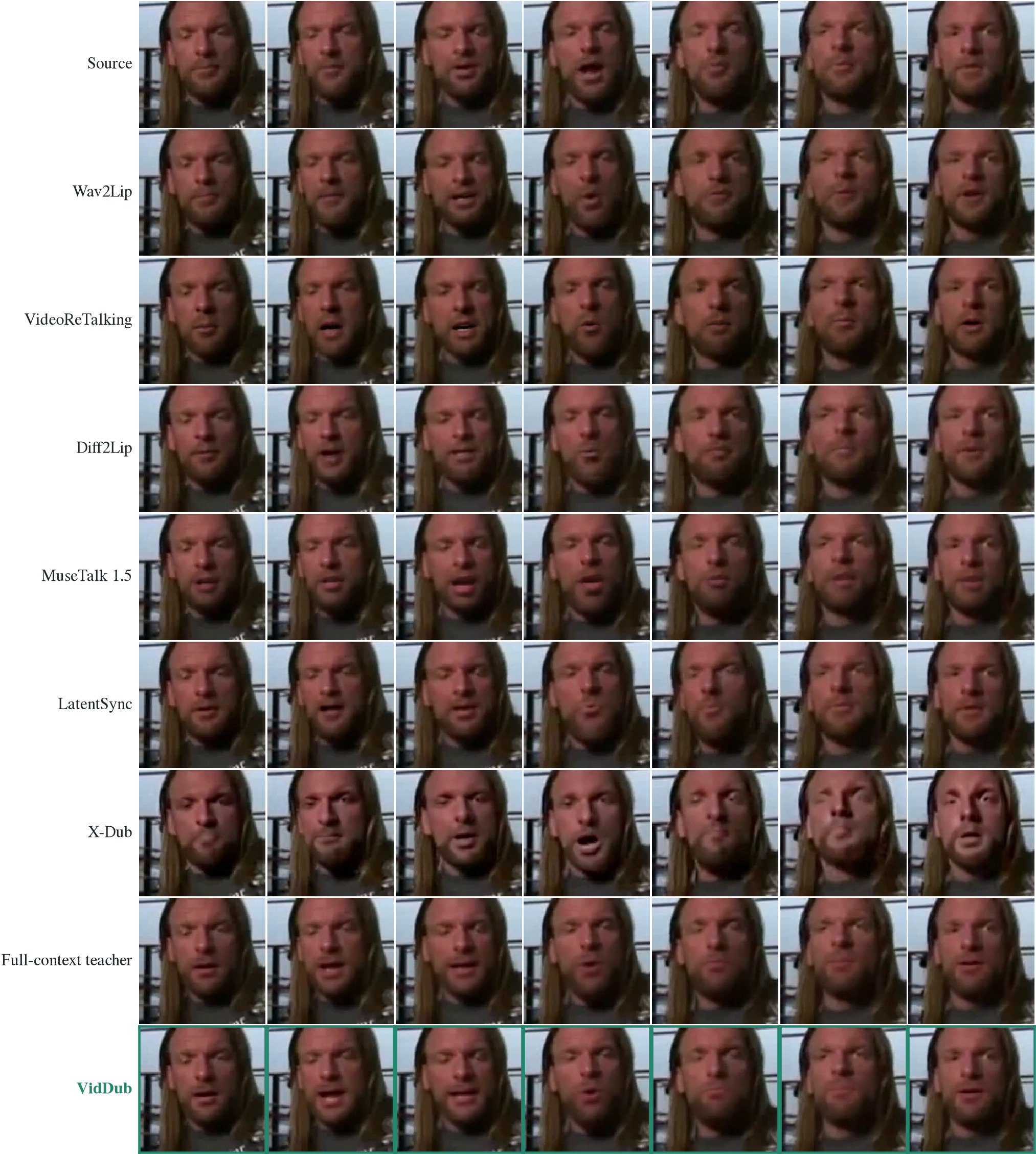}
\caption{VoxCeleb2-30 case 4, same-audio track. Rows compare the source and editing methods at matched temporal positions. S1--S7 mark the same seven temporal positions across methods.}
\label{fig:qual-vox2-4-same}
\end{figure*}

\begin{figure*}[p]
\centering
\includegraphics[width=.985\textwidth]{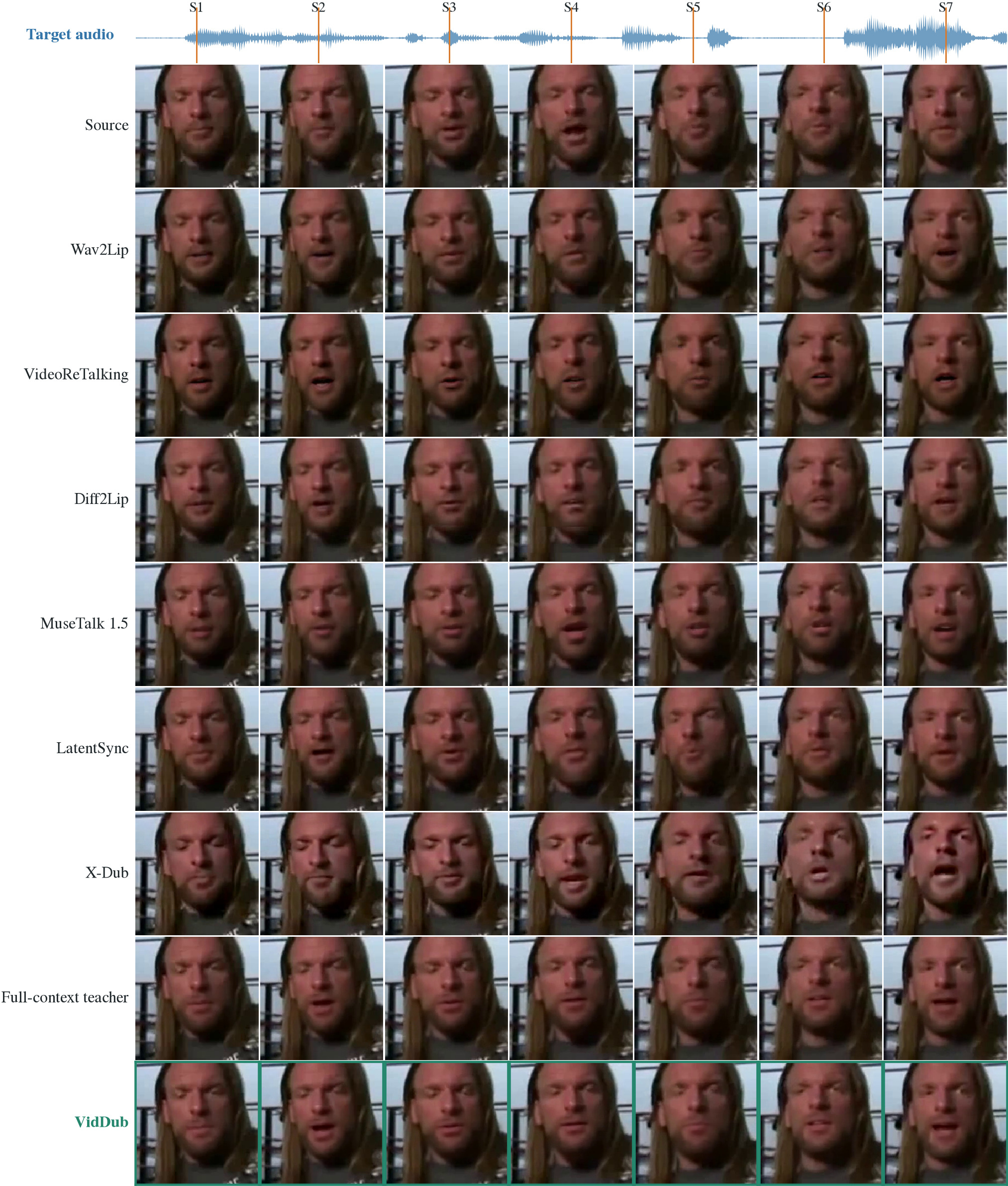}
\caption{VoxCeleb2-30 case 4, cross-audio track. Rows compare the source and editing methods at matched temporal positions. The panel includes the actual target waveform; S1--S7 mark aligned acoustic segments rather than phoneme ground truth.}
\label{fig:qual-vox2-4-cross}
\end{figure*}

\begin{figure*}[p]
\centering
\includegraphics[width=.985\textwidth]{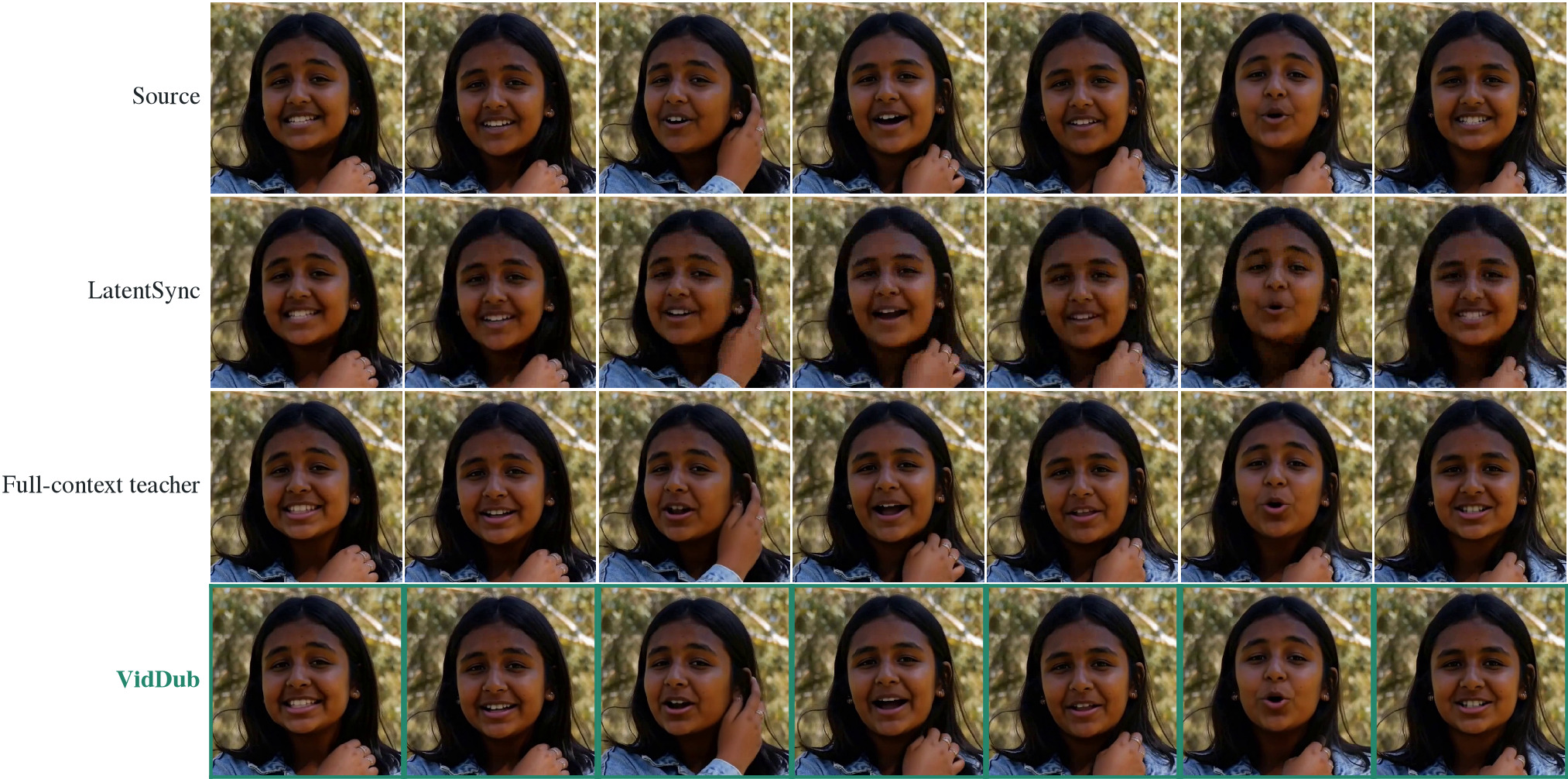}
\caption{VidDub-Bench case 1, same-audio track. Rows compare Source, LatentSync, the full-context teacher, and final full-context VidDub. All rows use the same seven temporal positions.}
\label{fig:qual-causaldub-1-same}
\end{figure*}

\begin{figure*}[p]
\centering
\includegraphics[width=.985\textwidth]{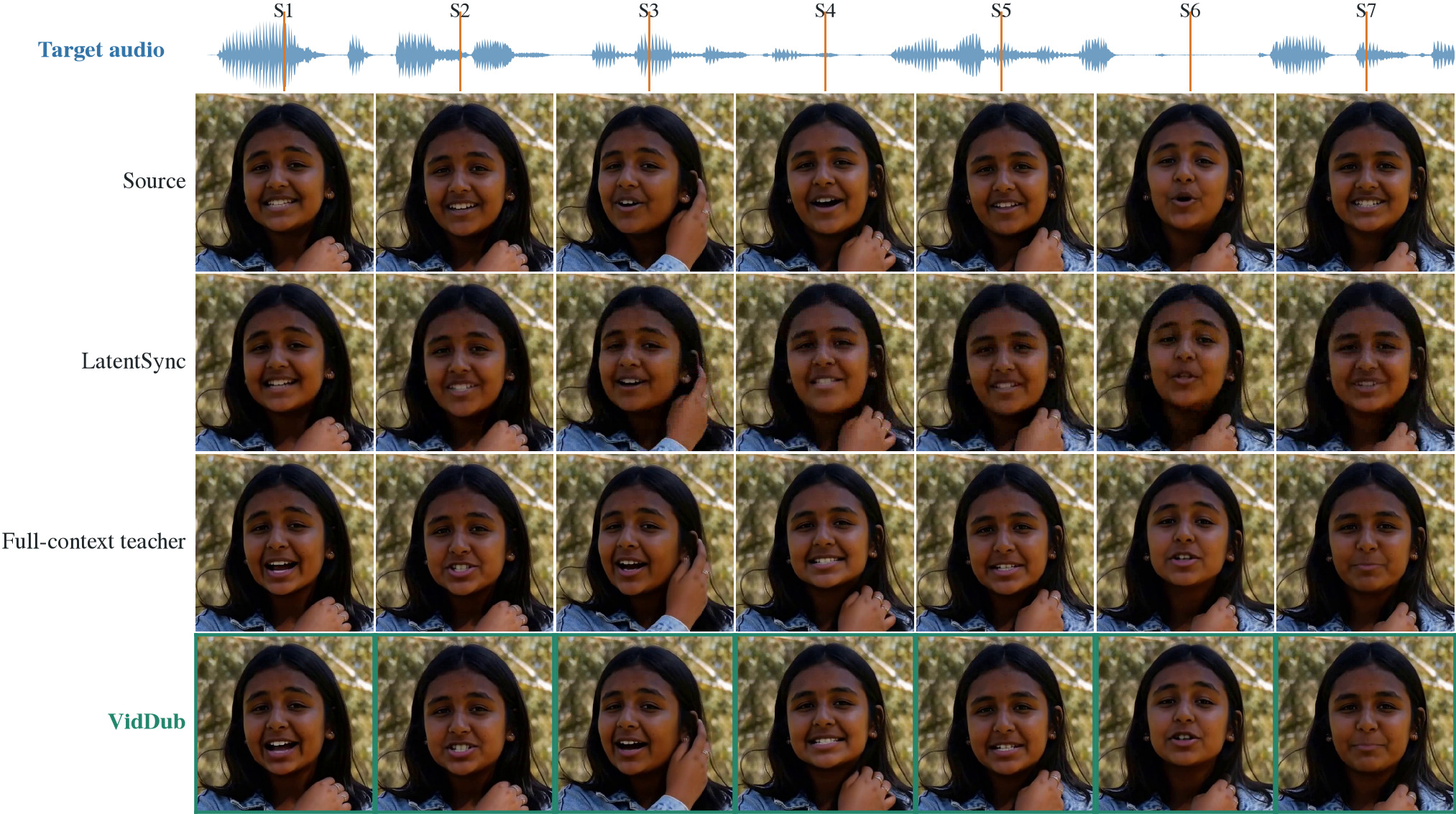}
\caption{VidDub-Bench case 1, cross-audio track. Rows compare Source, LatentSync, the full-context teacher, and final full-context VidDub. The panel includes the target waveform and aligned acoustic segments.}
\label{fig:qual-causaldub-1-cross}
\end{figure*}

\begin{figure*}[p]
\centering
\includegraphics[width=.985\textwidth]{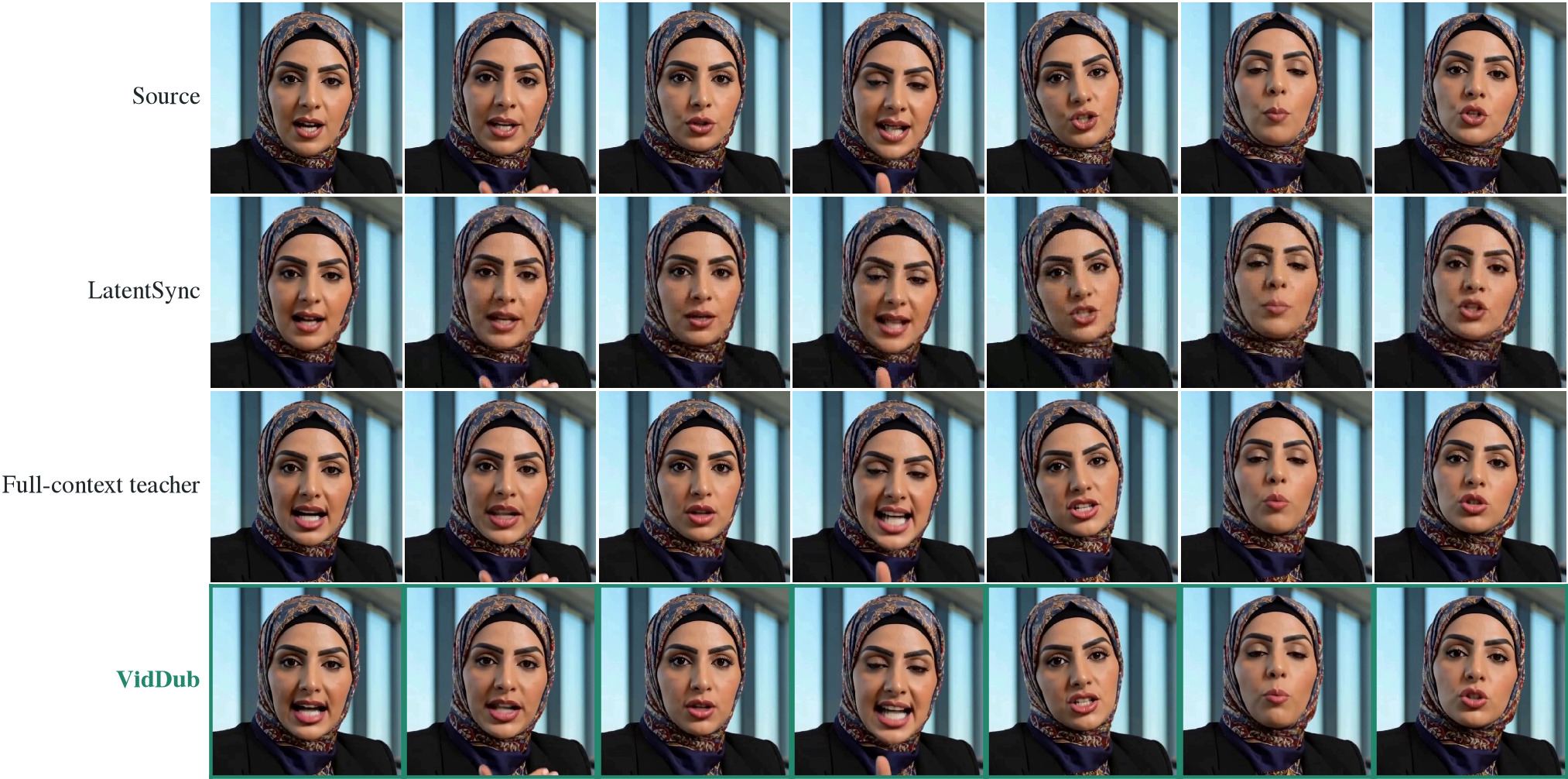}
\caption{VidDub-Bench case 2, same-audio track. Rows compare Source, LatentSync, the full-context teacher, and final full-context VidDub. All rows use the same seven temporal positions.}
\label{fig:qual-causaldub-2-same}
\end{figure*}

\begin{figure*}[p]
\centering
\includegraphics[width=.985\textwidth]{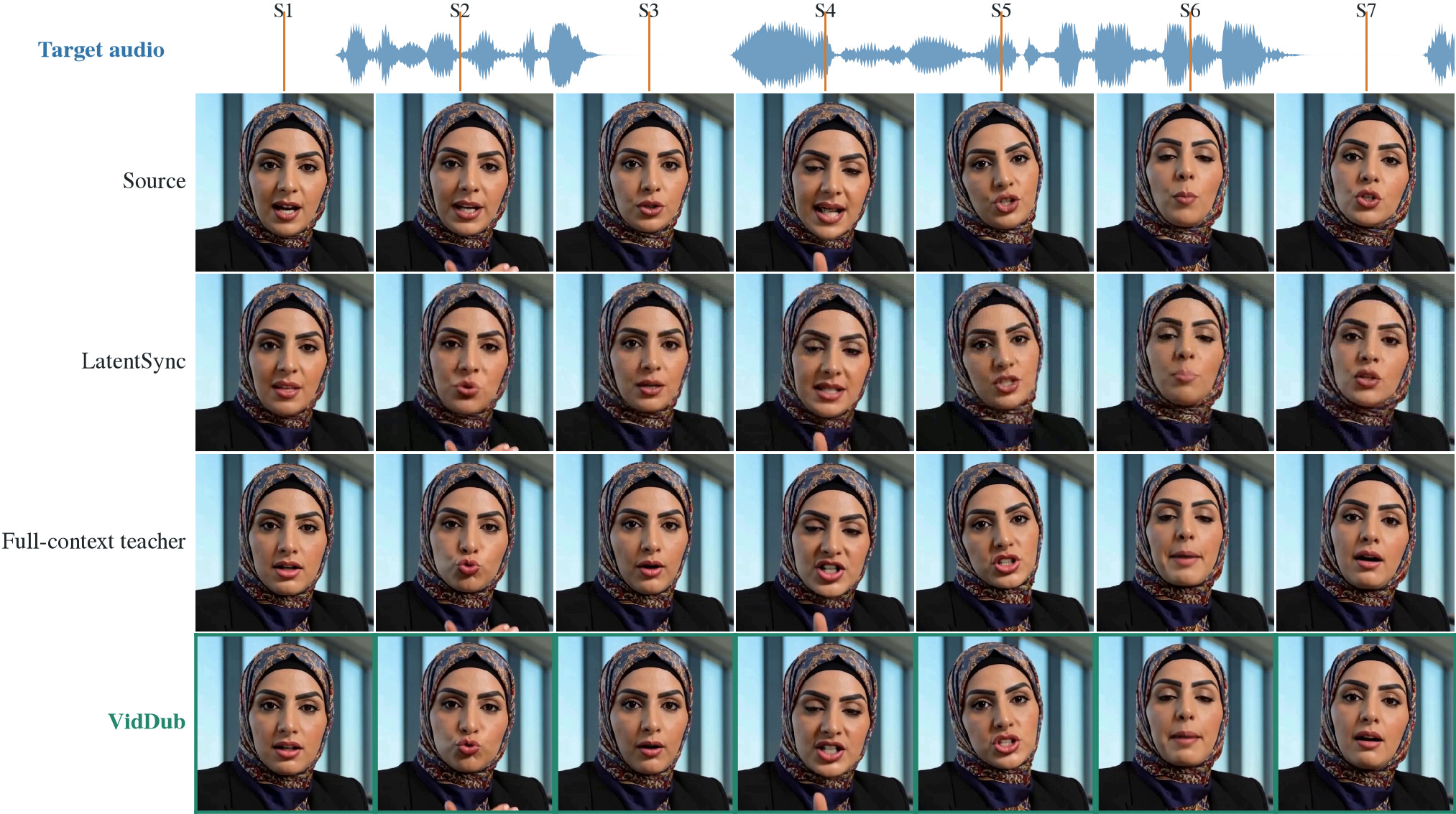}
\caption{VidDub-Bench case 2, cross-audio track. Rows compare Source, LatentSync, the full-context teacher, and final full-context VidDub. The panel includes the target waveform and aligned acoustic segments.}
\label{fig:qual-causaldub-2-cross}
\end{figure*}

}{%
  \textit{The complete qualitative appendix is generated by
  \texttt{figures/make\_video\_qualitative.py}.}
}

\section{Blinded Human Evaluation}
\label{supp:human-study}
This study evaluates the final full-context \method model on generated H3 sources.
Fifteen bilingual participants completed anonymized A/B trials covering 50 sources
and three opponents. Each trial asks which video is
better in lip synchronization, mouth realism, identity preservation, temporal
stability, and overall quality. Left/right assignment and the order of the first
four criteria are randomized; overall quality is always shown last. The interface
exposes neither method identity nor metrics. Table~\ref{tab:supp-human-study}
reports the retained post-aggregation export: one source-level
VidDub/opponent/tie decision per opponent and criterion, hence each cell triplet
sums to 150 rather than 750 raw participant ratings. Participant-level responses
were not retained, so we make no confidence-interval, significance, or
inter-rater-agreement claim; the table is descriptive, and the main-paper
conclusions do not rely on it.

\begin{figure*}[t]
\centering
\includegraphics[width=0.92\textwidth]{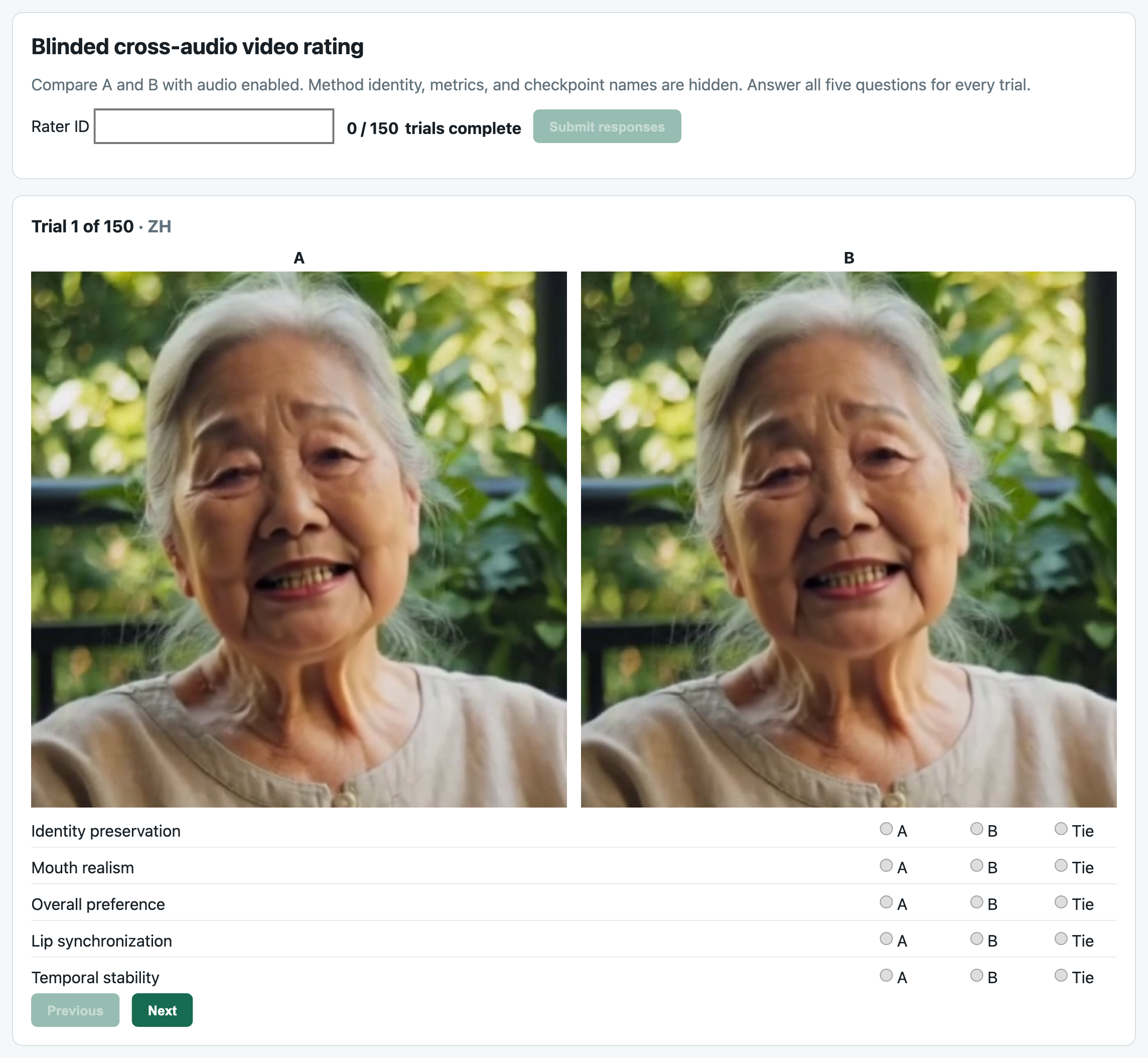}
\caption{\textbf{Blinded rating interface.} Each trial shows only an anonymized
A/B pair. The first four criteria are randomized, and overall quality is shown
last in every trial of the study.}
\label{fig:supp-blind-rater}
\end{figure*}
\IfFileExists{tables/human_study_detailed.tex}{%
\begin{table*}[t]
\centering
\caption{\textbf{Blinded human evaluation of full-context VidDub.} The retained post-aggregation export contains 150 source-level decisions per opponent; participant-level responses are unavailable. Each entry is
VidDub/Opp./Tie; ties are retained.}
\label{tab:supp-human-study}
\small
\setlength{\tabcolsep}{4.2pt}
\begin{tabular}{@{}lccccc@{}}
\toprule
Opponent & Lip sync & Mouth realism & Identity & Stability & Overall \\
\midrule
Teacher & 28/29/93 & 35/38/77 & 18/18/114 & 29/31/90 & 38/37/75 \\
LatentSync & 123/4/23 & 124/19/7 & 46/100/4 & 47/101/2 & 73/77/0 \\
Lip Forcing & 99/28/23 & 109/32/9 & 56/22/72 & 65/70/15 & 98/47/5 \\
\bottomrule
\end{tabular}
\end{table*}

}{%
\textit{The generated blinded-human-evaluation table is required for the submission
build. It is produced only from complete, package-hash-validated response files.}
}

\section{Reproducibility Checklist}
\label{supp:reproducibility}

The reproducibility package is defined by frozen split manifests, preprocessing
thresholds, exact model and code revisions, checkpoint hashes, optimizer and EMA
state, inference schedules, metric versions, failed-sample accounting, and commands
used to generate every table and supplementary video. Archived records include
training and DMD configurations, paired-bootstrap outputs, final-evaluation
protocols, and figure-selection manifests.
Training completion by itself is never treated as evidence of quality gain.

\section{Ethical Considerations}

High-fidelity lip synchronization can support translation, accessibility, and
post-production, but can also facilitate impersonation and misattributed speech.
\method edits a source video and can make a real person appear to say words they
never said. We use
only public research benchmarks and generated H3 subjects for figures; no private
recordings of identifiable people are released. We will not redistribute source
media whose licenses or consent basis do not permit redistribution, and will publish
a dataset/model card documenting provenance, permitted use, and a removal-request
channel before releasing artifacts. Model access will require a use agreement
forbidding non-consensual impersonation; released outputs will include provenance
metadata, and detector baselines will accompany the model. The blinded human study uses only generated H3
subjects, so no real person's likeness is edited for it; raters are adult volunteers who
may stop at any time, and nothing beyond an anonymous rater ID is stored.
\end{document}